\pdfoutput=1
\documentclass[10pt, logo, twocolumn, copyright]{nv}

\usepackage{graphicx}
\definecolor{nvidiagreen}{HTML}{76B900}

\usepackage{multirow}
\usepackage{makecell}
\usepackage{amsmath}

\newcommand{\model}{PixelDiT}

\usepackage{xcolor}
\definecolor{pearDark}{RGB}{34,139,34}
\definecolor{mygreen}{RGB}{34,139,34}
\definecolor{mylightblue}{RGB}{0,162,230}
\definecolor{deepyellow}{RGB}{255,215,0}
\definecolor{nvgreen}{RGB}{118, 185, 0}

\makeatletter
\renewcommand{\titlefont}{\normalfont\bfseries\fontsize{17}{20}\selectfont}
\makeatother

\makeatletter
\renewcommand{\copyrightext}{
  \footerfont
  $^*$ Work was done while Yongsheng was an intern at NVIDIA.\quad Contact: \texttt{\{yongshengy, wxiong\}@nvidia.com} \\
  \textcopyright\, \the\year{} NVIDIA. All rights reserved.%
}
\makeatother

\usepackage{mdframed}
\usepackage{color}
\usepackage{xcolor}
\usepackage[utf8]{inputenc} 
\usepackage[T1]{fontenc}    

\usepackage{amsfonts}       
\usepackage{nicefrac}       
\usepackage{microtype}      
\usepackage{multirow}
\usepackage{multicol}
\usepackage{tabto}
\usepackage{xspace}
\usepackage{amsmath}
\usepackage{adjustbox}
\usepackage{enumitem}
\usepackage{wrapfig}
\usepackage{dblfloatfix}
\usepackage{times}
\usepackage{verbatim}
\usepackage{amssymb}
\usepackage{mathtools}
\usepackage{caption}
\usepackage{subcaption}
\usepackage{array}
\usepackage{colortbl}
\usepackage{booktabs}
\usepackage{bbm}
\usepackage{makecell}
\usepackage{float}
\usepackage{siunitx}
\usepackage{pifont}
\usepackage{marvosym}
\usepackage{listings}
\usepackage{pdflscape}
\usepackage{footmisc}
\usepackage{url}
\usepackage{tabularx}
\usepackage{arydshln}
\usepackage{hhline}
\usepackage{diagbox}
\usepackage{tcolorbox}
\usepackage[nameinlink]{cleveref}
\usepackage{hyperref}
\usepackage[square,sort,comma,numbers]{natbib} 
\usepackage{fp}
\usepackage{authblk}
\usepackage{xspace}
\usepackage{inconsolata}

\crefname{section}{Sec.}{Sec.}
\crefname{proposition}{Proposition.}{Proposition.}
\crefname{equation}{Eq.}{Eqs.}
\crefname{figure}{Fig.}{Figs.}
\crefname{table}{Tab.}{Tabs.}
\crefname{algorithm}{Algorithm}{Algorithms}
\crefname{appendix}{Appendix}{Appendices}
\Crefname{thm}{Thm}{Thm}
\definecolor{codebg}{RGB}{245, 245, 245} 
\definecolor{keywordcolor}{RGB}{0, 0, 153} 
\definecolor{commentcolor}{RGB}{34, 139, 34} 
\definecolor{stringcolor}{RGB}{163, 21, 21}
\definecolor{numbercolor}{RGB}{128, 128, 128}

\title{
\textcolor{nvidiagreen}{\model{}}: %
\textcolor{nvidiagreen}{Pixel} %
\textcolor{nvidiagreen}{Di}ffusion %
\textcolor{nvidiagreen}{T}ransformers for Image Generation%
}

\author{
\centering
\fontsize{10pt}{18pt}\selectfont
Yongsheng Yu\textsuperscript{1,2 *} ~~
Wei Xiong\textsuperscript{1 \dag} ~~
Weili Nie\textsuperscript{1} ~~
Yichen Sheng\textsuperscript{1} ~~
Shiqiu Liu\textsuperscript{1} ~~
Jiebo Luo\textsuperscript{2} ~~
\\
\vspace{2mm}
{\centering \normalsize \textsuperscript{1}NVIDIA ~~
\textsuperscript{2}University of Rochester} \\
\vspace{2mm}
\centering \normalsize  \dag { {Project Lead and Main Advising}}
}

\begin{abstract}\small
\textbf{Abstract:} Latent-space modeling has been the standard for Diffusion Transformers (DiTs). However, it relies on a two-stage pipeline where the pretrained autoencoder introduces lossy reconstruction, leading to error accumulation while hindering joint optimization. To address these issues, we propose \textbf{PixelDiT}, a single-stage, end-to-end model that eliminates the need for the autoencoder and learns the diffusion process directly in the pixel space. PixelDiT adopts a fully transformer-based architecture shaped by a dual-level design: a patch-level DiT that captures global semantics and a pixel-level DiT that refines texture details, enabling efficient training of a pixel-space diffusion model while preserving fine details. PixelDiT achieves \textbf{1.61} FID on ImageNet 256 and \textbf{1.81} FID on ImageNet 512, surpassing existing pixel generative models. We further extend PixelDiT to text-to-image generation and  pretrain it at the $1024^{2}$ resolution in pixel space. It achieves \textbf{0.74} on GenEval and \textbf{83.5} on DPG-bench, approaching the best latent diffusion models.
    \newline
    \textbf{Links:} \hspace{2pt}
    {\hypersetup{urlcolor=nvidiagreen}
    \href{https://github.com/NVlabs/PixelDiT}{GitHub Code} |
    \href{https://huggingface.co/collections/nvidia/pixeldit}{HF Models} |
    \href{https://pixeldit.github.io}{Project Page}
    }\vspace{-5mm}
\end{abstract}

\begin{document}

\twocolumn[{
\renewcommand\twocolumn[1][]{#1}
\maketitle
\begin{center}
    \centering
    \captionsetup{type=figure}
    \begin{subfigure}[t]{\textwidth}
        \centering
        \includegraphics[width=\textwidth,page=1]{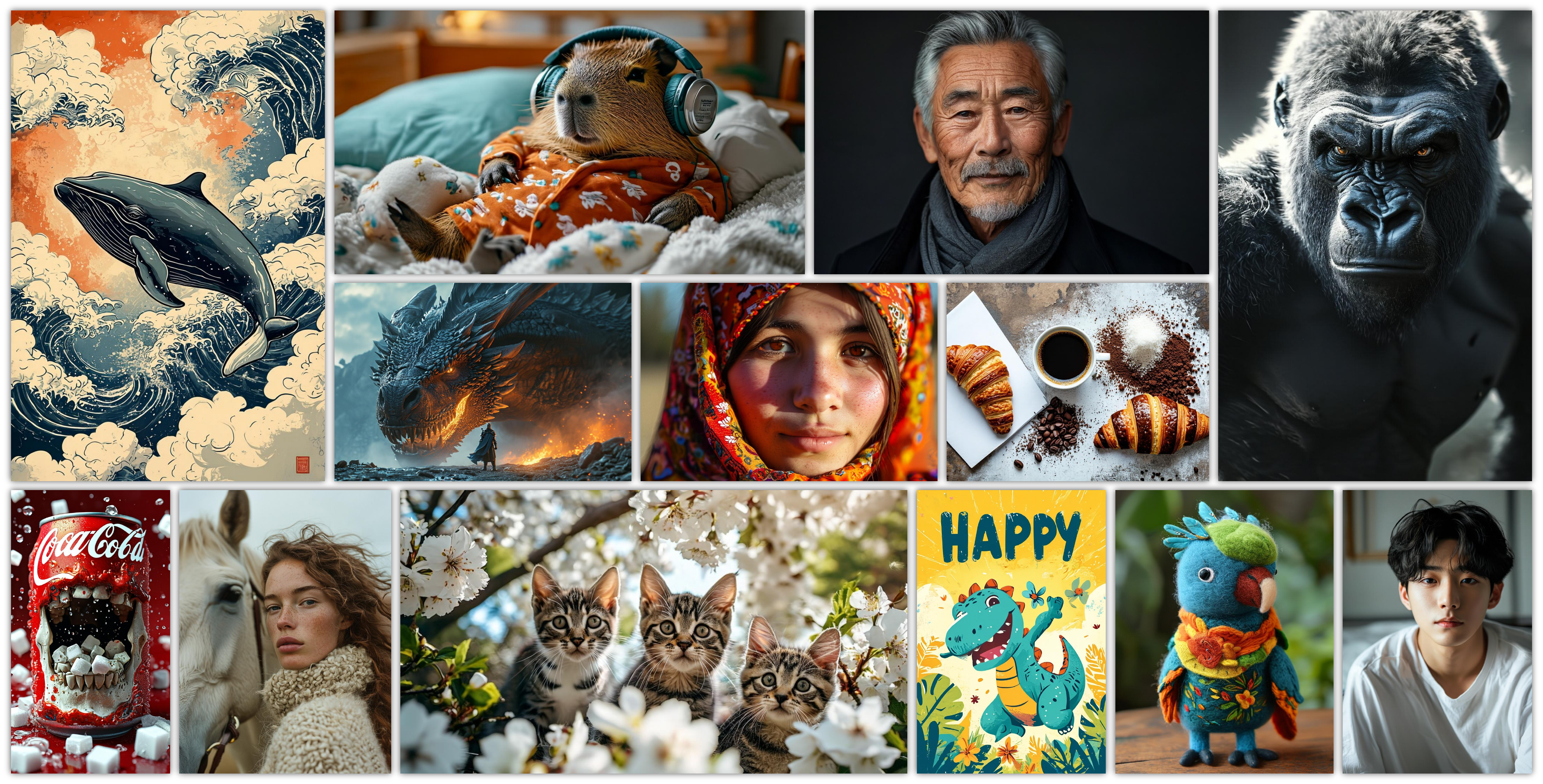}
        \caption{High-resolution text-to-image samples at the megapixel scale (approximately \(1024{\times}1024\)) generated by \model{}-T2I, which is  \textit{directly trained on pixel space}.}
        \label{fig:teaser:a}
    \end{subfigure}
    \begin{subfigure}[t]{\textwidth}
        \centering
        \includegraphics[width=\textwidth]{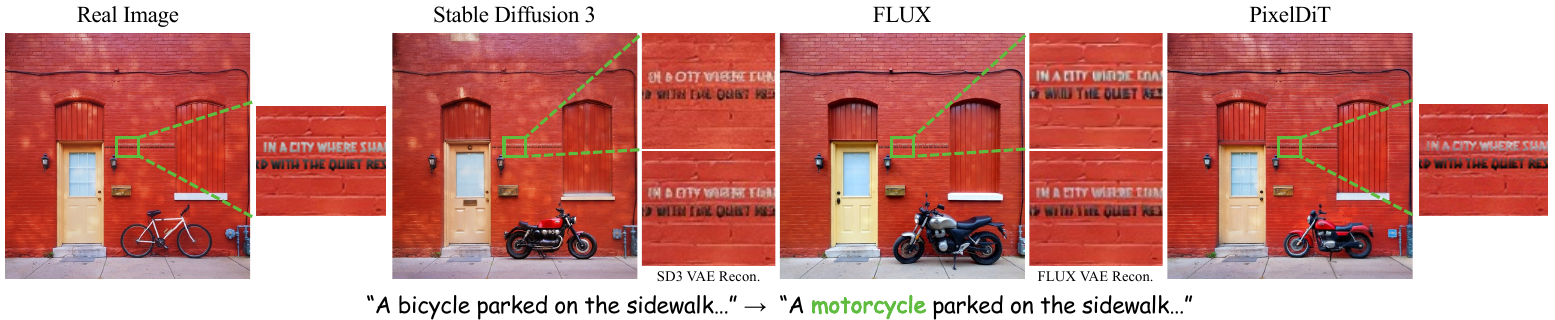}
        \caption{Train-free image editing with FlowEdit~\cite{kulikov2025flowedit} based on different diffusion models. The imperfect reconstruction of VAEs used by FLUX~\cite{flux} and Stable Diffusion~3~\cite{SD3} produce severe and non-invertible distortions on the small details such as the scene texts on the wall, therefore the subsequent editing of the latent diffusion models fails.  In contrast, by denoising directly in pixel space without a VAE, our \model{} avoids VAE reconstruction artifacts and helps maintain background consistency during local manipulations. For a fair comparison, all models are evaluated with \texttt{nmin}$=0$ in FlowEdit to preserve more structures.}
        \label{fig:teaser:b}
    \end{subfigure}
    \setcounter{figure}{0}
    \captionof{figure}{\textbf{Visual results of \model{}} on text-to-image generation and training-free image editing. Please zoom in for the details. Additional examples are provided in the Appendix.}
    \label{fig:teaser}
\end{center}
}]

\section{Introduction}

Latent diffusion models (LDMs)~\cite{LDM} perform denoising in a compressed representation space and have become the standard paradigm for Diffusion Transformers~\cite{dit,sit}. This choice yields substantial savings in compute and memory. However, it also inherits two structural limitations. First, LDMs couple diffusion to a separately pretrained autoencoder whose reconstruction objective is only partially aligned with the downstream generative objective~\cite{lgt,repa-e}. Second, the autoencoder introduces a lossy reconstruction that can remove high-frequency details and cap sample fidelity for image generation and editing tasks even when the diffusion model is strong~\cite{SD3,liu2024playground,Dalle-3,zheng2025rae}, as shown in Figure~\ref{fig:teaser:b}. These limitations motivate us to revisit pixel-space diffusion, where both learning and sampling operate directly in the original pixels without autoencoders.

The core challenge in pixel-space diffusion can be framed as \emph{pixel modeling}. By pixel modeling, we refer to the mechanism of capturing dense, per-pixel interactions and high-frequency details, which is distinct from the semantic structural generation typically handled by coarse patch tokens. Effectively modeling these per-pixel tokens is crucial for texture fidelity but computationally expensive. Prior attempts expose a fundamental trade-off in how pixel interactions are organized. One line of work adopts aggressive patchification~\cite{pixnerd,jetformer,farmer,epg, jit} to keep attention affordable, but this significantly weakens per-pixel token modeling and hinders the generation of finer visual contents. Another line pushes toward near-pixel granularity (e.g., small patch sizes or U-ViT-like designs) to better preserve details~\cite{pixelflow,simplediffusion,sid2}, but the global attention must process very long token sequences with quadratic complexity, or rely on heavy decoder-style stacks, resulting in high training and sampling costs. Cascaded  pipelines~\cite{cdm,li2025fractal} mitigate some cost, but may introduce additional stages and accumulated errors. Taken together, these observations suggest that the obstacle to practical pixel-space diffusion is the lack of an \emph{efficient pixel modeling mechanism} that can model both global semantics and per-pixel updates.

To address these challenges, we propose \model{}, a single-stage, fully transformer-based diffusion model that performs end-to-end training and sampling in pixel space while explicitly structuring pixel modeling. \model{} decouples image semantics from per-pixel learning with a dual-level architecture design: a patch-level DiT with an aggressive patch size that performs long-range attention on short patch token sequences to capture global layout and content, and a pixel-level DiT that performs dense, per-pixel token modeling to refine local texture details. We propose two key techniques to make this design effective: (i) a pixel-wise AdaLN modulation that conditions each pixel token using semantic tokens, aligning per-pixel updates with global context; and (ii) a pixel token compaction mechanism that compresses each pixel token before full attention and decompresses them back afterward, enabling per-pixel token modeling while keeping global attention efficient. The dual-level pathways, together with efficient pixel modeling via pixel-wise AdaLN and token compaction, yield high training efficiency and faster convergence while preserving fine details.

Extensive experiments show that \model{} generates high-quality images when trained end-to-end on pixel space \emph{without any autoencoders}. On ImageNet $256\times256$, \model{} achieves an FID of 1.61, significantly outperforming existing pixel-space models. We further demonstrate the \emph{scalability} of our architecture by extending \model{} to text-to-image generation. Using multi-modal DiT blocks in the patch-level pathway, we directly train \model{} at $1024^2$ resolution in pixel space. \model{} can generate high-fidelity text-aligned 1K resolution images as shown in Figure~\ref{fig:teaser:a} and achieves competitive scores on standard benchmarks compared to state-of-the-art latent diffusion models. Moreover, operating directly in pixel space enables \model{} to bypass VAE reconstruction artifacts, leading to significantly better content preservation for image editing tasks, as shown in Figure~\ref{fig:teaser:b}. 

We highlight our main \textbf{contributions} as follows:
\begin{itemize}
    \item We propose \model{}, a single-stage, fully transformer-based pixel-space diffusion model that is trained end-to-end without a separate autoencoder.
    \item We demonstrate that efficient pixel modeling is the key factor to practical pixel-space diffusion and propose a dual-level DiT architecture that disentangles the learning of global semantics from pixel-level texture details.
    \item We introduce a pixel-wise AdaLN modulation mechanism and a pixel token compaction mechanism that jointly enable dense per-pixel token modeling.
    \item \model{} achieves high image quality in both class-conditioned image generation and text-to-image generation, significantly outperforming existing pixel-space generative models and approaching the state-of-the-art latent diffusion models.
\end{itemize}

\begin{figure*}[t]
    \centering
    \includegraphics[width=\linewidth]{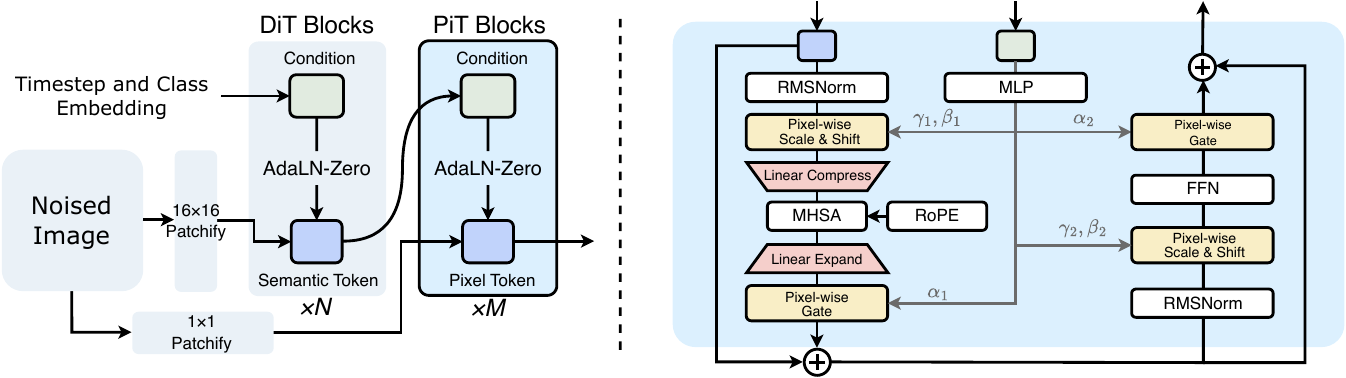}
    \caption{\textbf{Overview of \model{}}: a dual-level, fully transformer-based diffusion architecture that operates directly in pixel space. The left figure shows the overall framework of \model{}, while the right figure illustrates the detailed structure of the PiT blocks.}
    \label{fig:main_method}
\end{figure*}
\section{Related Works}
\subsection{Latent Diffusion Models with Autoencoders}
Latent diffusion models (LDMs) perform denoising in a latent space produced by an autoencoder, delivering substantial compute and memory savings 
and enabling training at higher resolutions with larger backbones on a fixed budget~\cite{LDM}.  
In practice, most LDMs adopt a variational autoencoder (VAE) that trades off reconstruction fidelity and compression rate; a rich line of work improves the autoencoder via better architecture or learning objective, including stronger compression schemes, tokenizers~\cite{dcae,dcae1p5,maetok,yu2025zipir} and analysis of the reconstruction–generation optimization dilemma~\cite{lgt}. End-to-end approaches have also been explored: REPA-E jointly tunes the VAE and diffusion transformer to align the latent representation with the generative objective~\cite{repa-e}. Recently, several works~\cite{shi2025svg,zheng2025rae} replace the variational bottleneck with representation autoencoders and report competitive latents without explicit variational modeling.

\noindent\textbf{Why revisit pixel space?} Despite their efficiency, LDMs inherit a reconstruction bottleneck: sample fidelity is bounded by the autoencoder, and aggressive compression tends to remove high-frequency details and fine structures~\cite{LDM,dcae}. Pretraining or co-training a large autoencoder adds additional data and compute overhead compared to pixel-space training. Moreover, misalignment between the autoencoder’s reconstruction objective and the downstream generative objective introduces distribution shift in latent space (e.g., texture smoothing or color shifts) that the diffusion model must compensate for~\cite{lgt,repa-e}. Finally, decoding latency at sampling time incurs additional cost compared to pixel-space models.
\subsection{Pixel-Space Diffusion Models}
Pixel-space diffusion models preceded latent-space methods and remain an active area of research. Early work establishes high image quality via direct denoising in pixel space~\cite{adm}, while cascaded diffusion improves high-resolution synthesis through multi-scale pipelines~\cite{cdm}. However, the quadratic compute and memory cost in image resolution renders end-to-end training at megapixel scales prohibitively expensive.
Recent efforts revisit pixel space with improved architectures and training: JetFormer formulates autoregressive generation over raw pixels and text~\cite{jetformer}; Simple Diffusion proposes simplified, memory-efficient convolutional networks with skip connections~\cite{simplediffusion,sid2}; Fractal Generative Models introduce fractal designs for long-range structure~\cite{li2025fractal}; PixelFlow develops hierarchical flow-based pixel-space models~\cite{pixelflow}; and PixNerd employs lightweight neural field layers for efficient pixel-space diffusion~\cite{pixnerd}. In contrast to these efforts, we present a purely transformer-based pixel-space diffusion architecture trained directly at $1024^2$ resolution.

Concurrent to our work, EPG~\cite{epg} adopts a two-stage framework bridging self-supervised pre-training and generative fine-tuning, while FARMER~\cite{farmer} integrates normalizing flows with autoregressive modeling to handle high-dimensional pixels. Distinctively, JiT~\cite{jit} demonstrates that plain Transformers can efficiently model high-dimensional data by predicting clean images ($x_0$-prediction).

\section{Method}

In this section, we present \model{}, a transformer-based diffusion model that directly performs denoising in pixel space. Our objective is to make pixel token modeling computationally efficient while preserving the convergence behavior and sample quality that have motivated latent space approaches.

\subsection{Dual-level DiT Architecture}

As illustrated in Figure~\ref{fig:main_method}, we adopt a dual-level transformer organization that concentrates semantic learning on a coarse patch-level pathway and leverages dedicated \textbf{Pi}xel \textbf{T}ransformer (PiT) blocks in the pixel-level pathway for detail refinement. This organization allows most semantic reasoning to occur on the low-resolution grid, which reduces the burden on the pixel-level pathway and accelerates learning, consistent with observations found in~\cite{zheng2025rae,ddt,pernias2024wurstchen}.

\noindent\textbf{Patch-level architecture: }
Let the input image be $x\in\mathbb{R}^{B\times C\times H\times W}$. We form non-overlapping $p\times p$ patch tokens $x_{\text{patch}}\in\mathbb{R}^{B\times L\times (p^2C)}$, where $L{=}(\frac{H}{p})(\frac{W}{p})$ is the number of tokens, and project them to hidden size $D$:
\begin{align}
 s_0 &= W_{\text{patch}}\,x_{\text{patch}},\\
 c &= \mathrm{SiLU}(W_t t + W_y y + b)\in\mathbb{R}^{B\times 1\times D}.
\end{align}
Following \cite{lgt}, we augment the DiT block by replacing LayerNorm with RMSNorm and applying 2D RoPE in all attention layers. The patch-level pathway consists of $N$ augmented DiT blocks; for block $i$, we write
\begin{align}
 \tilde s_i &= \mathrm{RMSNorm}(s_i),\\
 \overline{s}_i &= s_i 
	\;{+}\; \alpha_1(c)\odot \mathrm{Attn}\!\big(\gamma_1(c)\odot \tilde s_i + \beta_1(c); \mathrm{RoPE}\big),\\
 s_{i+1} &= \overline{s}_i + \alpha_2(c)\odot \mathrm{MLP}\!\big( \nonumber\\
&\quad\gamma_2(c)\odot \mathrm{RMSNorm}(\overline{s}_i) + \beta_2(c)\big),
\end{align}
where AdaLN modulation parameters are produced from the \emph{global} conditioning vector $c$ and then broadcast across the $L$ patch tokens. This global-to-patch broadcasting applies identical per-feature AdaLN parameters to all patch tokens (i.e., token-independent at the patch level), in contrast to the \emph{pixel-wise} AdaLN used later in the pixel-level pathway.

After $N$ blocks, we obtain semantic tokens $s_N\in\mathbb{R}^{B\times L\times D}$. In the spirit of designs \cite{ddt,zheng2025rae}, we define the conditioning signal for the pixel-level pathway as $s_\textrm{cond} := s_N + t$, where $t$ is the timestep embedding. These tokens provide semantic context to the PiT blocks via pixel-wise AdaLN.

\noindent\textbf{Pixel-level architecture: } The pixel-level DiT is composed of $M$ layers of PiT Blocks. It takes the pixel tokens and the output of the patch-level DiT $s_\textrm{cond}$ as inputs to perform the pixel token modeling and generate the final result. The details of each PiT block are described below.

\paragraph{Design notes.} The patch-level pathway exclusively processes patch tokens to capture global semantics. By delegating detail refinement to the pixel-level pathway, we can employ larger patch sizes $p$, which shortens the sequence length and accelerates inference while preserving per-pixel fidelity. Furthermore, the pixel-level pathway operates with a reduced hidden dimension $D_{\textrm{pix}} \ll D$ (e.g., $D_{\textrm{pix}}{=}16$), ensuring that dense per-pixel computations remain highly efficient.

\begin{figure}[t]
    \centering
    \includegraphics[width=0.72\linewidth]{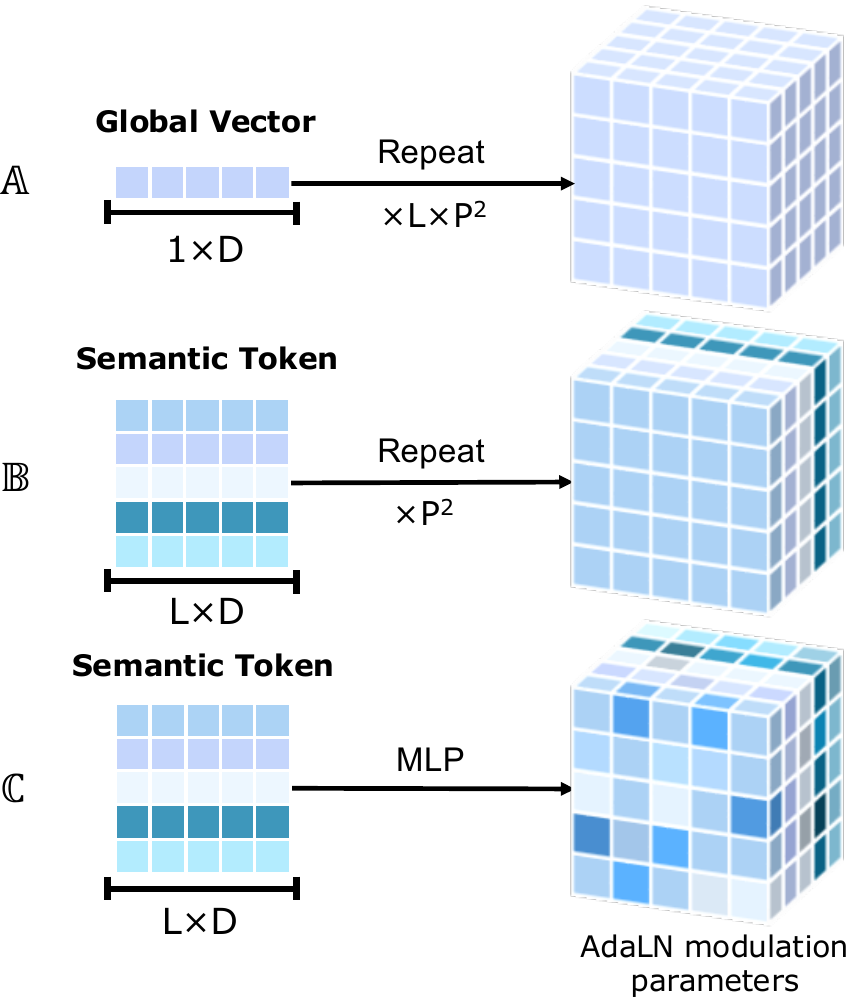}
    \vspace{-1mm}
    \caption{\textbf{AdaLN modulation strategies}. ($\mathbb{A}$) A naive AdaLN broadcasts a global conditioning vector to all pixels. ($\mathbb{B}$) Patch-wise AdaLN expands semantic tokens to the $p^2$ pixels within each patch. ($\mathbb{C}$) Pixel-wise AdaLN applies an MLP to each semantic token to produce per-pixel scale, shift, and gating parameters, enabling fully context-aligned updates at every pixel.}
    \label{fig:pixeladaln}  
\end{figure}

\begin{table*}[ht!]
    \footnotesize
    \centering
    \begin{tabular*}{\linewidth}{@{\extracolsep{\fill}}lccccccc@{}}
    \toprule
    \multirow{2}{*}{\textbf{Method}} & \multirow{2}{*}{\makecell{\textbf{Training} \\ \textbf{Epochs}}} & \multirow{2}{*}{\textbf{\#params}} & \multicolumn{5}{c}{\textbf{Generation@256}} \\
    \cmidrule(lr){4-8}
     &  &  & \textbf{gFID}$\downarrow$ & \textbf{sFID}$\downarrow$ & \textbf{IS}$\uparrow$ & \textbf{Precision}$\uparrow$ & \textbf{Recall}$\uparrow$ \\
    \midrule
    \multicolumn{8}{c}{\textbf{Latent Generative Models}} \\
    \midrule
    LDM-4-G~\cite{LDM} & 170 & 400M & 3.60 & - & 247.6 & 0.87 & 0.48 \\
    DiT-XL~\cite{dit}        & 1400 & 675M &  2.27 & 4.60  & 278.2  & 0.83 & 0.57 \\
    SiT-XL~\cite{sit}        & 1400 & 675M &  2.06 & 4.50  & 270.3  & 0.82 & 0.59 \\
    MaskDiT~\cite{maskdit} & 1600 & 675M  &  2.28 & 5.67 & 276.5 & 0.80 & 0.61 \\
    REPA~\cite{repa} & 800 & 675M &  1.42 & 4.70 & 305.7 & 0.80 & 0.65 \\
    LightningDiT~\cite{lgt} & 800 & 675M &  1.35 & 4.15 & 295.3 & 0.79 & 0.65 \\
    SVG-XL~\cite{shi2025svg} & 1400 & 675M & 1.92 & - & 264.9 & - & - \\
    DDT-XL~\cite{ddt} & 400 & 675M & 1.26 & - & 310.6 & 0.79 & 0.65 \\
    RAE-XL~\cite{zheng2025rae} & 800 & 839M &  \textbf{1.13} & - & 262.6 & 0.78 & 0.67 \\
    \midrule
    \multicolumn{8}{c}{\textbf{Pixel Generative Models}} \\
    \midrule
    StyleGAN-XL~\cite{styleganxl} & / & / & 2.30 & 4.02 & 265.1 & 0.78 & 0.53 \\
    ADM-U~\cite{adm} & 400 & 554M & 4.59 & 5.25 & 186.7 & 0.82 & 0.52 \\
    CDM~\cite{cdm} & 2160 & / & 4.88 & - & 158.7 & - & - \\
    RIN \cite{rin} & 480 & 410M & 3.42 & - & 182.0 & - & - \\
    VDM++ \cite{vdm++} & / & / & 2.12 & - & 267.7 & - & - \\
    JetFormer~\cite{jetformer} & / & 2.8B & 6.64 & - & - & 0.69 & 0.56 \\
    Simple Diffusion~\cite{simplediffusion} & / & 2.0B & 2.44 & - & 256.3 & - & - \\
    FractalMAR-H~\cite{li2025fractal} & 600 & 844M & 6.15 & - & 348.9 & 0.81 & 0.46 \\
    FARMER~\cite{farmer} & 320 & 1.9B & 3.60 & - & 269.2 & 0.81 & 0.51 \\
    EPG-XXL/16~\cite{epg} & 600 & 789M & 1.81 & - & 294.6 & 0.80 & 0.61 \\
    PixelFlow-XL~\cite{pixelflow} & 320 & 677M & 1.98 & 5.83 & 282.1 & 0.81 & 0.60 \\
    PixNerd-XL~\cite{pixnerd} & 320 & 700M & 1.93 & - & 298.0 & 0.80 & 0.60 \\
    JiT-G~\cite{jit} & 600 & 2B & 1.82 & - & 292.6 & 0.79 & 0.62 \\
    \midrule
    \model{}-XL & 80 & 797M & 2.36 & 5.11 & 282.3 & 0.80 & 0.57 \\
    \model{}-XL & 320 & 797M & \textbf{1.61} & 4.68 & 292.7 & 0.78 & 0.64 \\
    \bottomrule
    \end{tabular*}
    \caption{\textbf{Quantitative results} on ImageNet 256$\times$256 for class-conditioned generation.}
    \label{tab:imagenet256_main}
\end{table*}

\subsection{Pixel Transformer Block}

Each PiT block has two core components. First, pixel-wise AdaLN enables dense conditioning at the level of individual pixels, aligning per-pixel updates with global context. Second, a pixel token compaction mechanism reduces redundancy among pixel tokens so that global attention operates on a manageable sequence length.

\paragraph{Pixel-wise AdaLN Modulation.}
In the pixel-level pathway, each image is embedded into one token per pixel with a linear layer:
\begin{align}
 X &\in\mathbb{R}^{B\times C\times H\times W} \xrightarrow{\text{reshape + linear}} \mathbb{R}^{B\times H\times W\times D_{\textrm{pix}}}.
\end{align}
To align with patch-level semantic tokens, we reshape into $B\!\cdot\!L$ sequences of $p^2$ pixel tokens, i.e., $X\in\mathbb{R}^{(B\cdot L)\times p^2\times D_{\textrm{pix}}}$.
For each patch, we form a semantic conditioning token $s_{\text{cond}}\in\mathbb{R}^{(B\cdot L)\times D}$ that summarizes global context. A straightforward patch-wise modulation would repeat the same parameters for all $p^{2}$ pixels within a patch, as illustrated in Figure~\ref{fig:pixeladaln}($\mathbb{B}$). However, this cannot capture dense per-pixel variation. Instead, we assign independent modulation to each pixel by expanding $s_{\text{cond}}$ into $p^{2}$ sets of AdaLN parameters via a linear projection $\Phi:\mathbb{R}^{D}\!\to\!\mathbb{R}^{p^2\cdot 6D_{\textrm{pix}}}$:
\begin{align}
    \Theta \;=\; \Phi(s_{\text{cond}})\in\mathbb{R}^{(B\cdot L)\times p^2\times 6D_{\textrm{pix}}},
\end{align}
and we partition the last dimension of $\Theta$ into six groups of size $D_{\textrm{pix}}$, yielding \((\beta_{1},\gamma_{1},\alpha_{1},\beta_{2},\gamma_{2},\alpha_{2})\in\big(\mathbb{R}^{(B\cdot L)\times p^2\times D_{\textrm{pix}}}\big)^6\).
These modulation parameters are learned and are distinct at every pixel, as illustrated in Figure~\ref{fig:pixeladaln}($\mathbb{C}$).
They are applied to \(X\) via pixel-wise AdaLN, enabling pixel-specific updates; in contrast, patch-wise AdaLN broadcasts a single set of parameters to all pixels within a patch and thus cannot capture such spatial variation.

\begin{figure*}[t]
    \centering
    \setlength{\tabcolsep}{0pt}
    \renewcommand{\arraystretch}{0}
    \begin{tabular}{cccccc}
        \includegraphics[width=0.1666\textwidth]{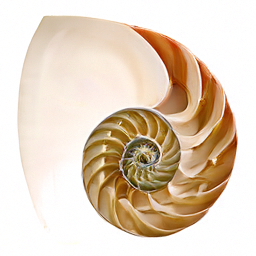} &
        \includegraphics[width=0.1666\textwidth]{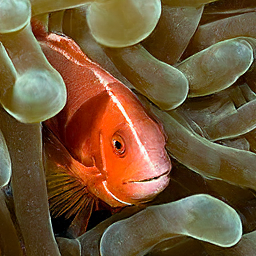} &
        \includegraphics[width=0.1666\textwidth]{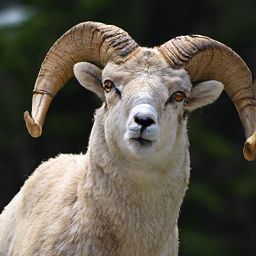} &
        \includegraphics[width=0.1666\textwidth]{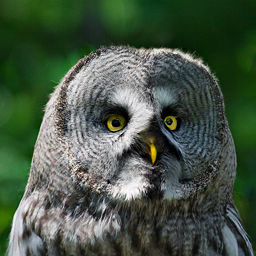} &
        \includegraphics[width=0.1666\textwidth]{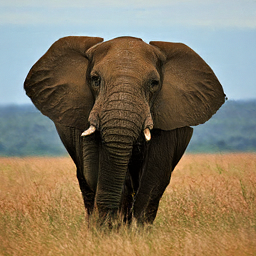} &
        \includegraphics[width=0.1666\textwidth]{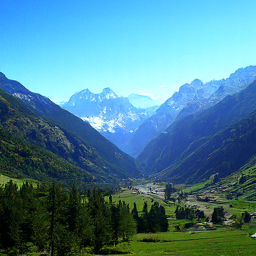} \\
        \includegraphics[width=0.1666\textwidth]{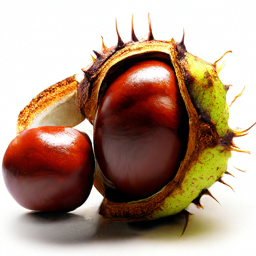} &
        \includegraphics[width=0.1666\textwidth]{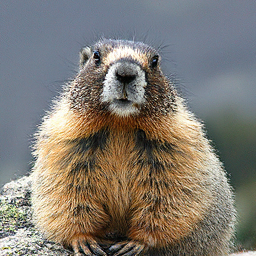} &
        \includegraphics[width=0.1666\textwidth]{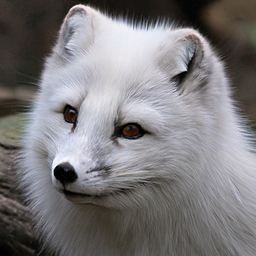} &
        \includegraphics[width=0.1666\textwidth]{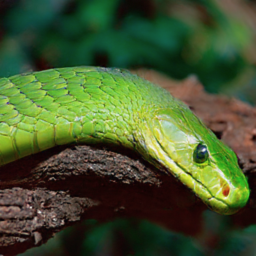} &
        \includegraphics[width=0.1666\textwidth]{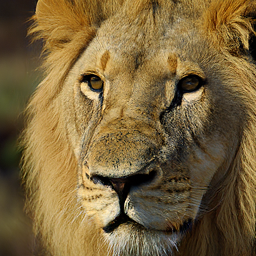} &
        \includegraphics[width=0.1666\textwidth]{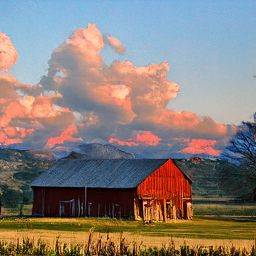}
    \end{tabular}
    \caption{\textbf{Qualitative results} on ImageNet $256 \times 256$ using \model{}-XL. We use a classifier-free guidance scale $\alpha_{\mathrm{cfg}} = 4.0$.}
    \label{fig:qual_imagenet256}
\end{figure*}

\paragraph{Pixel Token Compaction.}

In the pixel-level pathway, 
direct attention over all $H\times W$ pixel tokens is computationally prohibitive. We therefore compress the $p^2$ pixel tokens inside each patch into a compact patch token before global attention, and later expand the attended representation back to pixels. This reduces the attention sequence length from $H\times W$ to $L{=}(\frac{H}{p})(\frac{W}{p})$, a $p^2$-fold reduction; with $p{=}16$, this yields a $256\times$ shrinkage while preserving per-pixel updates through pixel-wise AdaLN and the learned expansion.

We instantiate the compaction operators with a learned flattening: a linear map $\mathcal{C}:\mathbb{R}^{p^2\times D_{\textrm{pix}}}\!\to\!\mathbb{R}^{D}$ that jointly mixes spatial and channel dimensions, paired with an expansion $\mathcal{E}:\mathbb{R}^{D}\!\to\!\mathbb{R}^{p^2\times D_{\textrm{pix}}}$. This compress–attend–expand pipeline keeps global attention efficient. Unlike the lossy bottleneck in VAEs, this mechanism only compresses the representation momentarily for the attention operation. Crucially, this compaction operates purely to reduce the computational overhead of self-attention; it does not compromise fine-grained details, because high-frequency information is preserved through residual connections and learned expansion layers that effectively bypass the pixel-token bottleneck.

\subsection{PixelDiT for Text-to-Image Generation}

We extend the patch-level pathway with multi-modal DiT (MM-DiT) blocks~\cite{SD3} that fuse text and image semantics while leaving the pixel-level pathway unchanged. In each MM-DiT block, image and text tokens form two streams with separate QKV projections.

Text embeddings $y\in\mathbb{R}^{B\times L_{\text{txt}}\times D_{\text{txt}}}$ are produced by a frozen Gemma-2 encoder~\cite{gemma2}. Following \cite{xie2025sana}, we prepend a concise system prompt to the user prompt before feeding the sequence to the text encoder. The resulting token embeddings are projected to the model width and used as the text stream in MM-DiT.

Empirically, we find the semantic tokens from the patch-level pathway are sufficient to convey textual intent to the pixel updates. The pixel-level pathway is therefore architecturally identical to the class-conditioned model: it operates on pixel tokens and is conditioned only through the semantic tokens together with the timestep. No text tokens are routed directly to the pixel stream.

\begin{table}[t]
    \centering
    \resizebox{\linewidth}{!}{%
    \begin{tabular}{lccccc}
    \toprule
    \multirow{2}{*}{\textbf{Method}} & \multicolumn{5}{c}{\textbf{Generation@512}} \\
    \cmidrule(lr){2-6}
     & \textbf{gFID}$\downarrow$ & \textbf{sFID}$\downarrow$ & \textbf{IS}$\uparrow$ & \textbf{Precision}$\uparrow$ & \textbf{Recall}$\uparrow$ \\
    \midrule
    \multicolumn{6}{c}{\textbf{Latent Generative Models}} \\
    \midrule
    DiT-XL~\cite{dit}        & 3.04 & 5.02 & 240.8 & 0.84 & 0.54 \\
    SiT-XL~\cite{sit}        & 2.62 & 4.18 & 252.2 & 0.84 & 0.57 \\
    MaskDiT~\cite{maskdit} & 2.50 & 5.10 & 256.2 & 0.83 & 0.56 \\
    U-ViT-H~\cite{uvit} & 4.05 & - & 263.8 & 0.84 & 0.48 \\
    REPA~\cite{repa} & 2.08 & 4.19 & 274.6 & 0.83 & 0.58 \\
    RAE-XL~\cite{zheng2025rae} & 1.13 & - & 259.6 & 0.80 & 0.63 \\
    \midrule
    \multicolumn{6}{c}{\textbf{Pixel Generative Models}} \\
    \midrule
    ADM~\cite{adm} & 3.85 & 5.86 & 221.7 & 0.84 & 0.53 \\
    RIN \cite{rin} & 3.95 & - & 216.0 & - & - \\
    VDM++ \cite{vdm++} & 2.65 & - & 278.1 & - & - \\
    PixNerd-XL~\cite{pixnerd} & 2.84 & 5.95 & 245.6 & 0.80 & 0.59 \\
    EPG-L/32~\cite{epg} & 2.35 & - & 295.4 & 0.82 & 0.57 \\
    JiT-H~\cite{jit} & 1.94 & - & 309.1 & - & - \\
    \midrule
    \model{} & \textbf{1.81} & \textbf{5.61} & 278.6 & 0.78 & \textbf{0.67} \\
    \bottomrule
    \end{tabular}%
    }
    \caption{\textbf{Quantitative comparison} on ImageNet 512$\times$512.}
    \label{tab:imagenet512_main}
    \vspace{-8pt}
\end{table}

\subsection{Training Objectives}
We adopt the Rectified Flow formulation~\cite{rf} in pixel space and train the model with its velocity-matching loss:
\begin{equation}
\mathcal{L}_{\text{diff}} = \mathbb{E}_{t,x,\varepsilon}\big[\lVert f_\theta(x_t,t,y) - v_t \rVert_2^2\big].
\end{equation}

Following~\cite{repa}, we include an alignment objective that encourages mid-level patch-pathway tokens to agree with features from a frozen DINOv2 encoder~\cite{oquab2023dinov2}. The overall objective is $\mathcal{L} = \mathcal{L}_{\text{diff}} + \lambda_{\text{repa}}\,\mathcal{L}_{\text{repa}}$. We use the same formulation for class- and text-conditional models.

\section{Experiments}

We evaluate the effectiveness of \model{} through extensive experiments. To demonstrate the scalability of our approach, we instantiate \model{} with three model sizes: Base (B), Large (L), and Extra Large (XL) for experiments on ImageNet. The detailed configurations are summarized in Table~\ref{tab:model_config}. Unless otherwise specified, we use \model{}-XL as the default model for all experiments on ImageNet.

\begin{table}[htbp]
    \centering
    \resizebox{\linewidth}{!}{
    \begin{tabular}{l c c c c c c}
        \toprule
        Config & $N$ & $M$ & $D$ & $D_{\text{pix}}$ & Heads & Params (M) \\
        \midrule
        \model{}-B & 12 & 2 & 768 & 16 & 12 & 184 \\
        \model{}-L & 22 & 4 & 1024 & 16 & 16 & 569 \\
        \model{}-XL & 26 & 4 & 1152 & 16 & 16 & 797 \\
        \bottomrule
    \end{tabular}}
    \caption{Model configurations for \model{} variants for experiments on ImageNet 256$\times$256. $N$ and $M$ denote the depth of the patch-level and pixel-level pathways, respectively. $D$ and $D_{\text{pix}}$ represent the hidden dimension of the patch-level and pixel-level pathways.}
    \label{tab:model_config}
\end{table}

\subsection{Implementation Details}\label{sec:implementation_details}

\noindent\textbf{Class-conditioned Image Generation.} We follow the training setup of~\cite{dit} and train on ImageNet-1K~\cite{imagenet}. Our model uses hidden dimension $1152$, patch-level branch depth $N{=}26$, and pixel-level branch depth $M{=}4$. We resize images to $512\times512$ using the utility from~\cite{edm2} and adopt the ADM~\cite{adm} preprocessing pipeline. For the representation alignment objective, we set $\lambda_{\text{repa}}{=}0.5$ and apply alignment at the eighth block of the patch-level pathway. Throughout diffusion training, we use logit-normal sampling~\cite{SD3}, EMA with decay $0.9999$, AdamW with betas $(0.9, 0.999)$, a batch size of $256$, and \textit{bfloat16} mixed precision. On ImageNet 256$\times$256, we train with a constant learning rate of $1\times10^{-4}$ for the first 160 epochs, then step down to $1\times10^{-5}$ for the remainder. We apply gradient clipping at $1.0$ initially and $0.5$ thereafter for stability. For ImageNet 512$\times$512, we fine-tune for 530 epochs from a 320-epoch ImageNet 256$\times$256 checkpoint using the same $1\times10^{-5}$ learning rate and $0.5$ clipping threshold.

\noindent\textbf{Text-to-Image Generation.} We use Gemma-2~\cite{gemma2} as the text encoder and incorporate the MM-DiT conditioning design~\cite{SD3} on the patch-level pathway. Our \model{}-T2I uses hidden size $1536$, patch-level pathway depth $N{=}14$, and pixel-level pathway depth $M{=}2$. We collect approximately 26M image–text pairs at $1024^2$ resolution with various aspect ratios to train our model. We first pre-train the model from scratch at $512\times512$ resolution for 400K iterations with AdamW (learning rate $1\times10^{-4}$; betas $(0.9, 0.999)$), batch size $1{,}024$, gradient clipping $0.5$, and the shifting strategy~\cite{SD3} with shift value $\alpha{=}3.0$. We then finetune the model at $1024^2$ resolution for another 100K iterations with AdamW (learning rate $2\times10^{-5}$), batch size $768$, a higher shift value $\alpha{=}4.0$, and gradient clipping $0.1$.

\noindent\textbf{Evaluation Settings on ImageNet.} Following ADM~\cite{adm}, we report FID (gFID), sFID, Inception Score, and Precision–Recall on 50K samples. On ImageNet 256$\times$256, we use a guidance scale $3.25$ with an interval~\cite{CFGinterval} $[0.1, 1.0]$ for the 80-epoch checkpoint. All other evaluations use a guidance $2.75$ with an interval $[0.1, 0.9]$, unless otherwise stated. On ImageNet 512$\times$512, the guidance scale is $3.5$ and the interval is $[0.1, 1.0]$. 

\noindent\textbf{Evaluation Settings for Text-to-Image.} We evaluate \model{}-T2I at $512\times512$ and $1024^2$ resolutions on GenEval~\cite{ghosh2024geneval} and DPG-Bench~\cite{hu2024ella} using 533 and 1{,}065 prompts, respectively. Unless noted, we use a fixed guidance scale $4.5$ and the shift value aligns with the training setting at corresponding resolutions.

\noindent\textbf{Sampling at Inference.} For both class-conditioned and text-conditional image generation tasks, we use FlowDPMSolver~\cite{xie2025sana}, a modified DPMSolver++~\cite{lu2022dpm_add} in the Rectified Flow formulation, with \textit{bfloat16} precision. By default, we perform sampling with 100 steps on ImageNet and 25 steps for text-to-image generation.

\begin{figure*}[t]
    \centering
    \begin{subfigure}{0.72\linewidth}
        \includegraphics[width=\linewidth]{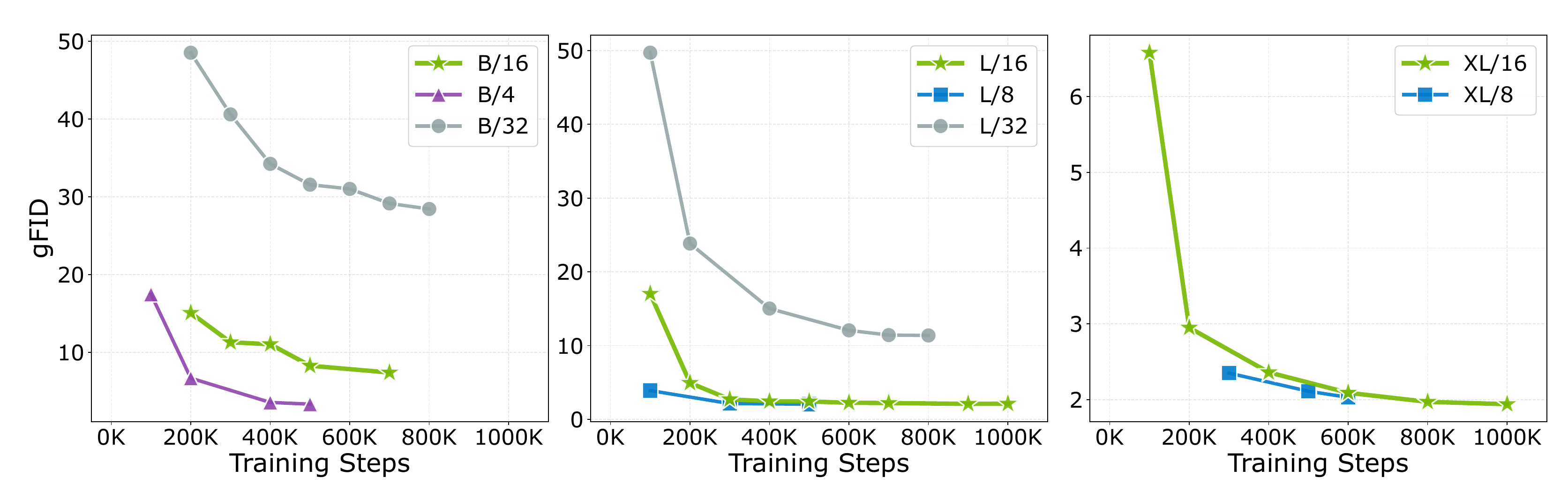}
        \caption{Patch-size ablations for B/L/XL models on ImageNet 256$\times$256.}
        \label{fig:app:convergence_full}
    \end{subfigure}
    \hfill
    \begin{subfigure}{0.26\linewidth}
        \includegraphics[width=0.965\linewidth]{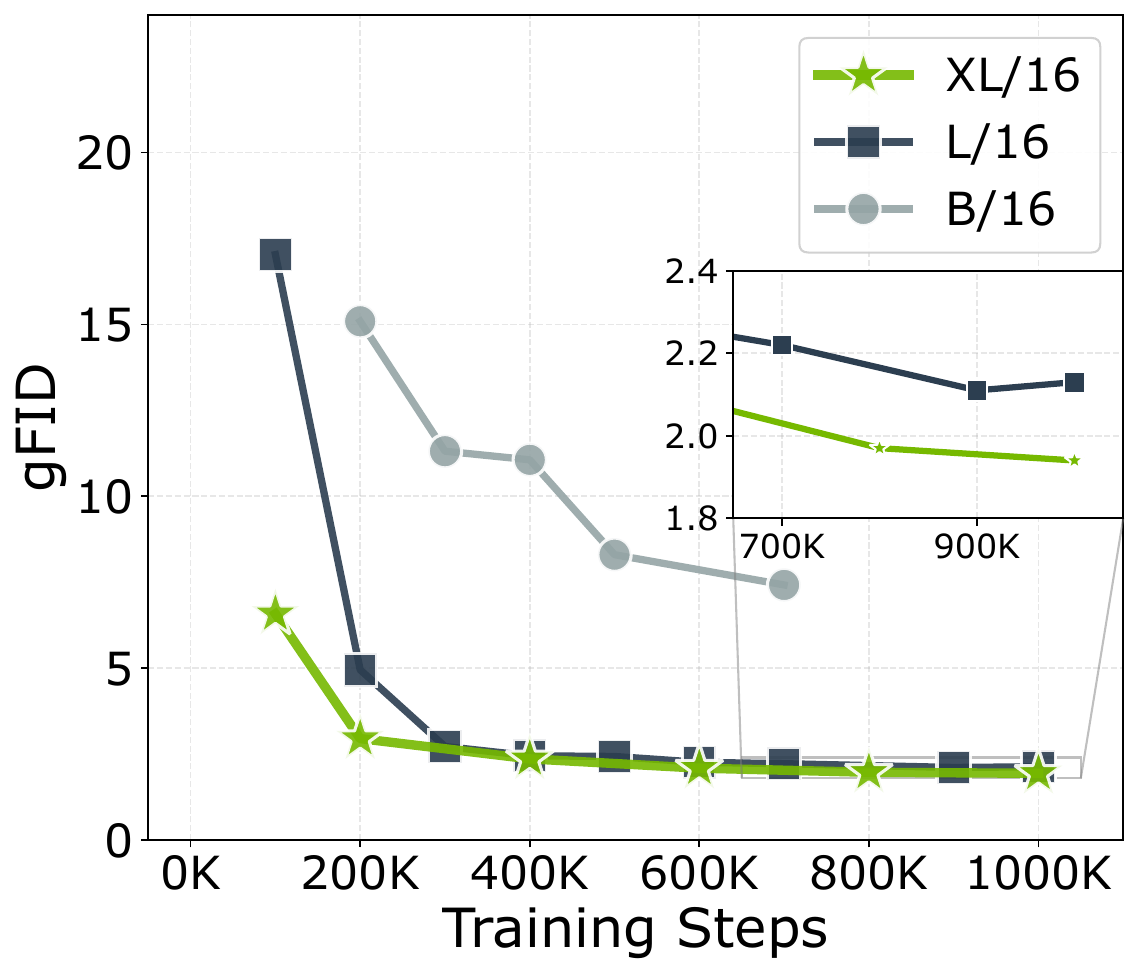}
        \caption{Comparison at fixed patch size.}
        \label{fig:app:comparison_16}
    \end{subfigure}
    \caption{Convergence analysis of \model{} on ImageNet 256$\times$256. (a) gFID vs.\ training iterations for B, L, and XL models with varying patch sizes. (b) Comparison of B/L/XL models at a fixed patch size $p{=}16$.}
    \label{fig:app:convergence_comparison}
\end{figure*}

\begin{table}[t]
    \centering
    \resizebox{\linewidth}{!}{
   \begin{tabular}{lcccc}
    \toprule
   \multirow{2}{*}{\textbf{Methods}} & \textbf{Params} & \multirow{2}{*}{\textbf{GenEval~$\uparrow$}} & \multirow{2}{*}{\textbf{DPG~$\uparrow$}} & \multirow{2}{*}{\textbf{\makecell{Throughput \\ (samples/s)~$\uparrow$}}} \\
   & \textbf{(B)} &  &  &  \\
   \toprule
   \multicolumn{5}{l}{\textit{512 $\times$ 512 resolution}} \\    \midrule
   PixArt-$\alpha$~\citep{chenpixart} & 0.6 & 0.48 & 71.6 & \textbf{1.5} \\
   PixArt-$\Sigma$~\citep{chen2024pixart} & 0.6 & 0.52 & 79.5 & \textbf{1.5} \\
   PixelFlow$\dagger$~\cite{pixelflow} & 0.9 & 0.60 & 77.9 & 0.05 \\
   PixNerd$\dagger$~\cite{pixnerd} & 1.2 & \underline{0.73} & \underline{80.9} & 1.04 \\
   \midrule
   \textbf{\model{}-T2I}$\dagger$ & 1.3 & \textbf{0.78} & \textbf{83.7} & \underline{1.07} \\
   \toprule
   \toprule
   \multicolumn{5}{l}{\textit{1024 $\times$ 1024 resolution}} \\    \midrule
   PixArt-$\Sigma$~\citep{chen2024pixart}  & 0.6   & 0.54              & 80.5              & 0.4 \\
   LUMINA-Next~\citep{zhuo2024lumina}      & 2.0   & 0.46              & 74.6              & 0.12 \\
   SDXL~\citep{podell2023sdxl}             & 2.6   & 0.55              & 74.7              & 0.15 \\
   Playground v2.5~\citep{li2024playground} & 2.6   & 0.56              & 75.5              & 0.21 \\
   Hunyuan-DiT~\citep{li2024hunyuan}       & 1.5   & 0.63              & 78.9              & 0.05 \\ 
   DALLE 3~\citep{Dalle-3}                 & -     & 0.67              & \underline{83.5}              & - \\
   FLUX-dev~\citep{flux}                   & 12.0  & 0.67              & 84.0              & 0.04 \\
   FLUX-schnell~\citep{flux}               & 12.0  & \underline{0.71}              & \textbf{84.8}            & \textbf{0.5} \\
    \midrule
    \textbf{\model{}-T2I}$\dagger$ & 1.3 & \textbf{0.74} & \underline{83.5} & \underline{0.33} \\
    \bottomrule
    \end{tabular}
    }
    \caption{\textbf{Comprehensive comparison of our method with text-to-image approaches.} We highlight the \textbf{best} and \underline{second-best} entries. $\dagger$ indicates pixel-space diffusion models.}
   \label{tab:t2i_overall}
\end{table}

\subsection{Class-conditioned Image Generation on ImageNet}

As shown in Table~\ref{tab:imagenet256_main}, at 320 epochs, our \model{}-XL obtains a gFID of \textbf{1.61}, surpassing recent pixel-space models including PixelFlow-XL (gFID 1.98), PixNerd-XL (gFID 1.93), and EPG-XXL/16 (gFID 1.81), showing a very significant improvement in image quality.  
Besides fidelity, \model{}-XL exhibits a stronger diversity–faithfulness trade-off, achieving a recall of \textbf{0.64} (vs. 0.60 for PixelFlow-XL), with competitive precision (\textbf{0.78}). 
Notably, \model{}-XL converges quickly: after only \textbf{80 epochs} it already reaches gFID \textbf{2.36} and IS \textbf{282.3}, outperforming classical pixel diffusion (ADM-U, gFID 4.59 at 400 epochs) and autoregressive baselines (JetFormer, gFID 6.64), indicating that \textit{structuring pixel modeling accelerates convergence}. Note that SiD2~\cite{sid2} reports a gFID of 1.38 at $256\times256$, but its model size, training epochs, IS, and Precision/Recall are unknown. For fairness, we include its gFID for reference but exclude it from Table~\ref{tab:imagenet256_main}.

\begin{figure*}[t]
    \centering
    \setlength{\tabcolsep}{2pt} 
    \begin{tabular}{cccccc}
        \includegraphics[width=0.1585\textwidth]{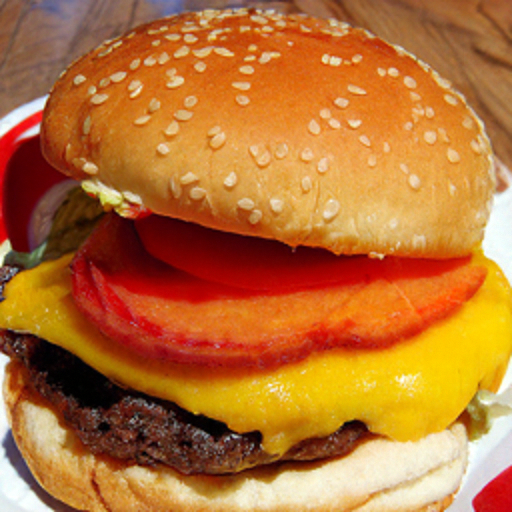} &
        \includegraphics[width=0.1585\textwidth]{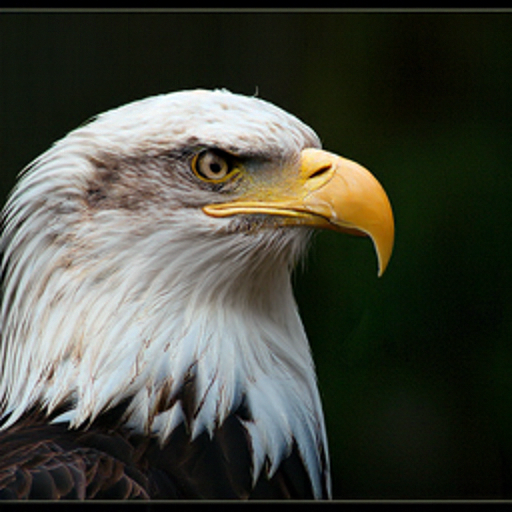} &
        \includegraphics[width=0.1585\textwidth]{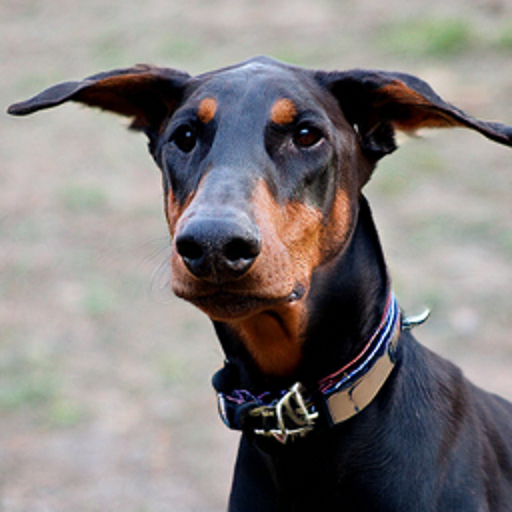} &
        \includegraphics[width=0.1585\textwidth]{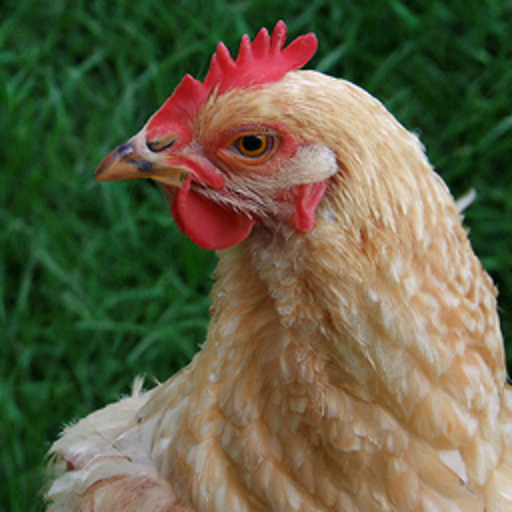} &
        \includegraphics[width=0.1585\textwidth]{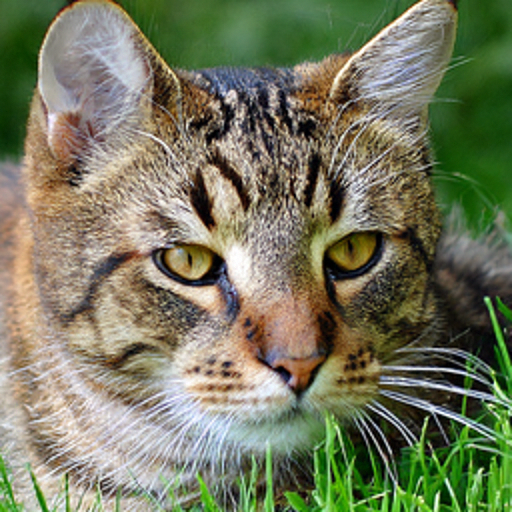} &
        \includegraphics[width=0.1585\textwidth]{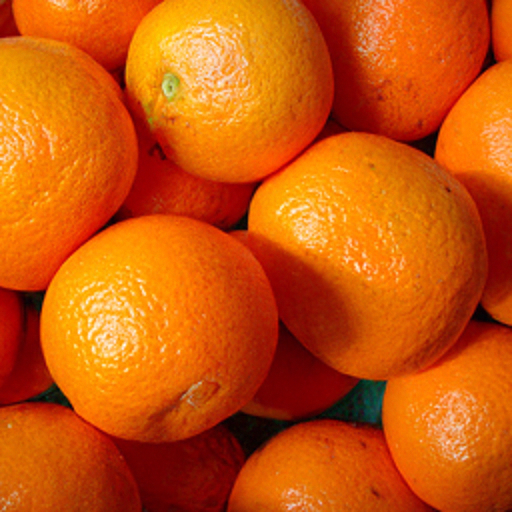} \\
        \includegraphics[width=0.1585\textwidth]{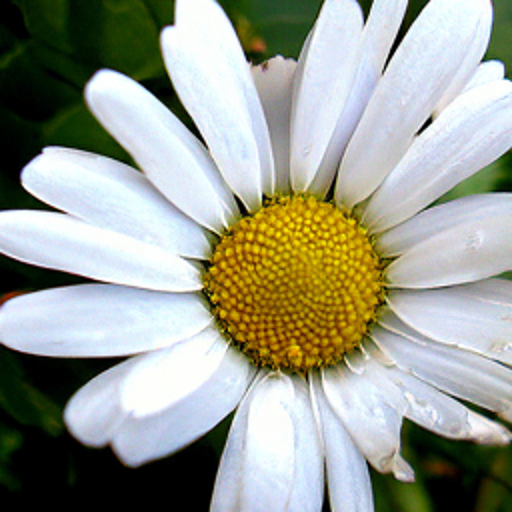} &
        \includegraphics[width=0.1585\textwidth]{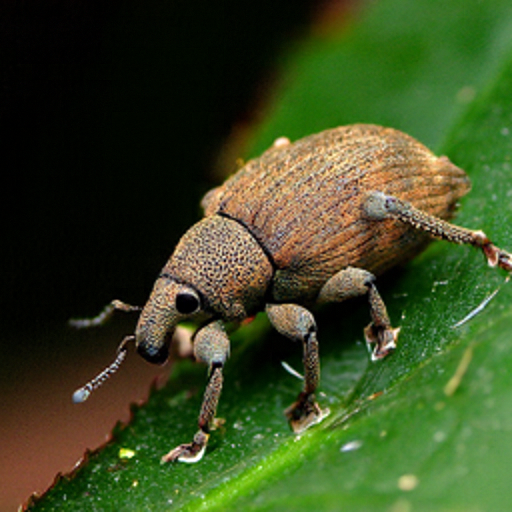} &
        \includegraphics[width=0.1585\textwidth]{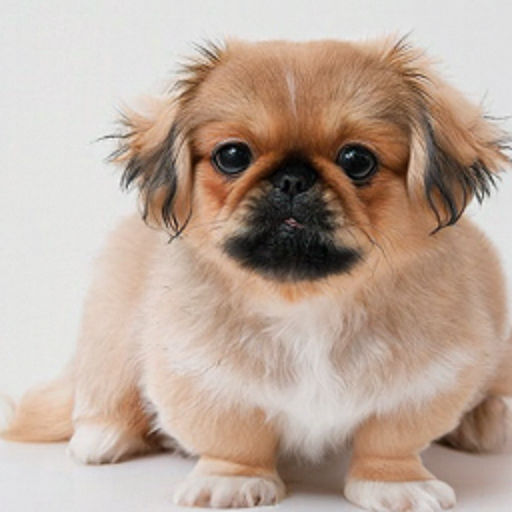} &
        \includegraphics[width=0.1585\textwidth]{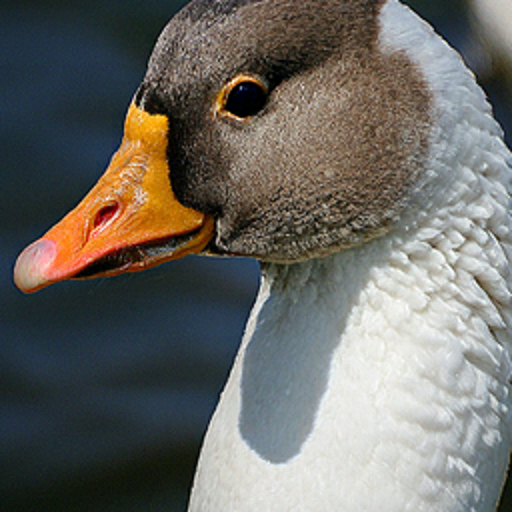} &
        \includegraphics[width=0.1585\textwidth]{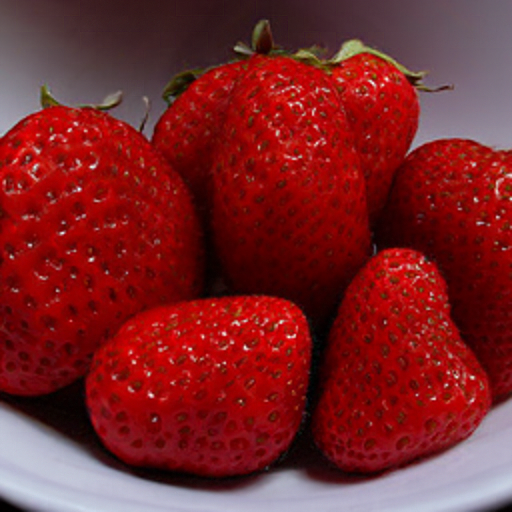} &
        \includegraphics[width=0.1585\textwidth]{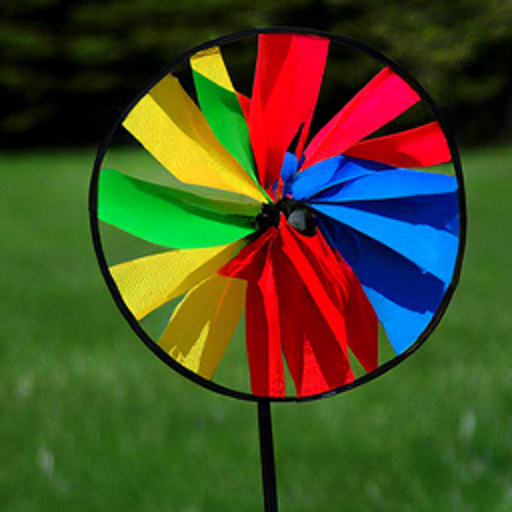} 
    \end{tabular}
    \caption{\textbf{Qualitative results} on ImageNet $512 \times 512$ using PixelDiT. We use a classifier-free guidance scale $\alpha_{\mathrm{cfg}} = 4.0$.}
    \label{fig:qual_imagenet512}
\end{figure*}

\noindent\textbf{ImageNet 512$\times$512.} Our
\model{} achieves a gFID of \textbf{1.81}, sFID \textbf{5.61}, IS \textbf{278.6}, precision \textbf{0.78}, and recall \textbf{0.67} as in Table~\ref{tab:imagenet512_main}, outperforming prior pixel-space models such as EPG-L/32 and PixNerd-XL and yielding the highest recall among pixel methods at this resolution.

Notably, \model{} even surpasses latent-space models such as REPA (gFID 2.08) \emph{without} any pretrained autoencoder, demonstrating the strong potential of pixel-space diffusion models in generating high-quality images. This is further supported by qualitative results in Figure~\ref{fig:qual_imagenet256} and Figure~\ref{fig:qual_imagenet512}. \model{} produces sharp textures, coherent object boundaries, and reasonable global structures, showing superiority of our dual-level architecture design.

\subsection{Text-to-Image Generation}

Table~\ref{tab:t2i_overall} presents the quantitative results of \model{}-T2I on text-to-image generation task. 
At $512\times512$ resolution, \model{}-T2I attains a GenEval score of 0.78 and DPG of 83.7, outperforming recent pixel-space models such as PixNerd and PixelFlow. At $1024^2$ resolution, \model{}-T2I achieves a GenEval score of 0.74 and DPG of 83.5, surpassing many latent diffusion models on GenEval and remaining competitive on DPG. These results indicate that end-to-end pixel-space diffusion with dual-level design scales to  text-to-image generation with strong text grounding and robust compositionality.
Qualitatively, Figure~\ref{fig:teaser}(a) shows representative samples produced by \model{}-T2I, illustrating high-resolution synthesis at the megapixel scale and consistent multi-resolution generation. More visualizations are provided in the appendix.

\noindent\textbf{Efficiency.} Throughput in Table~\ref{tab:t2i_overall} is measured with \textit{fp16} precision on a single NVIDIA A100 GPU. \model{}-T2I reaches 1.07 samples per second at $512\times512$ resolution, exceeding pixel-space baselines PixelFlow and PixNerd. At $1024^2$ resolution, our \model{}-T2I reaches 0.33 samples per second, comparable to latent diffusion models, despite denoising directly in pixel space.

\subsection{Detail Preservation in Image Editing}

In Figure~\ref{fig:teaser}(b), we perform image editing with  FlowEdit~\cite{kulikov2025flowedit} and compare the results using latent diffusion models (Stable Diffusion~3, FLUX) and PixelDiT.  
While both latent models generate motorcycle to replace the bicycle, the small scene text on the wall is heavily distorted. To investigate the cause of the distortion, Figure~\ref{fig:teaser}(b) presents the direct reconstruction results by the latent models' VAEs, which show that the scene text has already been distorted by the VAE encoder due to its lossy compression nature, and the error further accumulates in the latent diffusion process. In contrast, since PixelDiT is end-to-end trained in pixel space without relying on VAEs, it does not suffer from the lossy compression distortion. Therefore, it not only makes correct modifications, but also preserves the small scene text well, showing  significant advantages on preserving fine details for image editing. 

\begin{table}[t]
    \centering
    \resizebox{\linewidth}{!}{
    \begin{tabular}{lcr}
    \toprule
    \textbf{Model Components} & \textbf{Epoch} & \textbf{gFID}$\downarrow$ \\ 
    \midrule
    $\mathbb{A}$: Vanilla DiT/16 & 80 & 9.84 \\
    \midrule
    $\mathbb{A}$: + RoPE, RMSNorm & 80 & 8.53 \\
    $\mathbb{B}$: + Dual-level, patch-wise AdaLN (no compaction) & 80 & OOM \\
    $\mathbb{B}$: + Pixel Token Compaction & 80 & 3.50 \\ 
    \multirow{2}{*}{$\mathbb{C}$: + Pixel-wise AdaLN} & 80 & 2.36 \\
    & 320 & 1.61 \\
    \bottomrule
    \end{tabular}}
    \caption{\textbf{Ablations of \model{}-XL} on ImageNet 256$\times$256. Results start from Vanilla DiT and incrementally add architectural improvements and inference strategies. OOM indicates the dual-level variant without token compaction exceeds memory limits. Labels $\mathbb{A}$–$\mathbb{C}$ match the design schematic in Figure~\ref{fig:pixeladaln}.}
    \label{tab:main_ablation}
\end{table}

\subsection{Ablation Study}
\subsubsection{Contribution of Core Components}
Table~\ref{tab:main_ablation} quantifies the contribution of the proposed pixel-modeling components by comparing different model variants. Note that labels $\mathbb{A}$–$\mathbb{C}$ in Table~\ref{tab:main_ablation} correspond to the schematic variants in Figure~\ref{fig:pixeladaln}. Specifically,  we use a 30-layer,  $16\times16$ patchified DiT that directly performs denoising in pixel space as the baseline model (``Vanilla DiT/16''). This baseline operates solely on patch tokens without a dedicated pixel-level pathway,  treating each $16\times16$ patch as a high-dimensional vector. It obtains 9.84 gFID at 80 epochs. Introducing a dual-level architecture without pixel token compaction causes global attention to scale quadratically with the number of pixels and leads to an out-of-memory condition (OOM). 
Adding pixel token compaction resolves this bottleneck by shortening the global-attention sequence from $H{\times}W$ pixels to $L{=}(\frac{H}{p})(\frac{W}{p})$ patches, yielding a significant quality improvement to 3.50 gFID at the same 80-epoch budget. Incorporating pixel-wise AdaLN further aligns per-pixel updates with the semantic context produced by the patch-level pathway, improving gFID to 2.36 at 80 epochs and to 1.61 at 320 epochs. 

The comparison between model variants $\mathbb{A}$, $\mathbb{B}$, and $\mathbb{C}$ demonstrates the importance of each proposed component. More importantly, the comparison between our full PixelDiT model $\mathbb{C}$ and the vanilla DiT/16 $\mathbb{A}$ reveals that \textit{pixel-level token modeling plays a key role in pixel generative models}. Without pixel modeling, \textit{i.e.}, the visual content is only learned at the patch level, it will be challenging for the model to learn the fine details, and the visual quality will degrade significantly.

\begin{table}[t]
    \centering
    \resizebox{\linewidth}{!}{
    \begin{tabular}{l r cc cc}
        \toprule
        \multirow{2}{*}{\textbf{Configuration}} & \multirow{2}{*}{\textbf{GFLOPs}} & \multicolumn{2}{c}{\textbf{Epoch 80}} & \multicolumn{2}{c}{\textbf{Epoch 160}} \\
        & & \textbf{FID}$\downarrow$ & \textbf{IS}$\uparrow$ & \textbf{FID}$\downarrow$ & \textbf{IS}$\uparrow$ \\
        \midrule
        \model{}-XL & 311 & \textbf{2.36} & \textbf{282.3} & \textbf{1.97} & \textbf{299.4} \\
        No Pixel Token Compaction & 82247 & \multicolumn{4}{c}{OOM} \\
        No Pixel-Pathway Attention & \textbf{279} & 2.56 & 256.9 & 2.22 & 281.7 \\
        \bottomrule
    \end{tabular}}
    \caption{\textbf{Computation and convergence analyses of pixel token compaction.} 
    We report GFLOPs alongside ImageNet 256$\times$256 metrics at 80 and 160 epochs. 
    \textit{No Pixel Token Compaction} removes the compress--expand pathway, resulting in out-of-memory (OOM) at our training scale. 
    \textit{No Pixel-Pathway Attention} ablates self-attention in all PiT blocks.}
    \label{tab:ptc_efficiency}
\end{table}

\subsubsection{Analysis on Pixel Token Compaction} 
Token compaction is essential for making pixel-space training feasible. A global attention over $N{=}H\times W$ pixel tokens incurs $O(N^{2})$ memory and $O(N^{2}D)$ FLOPs, producing billions of attention entries even at $256\times256$ resolution, as reflected by the 82{,}247 GFLOPs reported for this variant in Table~\ref{tab:ptc_efficiency}. Grouping pixels into $p\times p$ patches using pixel token compaction reduces the sequence length to $L{=}N/p^{2}$, yielding a $p^{4}$-fold reduction in attention cost.

To analyze the role of attention in the pixel-level pathway, we include a ``No Pixel-Pathway Attention'' ablation that removes the attention and only keeps pixel-wise AdaLN and an MLP at the pixel level. As shown in Table~\ref{tab:ptc_efficiency}, while this variant reduces GFLOPs, it is consistently inferior to our full PixelDiT model across different training iterations (e.g., from 80 to 160 epochs), with visible degradation in gFID and IS. This indicates that compact global attention is necessary to align local updates with global context.

\noindent\textbf{Summary of Ablation.} These ablations support our central claim: efficient pixel modeling via pixel token compaction and pixel-wise modulation makes end-to-end pixel-space diffusion practical and closes much of the gap to latent-space training while preserving fine details. Labels $\mathbb{A}$–$\mathbb{C}$ in Table~\ref{tab:main_ablation} correspond to the schematic variants in Figure~\ref{fig:pixeladaln}. Please refer to the Appendix for more ablation studies.

\subsubsection{Impact of Model Size and Patch Size}
We investigate the impact of the patch size $p$ on the performance of models at different scales: \model{}-B, \model{}-L, and \model{}-XL. 
For all evaluations, we use the same CFG guidance scale $3.25$ with interval $[0.10, 1.00]$.
We evaluate patch sizes of $4$, $8$, $16$, and $32$ on ImageNet 256$\times$256; Figure~\ref{fig:app:convergence_comparison}(a) visualizes the resulting convergence behavior.
For the base model, moving from $p{=}32$ to $p{=}16$ and $p{=}4$ substantially accelerates convergence: at 200K iterations gFID drops from $48.5$ (B/32) to $15.1$ (B/16) and $6.7$ (B/4), and B/4 ultimately reaches $3.4$ gFID at 500K iterations.
Larger models follow a similar trend, but the benefit of very small patches diminishes with scale. 
For \model{}-L, using $p{=}8$ rather than $p{=}16$ improves gFID only modestly (from $2.72$ to $2.15$ at 300K iterations), and for \model{}-XL the gap between $p{=}8$ and $p{=}16$ essentially vanishes---both configurations converge to gFID near $2.0$.
These results highlight a clear trade-off: smaller patches yield better image quality or faster convergence but incur a quadratic cost in sequence length, and the relative gain shrinks as the model capacity increases.
In practice, we therefore use $p{=}16$ as the default patch size for \model{}-XL, which offers near-optimal quality at substantially lower compute.

To analyze the standalone effect of model size, Figure~\ref{fig:app:convergence_comparison}(b) directly compares B, L, and XL variants at a fixed patch size $p{=}16$.
Scaling the model yields consistent gains across the entire training trajectory: at 200K iterations, gFID improves from $15.1$ (B/16) to $4.95$ (L/16) and $2.95$ (XL/16), and at 1M iterations XL/16 reaches $1.94$ gFID compared to roughly $2.1$ for L/16.
Thus, increasing capacity improves both image quality and the speed at which a given quality level is reached, demonstrating the scalability of \model{}.

 We leave more ablation studies to the Appendix. 


\section{Conclusion}
Single-stage pixel-space diffusion offers a simpler, more elegant, and domain-agnostic paradigm for visual synthesis than the two-stage latent-space approach. However, existing pixel-space methods still show a substantial quality gap from latent diffusion models. This gap likely stems from two main challenges: 1) the inherently greater complexity of pixel-space data and noise distributions than the latent-space ones, and 2) the lack of a mature training recipe, including effective architecture, objective, noise scheduling, and optimization design.

In this work, we show that this gap can be largely closed. With appropriate architectural design, particularly for pixel token modeling, pixel-space diffusion transformers can achieve image quality on par with latent diffusion on both class-conditional and text-conditional generation tasks. Specifically, PixelDiT factorizes pixel modeling into a dual-level transformer and introduces pixel-wise AdaLN together with pixel token compaction, enabling efficient disentanglement of semantics modeling and fine-grained detail refinement. \textit{These findings highlight effective and efficient pixel modeling as the key to practical pixel-space diffusion}. We hope that this perspective inspires further research in pixel-space generative modeling.

\appendix

{
  \small
  \bibliographystyle{unsrt}
  \bibliography{main}
}

\clearpage

\begin{table}[bp]
    \vspace{-5mm}
    \centering
    \resizebox{\linewidth}{!}{
    \begin{tabular}{lccc}
        \toprule
        & \textbf{\model{}-B} & \textbf{\model{}-L} & \textbf{\model{}-XL} \\
        \midrule
        \multicolumn{4}{l}{\textit{Architecture}} \\
        Input dim. & \multicolumn{3}{c}{$256 \times 256 \times 3$} \\
        Patch-level depth $N$ & 12 & 22 & 26 \\
        Pixel-level depth $M$ & 2 & 4 & 4 \\
        Hidden size $D$ & 768 & 1024 & 1152 \\
        Heads & 12 & 16 & 16 \\
        Pixel hidden size $D_{\textrm{pix}}$ & 16 & 16 & 16 \\
        Patch size $p$ & 16 & 16 & 16 \\
        \#Params (M) & 184 & 569 & 797 \\
        \midrule
        \multicolumn{4}{l}{\textit{Representation Alignment}~\cite{repa}} \\
        Alignment depth (patch-level) & \multicolumn{3}{c}{8-th layer} \\
        Loss weight $\lambda_{\text{repa}}$ & \multicolumn{3}{c}{0.5} \\
        Alignment encoder & \multicolumn{3}{c}{Frozen DINOv2~\cite{oquab2023dinov2}} \\
        \midrule
        \multicolumn{4}{l}{\textit{Optimization}} \\
        Training iteration & 800K & 1M & 1.6M \\
        Batch size & \multicolumn{3}{c}{256} \\
        Timestep reweighting & \multicolumn{3}{c}{Logit-normal~\cite{SD3}} \\
        Optimizer & \multicolumn{3}{c}{AdamW~\cite{loshchilov2017decoupled}, $\beta_1{=}0.9, \beta_2{=}0.999$} \\
        EMA decay & \multicolumn{3}{c}{0.9999} \\
        Class drop prob. & \multicolumn{3}{c}{0.1} \\
        Gradient clipping & 1.0 & 1.0 & 1.0 $\to$ 0.5 \\
        Learning rate & 1e-4 & 1e-4 & 1e-4 $\to$ 1e-5 \\
        Weight decay & 0 & 0 & 0 \\
        \midrule
        \multicolumn{4}{l}{\textit{Inference}} \\
        Sampler & \multicolumn{3}{c}{FlowDPMSolver~\cite{xie2025sana,lu2022dpm_add}} \\
        Sampling steps & \multicolumn{3}{c}{100} \\
        CFG scale (80-ep) & 3.25 & 3.25 & 3.25 \\
        CFG interval (80-ep) & \multicolumn{3}{c}{$[0.10,1.00]$} \\
        CFG scale (320-ep) & -- & -- & 2.75 \\
        CFG interval (320-ep) & -- & -- & $[0.10,0.90]$ \\
        Guidance for Figs.~\ref{fig:app:convergence_comparison},~\ref{fig:app:dpm_steps}--\ref{fig:app:ptc_curves} & \multicolumn{3}{c}{$3.25$, $[0.10,1.00]$ for all checkpoints}  \\
        \bottomrule
    \end{tabular}
    }
    \caption{Detailed architecture and training configurations for PixelDiT B/L/XL models on ImageNet-256.}
    \label{tab:app:model_summary_imagenet}
    \vspace{-5mm}
\end{table}

\begin{table}[bp]
    \centering
    \resizebox{\linewidth}{!}{
    \begin{tabular}{lcc}
        \toprule
        \textbf{Hyperparameter} & \multicolumn{2}{c}{\textbf{\model{}-T2I}} \\
        \midrule
        \multicolumn{3}{l}{\textit{Architecture}} \\
        Input dim. & $512^2 \times 3$ & $1024^2 \times 3$ \\
        Patch-level depth $N$ & \multicolumn{2}{c}{14} \\
        Pixel-level depth $M$ & \multicolumn{2}{c}{2} \\
        Hidden size $D$ & \multicolumn{2}{c}{1536} \\
        Heads & \multicolumn{2}{c}{24} \\
        Pixel hidden size $D_{\textrm{pix}}$ & \multicolumn{2}{c}{16} \\
        Patch size $p$ & \multicolumn{2}{c}{16} \\
        \#Params (M) & \multicolumn{2}{c}{1311} \\
        \midrule
        \multicolumn{3}{l}{\textit{Representation Alignment}~\cite{repa}} \\
        Alignment depth (patch-level) & 6-th layer & -- \\
        Loss weight $\lambda_{\text{repa}}$ & 0.5 & 0.0 (disabled) \\
        Alignment encoder & \multicolumn{2}{c}{Frozen DINOv2~\cite{oquab2023dinov2}} \\
        \midrule
        \multicolumn{3}{l}{\textit{Optimization}} \\
        Training iteration & 400K & 100K \\
        Batch size & 1024 & 768 \\
        Learning rate & 1e-4 & 2e-5 \\
        Gradient clipping & 0.5 & 0.1 \\
        Text drop prob. & \multicolumn{2}{c}{0.1} \\
        \midrule
        \multicolumn{3}{l}{\textit{Inference}} \\
        Sampler & \multicolumn{2}{c}{FlowDPMSolver~\cite{xie2025sana,lu2022dpm_add}} \\
        Sampling steps & \multicolumn{2}{c}{25} \\
        CFG scale & \multicolumn{2}{c}{4.5} \\
        \bottomrule
    \end{tabular}
    }
    \caption{Implementation details for PixelDiT-T2I model.}
    \label{tab:app:model_summary_t2i}
    \vspace{-2mm}
\end{table}

\begin{figure*}[t]
    \centering
    \includegraphics[width=\linewidth]{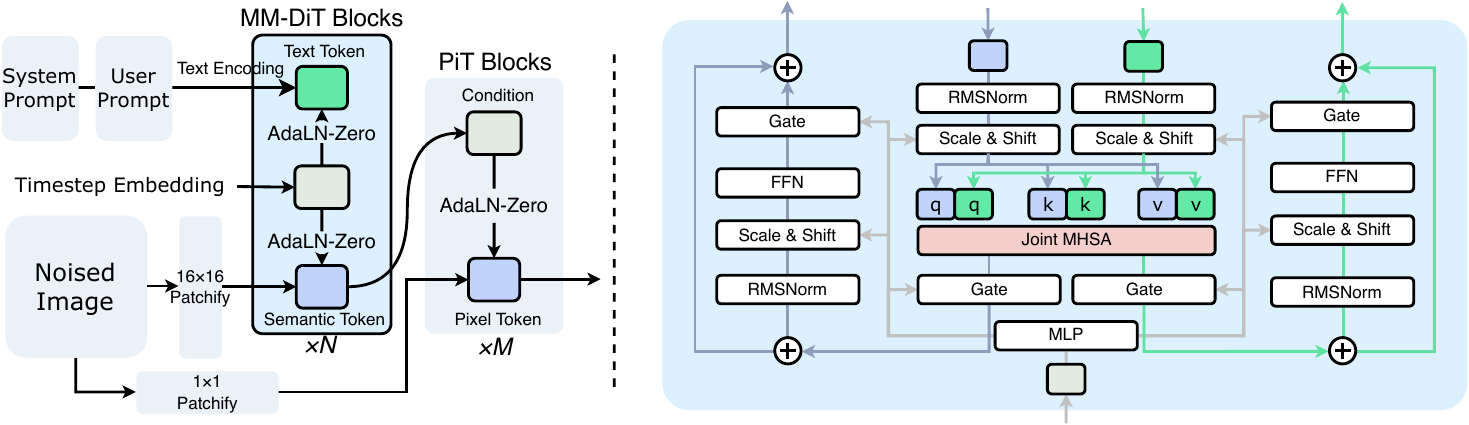}
    \caption{T2I architecture of \model{} with MM-DiT blocks on the patch-level pathway. The pixel-level pathway performs dense per-pixel modeling conditioned on semantic tokens.}
    \label{fig:app:t2i_schematic}
\end{figure*}

\section{Architecture and System Details}

\begin{figure*}[t]
    \centering
    \includegraphics[width=0.95\linewidth]{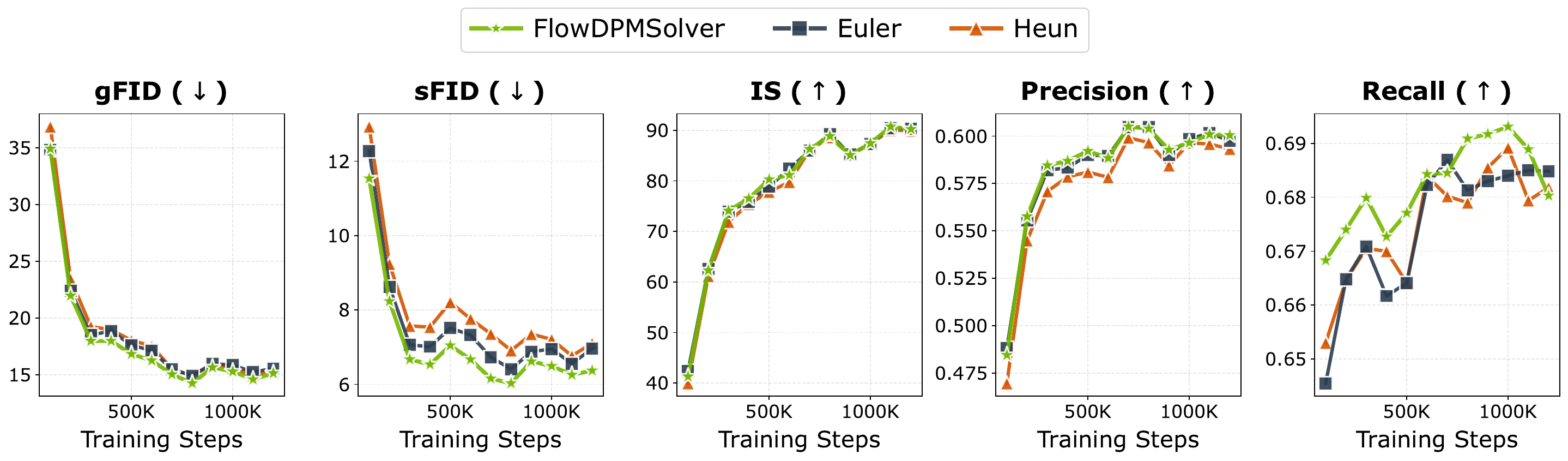}
    \caption{Comparison of FlowDPMSolver, Euler, and Heun samplers on ImageNet 256$\times$256 with 100 inference steps and no classifier-free guidance. FlowDPMSolver achieves the best combined trade-off between fidelity in gFID and sFID and diversity in IS, precision, and recall, which motivates its use as our default sampler.}
    \label{fig:app:solver_curves}
\end{figure*}

\begin{figure}[t]
    \centering
    \includegraphics[width=0.7\linewidth]{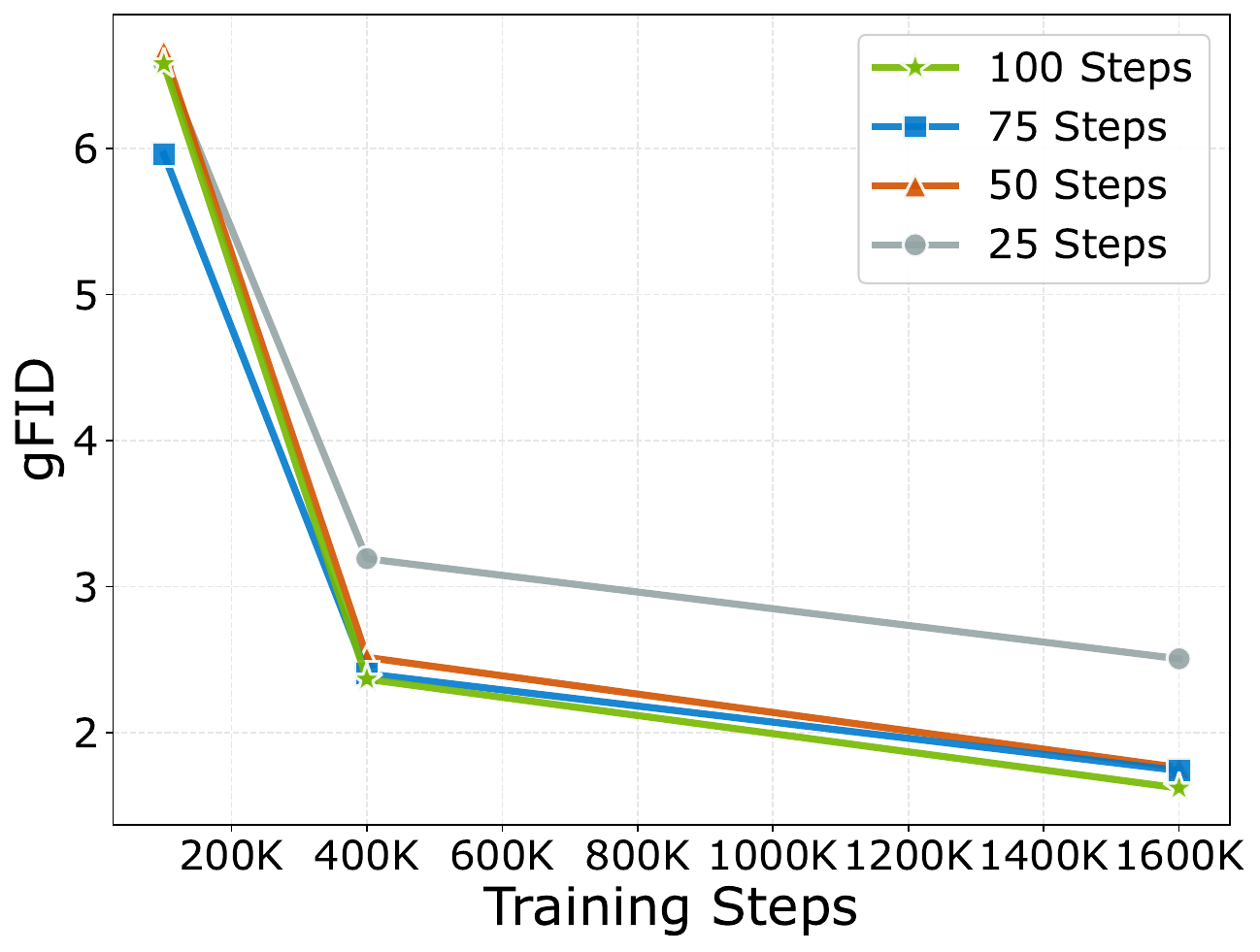}
    \vspace{-1mm}
    \caption{Effect of the number of FlowDPMSolver inference steps on gFID for \model{}-XL at different training stages on ImageNet 256$\times$256. Increasing the number of steps is most beneficial once the model is moderately or fully trained (400K and 1.6M iterations), with diminishing returns beyond 50 steps; we adopt 100 steps as the default for ImageNet experiments.}
    \label{fig:app:dpm_steps}
    \vspace{-2mm}
\end{figure}

\subsection{Summary of Model Size}
To study the impact of model size, we evaluate the base (B), large (L), and extra-large (XL) variants of \model{} on ImageNet 256$\times$256.
Tables~\ref{tab:app:model_summary_imagenet} and \ref{tab:app:model_summary_t2i} summarize the detailed architectural specifications and training settings for B, L, XL and T2I variants. \textbf{Note that the default configuration of all experiments in the main paper is PixelDiT-XL.} If not otherwise specified, we use the XL configuration for all ImageNet 256$\times$256 experiments in this appendix.

\subsection{Text-to-Image Architecture with MM-DiT}
Figure~\ref{fig:app:t2i_schematic} illustrates the T2I variant of \model{}, where the patch-level pathway is extended with MM-DiT blocks~\cite{SD3} to fuse text embeddings, while the pixel-level pathway remains unchanged. The figure emphasizes stream separation, conditioning flow, and the pixel-wise modulation interface used by the pixel-level pathway.

\section{Solvers and Guidance Scales}
\subsection{Ablation of Solvers}
We compare three diffusion samplers for denoising on ImageNet 256$\times$256: FlowDPMSolver~\cite{xie2025sana,lu2022dpm_add}, Euler, and Heun, all run for 100 steps without classifier-free guidance. 
Figure~\ref{fig:app:solver_curves} plots gFID, sFID, Inception Score (IS), precision, and recall as training progresses from 100K to 1,200K iterations.
Across most of the training trajectory, FlowDPMSolver achieves lower or comparable gFID and sFID than Euler and Heun, with the gap particularly pronounced in the low- and mid-epoch regimes (up to roughly 1--2 gFID points around $400$K--$800$K iterations).
FlowDPMSolver maintains the best overall trade-off: it matches or exceeds the competing solvers on sFID and IS while keeping precision and recall high.
These results motivate our choice of FlowDPMSolver as the default sampler for all main ImageNet and text-to-image evaluations.

\begin{figure}[t]
    \centering
    \includegraphics[width=0.9\linewidth]{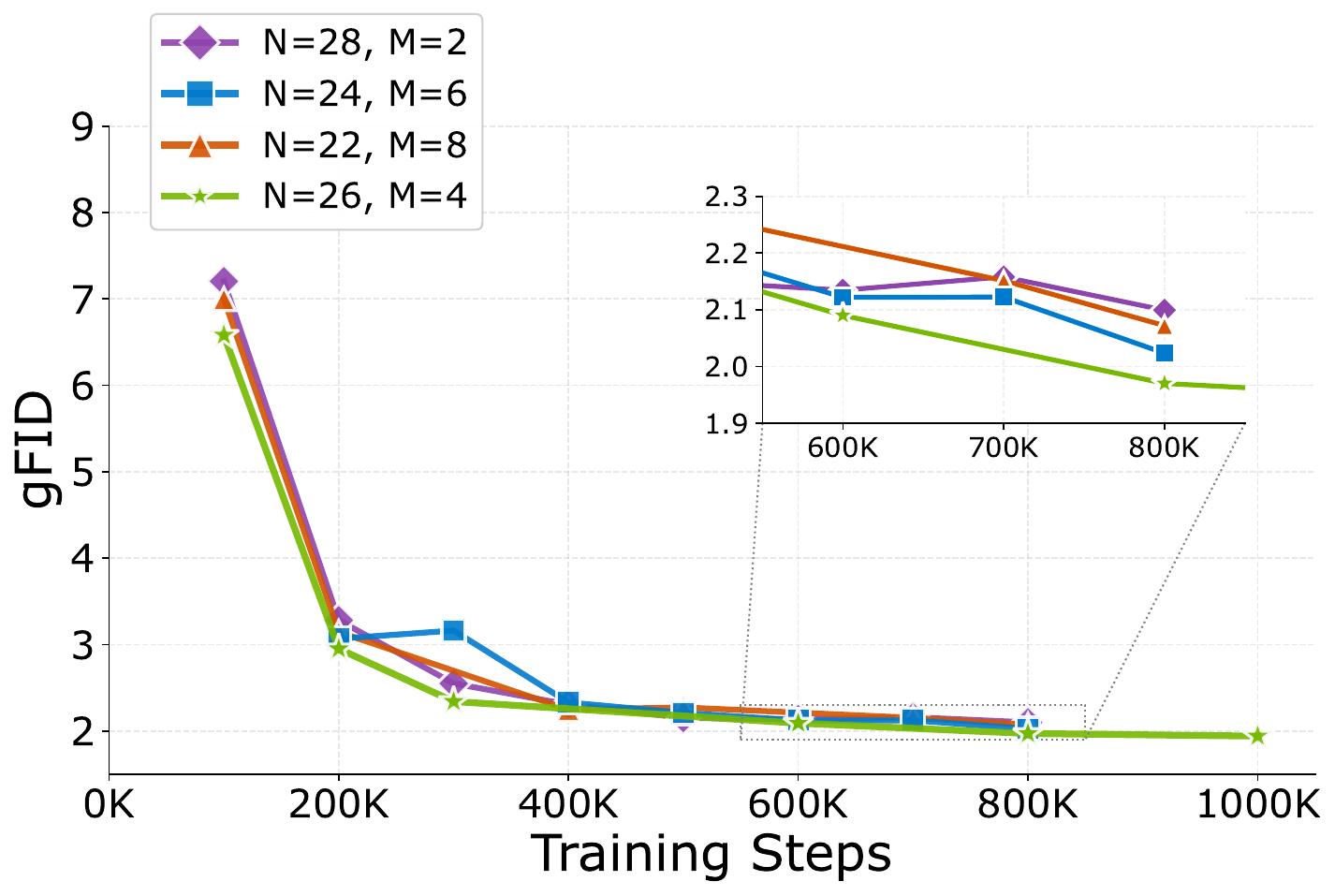}
    \caption{Ablation of depth allocation between patch-level ($N$) and pixel-level ($M$) pathways on ImageNet 256$\times$256. Our chosen configuration ($N{=}26, M{=}4$), highlighted in brown, offers the best trade-off between early convergence and final image quality.}
    \label{fig:app:nm_curves}
\end{figure}

\subsection{Inference Steps}
We further analyze the impact of the number of inference steps when using FlowDPMSolver.
Figure~\ref{fig:app:dpm_steps} shows gFID for \model{}-XL at three training stages (100K, 400K, and 1.6M iterations) as we vary the sampling budget from 25 to 100 steps.
At 100K iterations the model is undertrained and additional steps give only modest improvements, with gFID remaining in the 6--7 range.
Once the model has learned reasonable global structure (400K iterations), increasing the budget from 25 to 50 steps reduces gFID from about $3.19$ to $2.51$, and 100 steps further improves it to $2.36$.
For the fully converged checkpoint at 1.6M iterations, 25 steps already achieve gFID $\approx 2.50$, but 50--75 steps lower it to around $1.76$--$1.74$, and 100 steps obtain the best score of approximately $1.61$ gFID.
Overall, more inference steps consistently benefit well-trained models, though the marginal gain beyond 50 steps becomes small.
In practice we therefore use 100 steps for class-conditioned ImageNet experiments to match the strongest quality, and 25 steps for text-to-image generation where sampling latency is more critical.

\begin{figure}[t]
    \centering
    \includegraphics[width=0.9\linewidth]{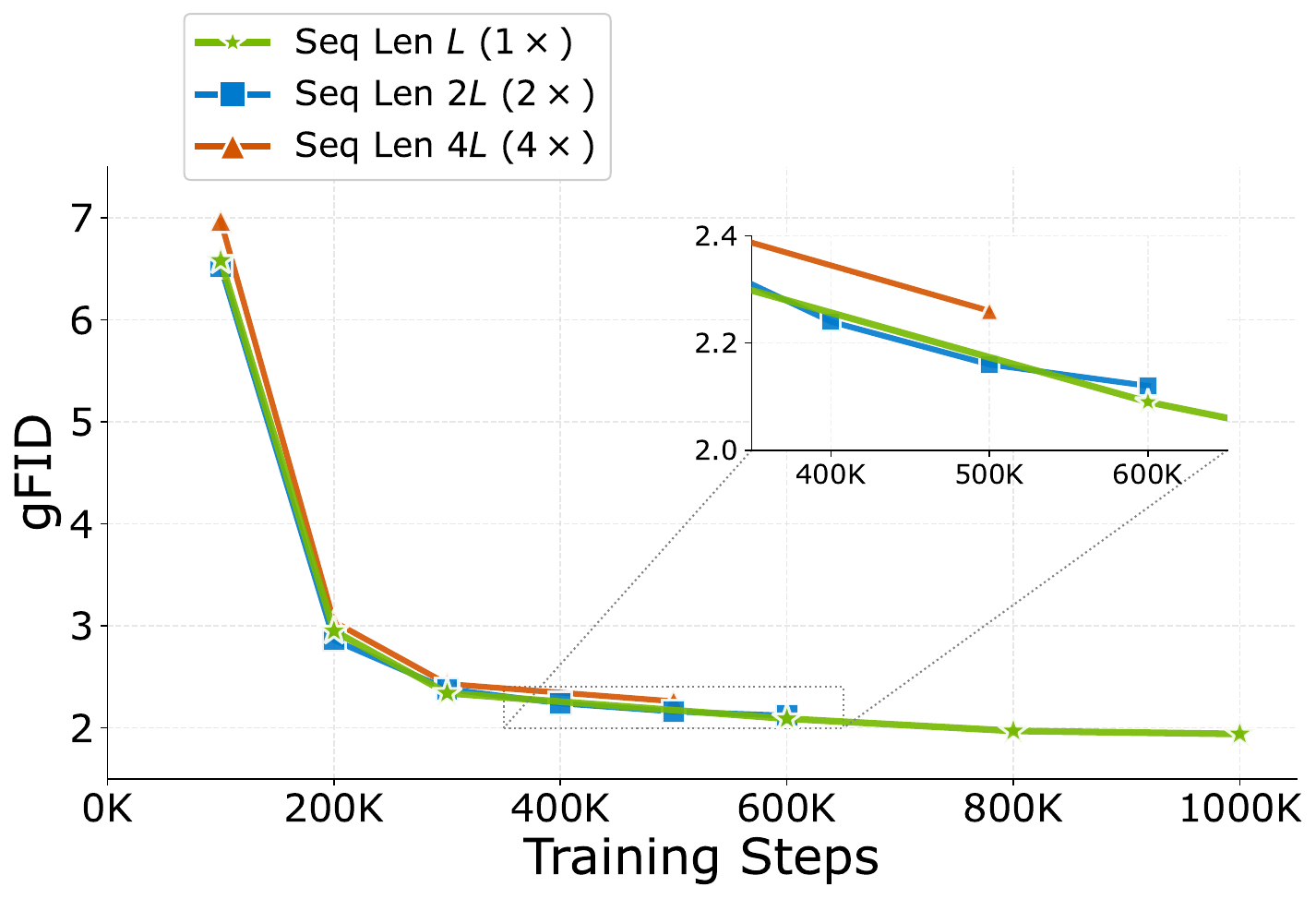}
    \caption{Ablation of Pixel Token Compaction rates on ImageNet 256$\times$256. The curves compare three post-compaction sequence lengths, denoted as ``Seq Len $L$ ($1\times$)'', ``Seq Len $2L$ ($2\times$)'', and ``Seq Len $4L$ ($4\times$)'', where $L$ is the number of patch tokens after compaction. ``Seq Len $L$ ($1\times$)'' corresponds to our default configuration in which each $p^2$ pixel block is compacted into a single token.}
    \label{fig:app:ptc_curves}
\end{figure}

\subsection{Guidance Scale and Interval}
We report the classifier-free guidance (CFG) settings used for \model{}-XL at 80 and 320 epochs on ImageNet 256$\times$256.
Table~\ref{tab:app:cfg_sweep} lists the CFG scale, active time interval, and the resulting gFID, sFID, IS, precision, and recall.
For the 80-epoch checkpoint, the best gFID is 2.36, obtained with a relatively strong guidance scale of 3.25 applied over the entire denoising trajectory from $t{=}0.10$ to $t{=}1.00$.
Increasing the scale to 3.50 or decreasing it to 3.00 slightly worsens gFID while mainly trading off IS and recall, and restricting the active interval to $[0.10,0.95]$ or $[0.10,0.90]$ does not lead to better performance.
For the 320-epoch checkpoint, the optimum shifts toward milder guidance: a scale of 2.75 active on the interval $[0.10,0.90]$ achieves the best gFID of 1.61 together with strong recall of 0.64, whereas both larger and smaller scales yield at most marginal IS gains at the cost of higher gFID.
In the main paper we therefore adopt a guidance scale of 3.25 with interval $[0.10,1.00]$ for the 80-epoch ImageNet 256$\times$256 results, and a scale of 2.75 with interval $[0.10,0.90]$ for the 320-epoch checkpoint that underpins our best reported scores.

\begin{table*}[t]
    \centering
    \resizebox{0.75\linewidth}{!}{
    \begin{tabular}{lccccccccc}
        \toprule
        Model & Epochs & Training Steps & CFG & Interval & gFID$\downarrow$ & sFID$\downarrow$ & IS$\uparrow$ & Prec.$\uparrow$ & Rec.$\uparrow$ \\
        \midrule
        \model{}-XL & 80  & 400K  & 3.25 & [0.10, 1.00] & 2.36 & 5.11 & 282.3 & 0.80 & 0.57 \\
        \model{}-XL & 80  & 400K  & 3.50 & [0.10, 1.00] & 2.60 & 5.07 & 305.2 & 0.82 & 0.57 \\
        \model{}-XL & 80  & 400K  & 3.00 & [0.10, 1.00] & 2.60 & 5.06 & 277.6 & 0.80 & 0.58 \\
        \model{}-XL & 80  & 400K  & 2.75 & [0.10, 1.00] & 2.76 & 5.17 & 259.2 & 0.79 & 0.59 \\
        \model{}-XL & 80  & 400K  & 3.25 & [0.10, 0.95] & 2.73 & 5.24 & 285.8 & 0.80 & 0.58 \\
        \model{}-XL & 80  & 400K  & 3.25 & [0.10, 0.90] & 2.75 & 5.32 & 285.4 & 0.80 & 0.58 \\ \midrule
        \model{}-XL & 320 & 1600K & 2.75 & [0.10, 0.90] & 1.61 & 4.68 & 292.7 & 0.78 & 0.64 \\
        \model{}-XL & 320 & 1600K & 2.75 & [0.10, 0.95] & 1.65 & 4.64 & 293.8 & 0.77 & 0.64 \\
        \model{}-XL & 320 & 1600K & 2.75 & [0.10, 1.00] & 1.66 & 4.60 & 294.2 & 0.78 & 0.64 \\
        \model{}-XL & 320 & 1600K & 2.50 & [0.10, 0.95] & 1.69 & 4.68 & 276.9 & 0.77 & 0.65 \\
        \model{}-XL & 320 & 1600K & 2.50 & [0.10, 0.90] & 1.71 & 4.60 & 275.2 & 0.77 & 0.65 \\
        \model{}-XL & 320 & 1600K & 2.50 & [0.10, 1.00] & 1.71 & 4.62 & 277.9 & 0.77 & 0.65 \\
        \bottomrule
    \end{tabular}}
    \caption{CFG settings and results for \model{}-XL on ImageNet 256$\times$256.}
    \label{tab:app:cfg_sweep}
\end{table*}

\section{Model Architecture Design}

\begin{table*}[t]
    \footnotesize
    \centering
    \begin{tabular*}{\linewidth}{@{\extracolsep{\fill}}lccccccccc@{}}
    \toprule
    \multirow{2}{*}{\textbf{Model}} & \multirow{2}{*}{\textbf{Params}~(B)} & \multirow{2}{*}{\textbf{Overall}~$\uparrow$} & \multicolumn{2}{c}{Objects} & \multirow{2}{*}{Counting} & \multirow{2}{*}{Colors} & \multirow{2}{*}{Position} & \multirow{2}{*}{Color} \\
    \cmidrule(lr){4-5}
    &  &  & Single & Two &  & & & Attribution \\
    \midrule
    \multicolumn{8}{l}{\textbf{512 \boldmath$\times$\unboldmath 512 resolution}} \\
    \midrule
    PixArt-$\alpha$             & 0.6 & \textbf{0.48} & 0.98 & 0.50 & 0.44 & 0.80 & 0.08 & 0.07 \\
    PixArt-$\Sigma$             & 0.6 & \textbf{0.52} & 0.98 & 0.59 & 0.50 & 0.80 & 0.10 & 0.15 \\
    PixelFlow~\citep{pixelflow}   & 0.9 & \textbf{0.60} & - & - & - & - & - & - \\
    PixNerd~\citep{pixnerd}       & 1.2 & \textbf{0.73} & 0.97 & 0.86 & 0.44 & 0.83 & 0.71 & 0.53 \\
    \midrule
    \textbf{\model{}-T2I}     & 1.3 & \textbf{0.78} & 1.00 & 0.94 & 0.70 & 0.90 & 0.53 & 0.65 \\
    \toprule
    \toprule
    \multicolumn{8}{l}{\textbf{1024 \boldmath$\times$\unboldmath 1024 resolution}} \\
    \midrule
    LUMINA-Next~\citep{zhuo2024lumina}      & 2.0  & \textbf{0.46}   & 0.92   & 0.46   & 0.48   & 0.70  & 0.09 & 0.13 \\
    SDXL~\citep{podell2023sdxl}             & 2.6  & \textbf{0.55}   & 0.98   & 0.74   & 0.39   & 0.85  & 0.15 & 0.23 \\
    PlayGroundv2.5~\citep{li2024playground} & 2.6  & \textbf{0.56}   & 0.98   & 0.77   & 0.52   & 0.84  & 0.11 & 0.17 \\
    Hunyuan-DiT~\citep{li2024hunyuan}       & 1.5  & \textbf{0.63}   & 0.97   & 0.77   & 0.71   & 0.88  & 0.13 & 0.30 \\ 
    DALLE3~\citep{Dalle-3}                  & -    & \textbf{0.67}   & 0.96   & 0.87   & 0.47   & 0.83  & 0.43 & 0.45 \\
    FLUX-dev~\citep{flux}                   & 12.0 & \textbf{0.67}   & 0.99	 & 0.81	  & 0.79   & 0.74  & 0.20 & 0.47 \\
    \midrule
    \textbf{\model{}-T2I}                   & 1.3  & \textbf{0.74}   & 1.00   & 0.95   & 0.55   & 0.88  & 0.41 & 0.68 \\ 
    \bottomrule
    \end{tabular*}
    \caption{\textbf{GenEval category-wise results} at $512\times512$ and $1024\times1024$ for text-to-image generation.
    \emph{Overall} is the unweighted mean over Single Object, Two Objects, Counting, Colors, Position, and Color Attribution.}
    \label{tab:geneval_details}
\end{table*}

\subsection{Ablation on Depth N and M}
We analyze how to allocate depth between the patch-level pathway ($N$ layers) and the pixel-level pathway ($M$ layers) under a fixed total budget of roughly $N{+}M{\approx}30$ layers.
Figure~\ref{fig:app:nm_curves} shows convergence curves for several $(N,M)$ configurations evaluated on ImageNet 256$\times$256. All evaluations use the same CFG guidance scale $3.25$ with interval $[0.10, 1.00]$.
Introducing even a shallow pixel pathway (e.g., $N{=}28, M{=}2$) dramatically improves convergence and reduces final gFID to around $2.1$.
Our default configuration ($N{=}26, M{=}4$) provides the best overall behavior: it reaches gFID $2.34$ by 300K iterations and continues to improve to $1.94$ at 1M iterations, outperforming both the shallower pixel pathway ($N{=}28, M{=}2$) and the deeper one ($N{=}22, M{=}8$).
The latter attains similar final gFID but converges more slowly in early epochs.
These trends indicate that dedicating a moderate but not excessive number of layers to the pixel-level pathway is crucial for efficient pixel modeling and underpins the strong ImageNet results reported in the main paper.

\subsection{Ablation on Representation Alignment (REPA)}
\label{sec:app:repa_ablation}

We ablate the representation alignment (REPA) loss that encourages patch-level features to stay aligned with frozen DINOv2 encoder features. Table~\ref{tab:app:repa_ablation} compares \model{}-XL with and without REPA on ImageNet 256$\times$256 at 80 and 160 epochs. Removing REPA leads to substantially worse FID and IS at both checkpoints: at 80 epochs, FID degrades from 2.36 to 6.58 and IS drops from 282.3 to 165.96. While longer training partially closes the gap (FID improves from 6.58 to 4.33 at 160 epochs without REPA), the model with REPA consistently outperforms by a large margin. These results confirm that representation alignment is an essential component of our pixel-space diffusion framework, providing critical regularization for the patch-level pathway to maintain semantically meaningful representations throughout training.

\begin{table}[h]
    \centering
    \footnotesize
    \begin{tabular*}{\linewidth}{@{\extracolsep{\fill}}lcccc@{}}
        \toprule
         & \multicolumn{2}{c}{\textbf{80 epochs}} & \multicolumn{2}{c}{\textbf{160 epochs}} \\ \cmidrule(lr){2-3} \cmidrule(lr){4-5}
        \textbf{Configuration} & \textbf{FID}$\downarrow$ & \textbf{IS}$\uparrow$ & \textbf{FID}$\downarrow$ & \textbf{IS}$\uparrow$ \\
        \midrule
        PixelDiT-XL w/ REPA & \textbf{2.36} & \textbf{282.3} & \textbf{1.97} & \textbf{299.4} \\
        PixelDiT-XL w/o REPA & 6.58 & 165.9 & 4.33 & 242.4 \\
        \bottomrule
    \end{tabular*}
    \caption{\textbf{Ablation of representation alignment (REPA)} on ImageNet 256$\times$256. Removing REPA leads to substantially degraded FID and IS at both training stages.}
    \label{tab:app:repa_ablation}
\end{table}

\begin{table*}[t]
    \footnotesize
    \centering
    \begin{tabular*}{\linewidth}{@{\extracolsep{\fill}}lccccccc@{}}
    \toprule
    \textbf{Model} & \textbf{Params}~(B) & \textbf{Overall}~$\uparrow$ & Global & Entity & Attribute & Relation & Other \\
    \midrule
    \multicolumn{8}{l}{\textbf{512 \boldmath$\times$\unboldmath 512 resolution}} \\
    \midrule
    PixArt-$\alpha$~\citep{chenpixart}      & 0.6 & \textbf{71.6} & 81.7 & 80.1 & 80.4 & 81.7 & 76.5 \\
    PixArt-$\Sigma$~\citep{chen2024pixart}  & 0.6 & \textbf{79.5} & 87.5 & 87.1 & 86.5 & 84.0 & 86.1 \\
    PixelFlow~\citep{pixelflow}   & 0.9 & \textbf{77.9} & - & - & - & - & - \\
    PixNerd~\citep{pixnerd}       & 1.2 & \textbf{80.9} & 80.5 & 87.9 & 87.2 & 91.3 & 72.8 \\
    \midrule
    \textbf{\model{}-T2I}     & 1.3 & \textbf{83.7} & 88.0 & 90.9 & 87.6 & 89.8 & 88.5 \\
    \toprule
    \toprule
    \multicolumn{8}{l}{\textbf{1024 \boldmath$\times$\unboldmath 1024 resolution}} \\
    \midrule
    LUMINA-Next~\citep{zhuo2024lumina}      & 2.0    & \textbf{74.6} & 82.8 & 88.7 & 86.4 & 80.5 & 81.8 \\
    SDXL~\citep{podell2023sdxl}             & 2.6    & \textbf{74.7} & 83.3 & 82.4 & 80.9 & 86.8 & 80.4 \\
    PlayGroundv2.5~\citep{li2024playground} & 2.6    & \textbf{75.5} & 83.1 & 82.6 & 81.2 & 84.1 & 83.5 \\
    Hunyuan-DiT~\citep{li2024hunyuan}       & 1.5    & \textbf{78.9} & 84.6 & 80.6 & 88.0 & 74.4 & 86.4 \\ 
    PixArt-$\Sigma$~\citep{chen2024pixart}  & 0.6    & \textbf{80.5} & 86.9 & 82.9 & 88.9 & 86.6 & 87.7 \\
    DALLE3~\citep{Dalle-3}                  & -      & \textbf{83.5} & 91.0 & 89.6 & 88.4 & 90.6 & 89.8 \\
    FLUX-dev~\citep{flux}                   & 12.0   & \textbf{84.0} & 82.1 & 89.5 & 88.7 & 91.1 & 89.4 \\
    \midrule
    \textbf{\model{}-T2I}                   & 1.3  & \textbf{83.5}   & 83.0 & 88.6 & 87.8 & 91.2 & 89.6 \\ 
    \bottomrule
    \end{tabular*}
    \caption{\textbf{DPG-Bench category-wise results} at $512\times512$ and $1024\times1024$ resolutions for text-to-image generation.
    }
    \label{tab:dpgbench_imagereward_details}
\end{table*}

\subsection{Study on Pixel Token Compaction (PTC) Rate}

We investigate the effectiveness of Pixel Token Compaction (PTC) by varying the compaction rate. 
Recall that our default patch size is $p$. Without compaction, the pixel-level pathway would process a sequence of length $H \times W$. 
With standard compaction (denoted as ``Seq Len $L$ ($1\times$)'' or Base), the $p \times p$ pixels in a patch are compressed into a single token, reducing the sequence length to $L = (H/p) \times (W/p)$.
We explore relaxing this compression by allowing the compacted sequence length to be multiples of the base length $L$. Specifically:
\begin{itemize}
    \item \textbf{Seq Len $L$ ($1\times$)}: The default setting. Compresses $p^2$ pixels to 1 token. Compression rate: $p^2$.
    \item \textbf{Seq Len $2L$ ($2\times$)}: Compresses $p^2$ pixels to 2 tokens. Compression rate: $p^2/2$.
    \item \textbf{Seq Len $4L$ ($4\times$)}: Compresses $p^2$ pixels to 4 tokens. Compression rate: $p^2/4$.
\end{itemize}

Figure~\ref{fig:app:ptc_curves} presents the ablation results.
Across training, all three settings converge to strong gFID values around $2.0$, while the model with the most aggressive compression (Seq Len $1\times$) obtains slightly better results.
For example, at 300K iterations the three configurations obtain gFID of roughly $2.34$ ($1\times$), $2.38$ ($2\times$), and $2.43$ ($4\times$), respectively.  At 1M iterations the $1\times$ variant is further improved to $1.94$ while the longer sequences plateau slightly higher.
The result suggests that, for the pixel-level pathway, a compact representation is sufficient to capture the residual information needed for texture refinement. This could be due to the redundant nature of the pixel-space tokens. Interestingly, the results indicate that lower compression rates do not necessarily lead to better image quality. We suspect that a longer, redundant token sequence and a larger attention space can be more challenging to optimize and slower to converge under the same training setting. \textit{Using more tokens in the pixel-level pathway may require carefully adjusted training settings to unlock its full potential}. 
Since the attention cost grows almost quadratically with the compressed token length, the model with $1\times$ compaction performs similarly to other configurations, thus we adopt it as the default setting throughout the paper for efficiency.

\subsection{Patch Size Ablation on ImageNet-512}

We further extend the patch size analysis to ImageNet-512 to examine whether a larger patch size ($p{=}32$) could provide a favorable compute--quality trade-off at higher resolution. As shown in Table~\ref{tab:app:patch512}, while $p{=}32$ reduces computational cost (see Table~\ref{tab:app:gflops_resolution}), it consistently underperforms $p{=}16$ at both 80 and 120 epochs. At 120 epochs, $p{=}16$ achieves an FID of 2.23 compared to 3.78 for $p{=}32$. These results suggest that the finer-grained patch tokenization remains important at higher resolutions, and that the quality benefit of smaller patches outweighs the compute savings from larger patches.

\begin{table}[t]
    \centering
    \footnotesize
    \begin{tabular*}{\linewidth}{@{\extracolsep{\fill}}lcccc@{}}
        \toprule
        \multirow{2}{*}{\textbf{Patch}} & \multicolumn{2}{c}{\textbf{80 epochs}} & \multicolumn{2}{c}{\textbf{120 epochs}} \\
        \cmidrule(lr){2-3}\cmidrule(lr){4-5}
        & \textbf{FID}$\downarrow$ & \textbf{IS}$\uparrow$ & \textbf{FID}$\downarrow$ & \textbf{IS}$\uparrow$ \\
        \midrule
        $p{=}16$ & \textbf{2.97} & \textbf{270.29} & \textbf{2.23} & \textbf{287.62} \\
        $p{=}32$ & 4.66 & 235.52 & 3.78 & 256.06 \\
        \bottomrule
    \end{tabular*}
    \caption{\textbf{Patch size ablation on ImageNet-512.} $p{=}16$ consistently outperforms $p{=}32$ despite the latter's lower compute cost.}
    \label{tab:app:patch512}
\end{table}

\section{Benchmark Details}
\subsection{DPG-Bench and GenEval Category Breakdown}
Tables~\ref{tab:geneval_details} and \ref{tab:dpgbench_imagereward_details} report category-wise results for GenEval and DPG-Bench at $512^2$ and $1024^2$ resolutions. 
On GenEval at $512^2$, \model{}-T2I outperforms prior pixel-space models.
At $1024^2$ resolution, \model{}-T2I matches or surpasses several widely used latent diffusion systems despite using fewer parameters, and maintains competitive performance across all individual categories.
On DPG-Bench, \model{}-T2I ranks among the top-performing models while maintaining balanced scores across categories.
These detailed breakdowns corroborate that PixelDiT delivers strong text–image alignment and compositional reasoning, closing much of the gap to heavily engineered latent diffusion models.

\subsection{Editing Background Preservation}
\label{sec:app:editing_eval}

A key advantage of pixel-space diffusion models over latent-space counterparts is the ability to preserve fine-grained details in unedited regions during image editing, since pixel-space models avoid the information loss introduced by VAE encoding and decoding. To quantify this advantage, we evaluate background preservation using the FlowEdit~\cite{kulikov2025flowedit} dataset, which consists of 281 source--target editing pairs. For each sample, we obtain the editing bounding box using SAM3 and compute MSE and SSIM on the background region outside the bounding box. As shown in Table~\ref{tab:app:editing_bg}, our pixel-space \model{} achieves substantially better background consistency than the latent-space models FLUX and SD3, with $6.0\times$ lower MSE and higher SSIM compared to FLUX. This confirms that pixel-space diffusion models naturally preserve unedited regions more faithfully, which is a practically important property for editing applications.

\begin{table}[h]
    \centering
    \footnotesize
    \begin{tabular*}{\linewidth}{@{\extracolsep{\fill}}lcc@{}}
        \toprule
         & \multicolumn{2}{c}{\textbf{Background Consistency}} \\ \cmidrule(lr){2-3}
         \textbf{Method} & \textbf{MSE}$\downarrow$ & \textbf{SSIM}$\uparrow$ \\
        \midrule
        FLUX & 0.009105 & 0.8254 \\
        SD3 & 0.004349 & 0.8400 \\
        \textbf{PixelDiT} & \textbf{0.001522} & \textbf{0.8628} \\
        \bottomrule
    \end{tabular*}
    \caption{\textbf{Background preservation in image editing.} We evaluate background consistency (MSE and SSIM outside the editing bounding box) on 281 FlowEdit samples. \model{} preserves unedited regions significantly better than latent-space models.}
    \label{tab:app:editing_bg}
\end{table}

\section{FLOPs Estimation and Comparison}

We estimate GFLOPs for a single forward pass at $256^2$ input resolution.   For the GFLOPs of prior work, we reuse the numbers reported in their papers~\cite{sid2,jit} and convert them to a \textbf{unified convention where one multiply-add counts as two FLOPs}.
Table~\ref{tab:app:gflops_comparison} compares the compute cost and  FID of \model{}-XL  with representative latent-space and pixel-space generative models on ImageNet 256$\times$256.
Latent models achieve very strong FIDs with around $240$--$290$ GFLOPs, whereas many pixel models require several hundred to several thousand GFLOPs to close this quality gap.
In contrast, \model{}-XL obtains a \textbf{1.61} FID with only \textbf{311} GFLOPs, offering superior image quality over the state-of-the-art pixel generators and closing much of the gap between the best latent models, while using a compute that is close to latent models and substantially less than most prior pixel-space models.

\begin{table}[t]
    \centering
    \resizebox{\linewidth}{!}{
    \begin{tabular}{l c c c}
        \toprule
        \textbf{Methods} & \textbf{Params} & \makecell{\textbf{GFLOPs} \\ (multi-add = 2 FLOPs)} & \textbf{FID}$\downarrow$ \\
        \midrule
        \multicolumn{4}{l}{\textit{Latent Generative Models}} \\ \midrule
        DiT-XL/2~\cite{dit} & 675+49M & 238 & 2.27 \\
        SiT-XL/2~\cite{sit} & 675+49M & 238 & 2.06 \\
        REPA, SiT-XL/2~\cite{repa} & 675+49M & 238 & 1.42 \\
        LightningDiT-XL/2~\cite{lgt} & 675+49M & 238 & 1.35 \\
        DDT-XL/2~\cite{ddt} & 675+49M & 238 & 1.26 \\
        RAE, DiT$^{\text{DH}}$-XL/2~\cite{zheng2025rae} & 839+415M & 292 & 1.13 \\
        \midrule
        \multicolumn{4}{l}{\textit{Pixel Generative Models}} \\\midrule
        ADM-G~\cite{adm} & 559M & 2240 & 7.72 \\
        RIN~\cite{rin} & 320M & 668 & 3.95 \\
        SiD, UViT/2~\cite{simplediffusion} & 2B & 1110 & 2.44 \\
        VDM++, UViT/2~\cite{vdm++} & 2B & 1110 & 2.12 \\
        SiD2, UViT/2~\cite{sid2} & N/A & 274 & 1.73 \\
        SiD2, UViT/1~\cite{sid2} & N/A & 1306 & 1.38 \\
        PixelFlow-XL/4~\cite{pixelflow} & 677M & 5818 & 1.98 \\
        PixNerd-XL/16~\cite{pixnerd} & 700M & 268 & 2.15 \\
        JiT-G/16~\cite{jit} & 2B & 766 & 1.82 \\
        \midrule
        \textbf{\model{}-XL (ours)} & 797M & 311 & 1.61 \\
        \bottomrule
    \end{tabular}}
    \vspace{-.2em}
    \caption{\textbf{Compute comparison on ImageNet 256$\times$256.}
    We report model parameters, GFLOPs per forward pass, and FID under our convention that one multiply-add equals two FLOPs.
    }
    \label{tab:app:gflops_comparison}
\end{table}

We further provide a detailed GFLOPs breakdown of \model{}-XL across different input resolutions and patch sizes in Table~\ref{tab:app:gflops_resolution}. The compute cost scales roughly quadratically with resolution at a fixed patch size, reflecting the quadratic cost of self-attention over the patch token sequence. Increasing the patch size dramatically reduces the cost: at $1024^2$, moving from $p{=}8$ to $p{=}16$ reduces GFLOPs by $7.4\times$, and $p{=}32$ further reduces it by $3.1\times$. These results provide practical guidance for resolution--compute trade-offs and motivate our default choice of $p{=}16$, which balances quality (see Table~\ref{tab:app:patch512}) and efficiency.

\begin{table}[h]
    \centering
    \footnotesize
    \begin{tabular*}{\linewidth}{@{\extracolsep{\fill}}lccc@{}}
        \toprule
        \textbf{Patch Size} & \textbf{256$^2$} & \textbf{512$^2$} & \textbf{1024$^2$} \\
        \midrule
        8$\times$8 & 1115.69 & 6200.99 & 52634.10 \\
        16$\times$16 & 311.19 & 1352.24 & 7147.17 \\
        32$\times$32 & 135.54 & 547.74 & 2298.42 \\
        \bottomrule
    \end{tabular*}
    \caption{\textbf{GFLOPs of \model{}-XL} across input resolutions and patch sizes. Compute scales roughly quadratically with resolution at a fixed patch size.}
    \label{tab:app:gflops_resolution}
\end{table}

\section{Extended Training Iterations}
We extend the training of \model{}-XL on ImageNet 256$\times$256 beyond the 320-epoch setting used in the main paper to an 800-epoch schedule.
Table~\ref{tab:app:extended_training} reports gFID, sFID, IS, precision, and recall at 80, 320, and 800 epochs.
Performance steadily improves from 80 to 320 epochs: gFID decreases from $2.36$ to $1.61$ and recall rises from $0.57$ to $0.64$, while sFID and IS also become better.
Extending training further to 800 epochs yields additional but smaller gains, with gFID reaching $1.54$ and recall improving to $0.65$.

\begin{table}[htbp]
    \centering
    \resizebox{\linewidth}{!}{
    \begin{tabular}{lcccccc}
        \toprule
        Model & Epochs & gFID$\downarrow$ & sFID$\downarrow$ & IS$\uparrow$ & Prec.$\uparrow$ & Rec.$\uparrow$ \\
        \midrule
        PixelDiT-XL & 80     &  2.36 & 5.11 & 282.3 & 0.80 & 0.57  \\
        PixelDiT-XL & 320    & 1.61  & 4.68 & 292.7 & 0.78 & 0.64  \\
        PixelDiT-XL & 800    & 1.54  & 4.49 & 297.0 & 0.78 & 0.65  \\
        \bottomrule
    \end{tabular}
    }
    \caption{Quantitative results of PixelDiT-XL on ImageNet 256$\times$256 across extended training epochs.}
    \label{tab:app:extended_training}
\end{table}

In addition to extended pre-training at $256\times256$, we evaluate the effect of longer fine-tuning at $512\times512$ resolution. Starting from the 320-epoch ImageNet-256 checkpoint, we fine-tune \model{}-XL for 40 and 530 epochs.
Table~\ref{tab:app:imagenet512_finetune} presents the results. Increasing the fine-tuning duration improves generation quality.

\begin{table}[htbp]
    \centering
    \resizebox{\linewidth}{!}{%
    \begin{tabular}{lcccccc}
    \toprule
    \multirow{2}{*}{\textbf{Method}} & \multirow{2}{*}{\textbf{Epochs}} & \multicolumn{5}{c}{\textbf{Generation@512}} \\
    \cmidrule(lr){3-7}
     & & \textbf{gFID}$\downarrow$ & \textbf{sFID}$\downarrow$ & \textbf{IS}$\uparrow$ & \textbf{Precision}$\uparrow$ & \textbf{Recall}$\uparrow$ \\
    \midrule
    \model{}-XL & 320 + 40 & 2.21 & 5.84 & 271.1 & 0.78 & 0.65 \\
    \model{}-XL & 320 + 530 & 1.81 & 5.61 & 278.6 & 0.78 & 0.67 \\
    \bottomrule
    \end{tabular}%
    }
    \caption{\textbf{Quantitative comparison} on ImageNet 512$\times$512 with different fine-tuning durations. ``320 + $N$'' denotes fine-tuning for $N$ epochs from a 320-epoch ImageNet-256 checkpoint.}
    \label{tab:app:imagenet512_finetune}
\end{table}

\section{Empirical Insights and Failed Attempts -- (\textit{Things We Tried But Did Not Work})}
\label{sec:empirical_insights}

During the development of PixelDiT, we explored a large number of architectural variants and training strategies beyond what has been reported in the main paper. Many of these explorations involved not only our final PixelDiT design but also alternative pixel-space modeling paradigms. In this section, we share empirical observations and negative results that may be informative for future research on pixel-space diffusion models. We note that the observations below reflect trends across our experimental campaigns; since the experiments were not all conducted under perfectly controlled settings, we report qualitative findings rather than specific numbers.

\subsection{Pixel-Level Modeling Paradigms}

\paragraph{Factorized Spatial Modulation.}
As an alternative to our pixel-wise AdaLN, we explored \emph{factorized AdaLN}, where instead of producing distinct modulation parameters for every pixel ($P^2$ sets of parameters per patch), we generate modulation coefficients in a low-frequency DCT basis. Concretely, the conditioning vector produces $K$ spectral coefficients per modulation channel, and the per-pixel parameters are reconstructed via $\mathbf{m}(i) = \sum_{k} c_k \cdot \phi_k(i)$, where $\phi_k$ is a 2D separable cosine basis function. This reduces the parameter count from $O(P^2 \cdot C)$ to $O(K \cdot C)$ where $K \ll P^2$. In our experiments, this factorization did not match the performance of the full pixel-wise AdaLN. This is perhaps not surprising given that the low-frequency basis inherently limits the spatial resolution of the modulation signal.

\paragraph{Haar Wavelet Pretransform.}
We investigated applying a reversible Haar wavelet decomposition as a preprocessing step, transforming each patch from spatial domain into frequency subbands before the diffusion process. The forward transform uses pixel unshuffle followed by the $4\times4$ Haar matrix, decomposing the image into low-frequency and three high-frequency bands. We also explored band-specific loss weighting, downweighting high-frequency components. This approach did not yield improvements over direct pixel-space modeling in our setting.

\subsection{Conditioning and Feature Interaction}

\paragraph{Conditioning Mechanisms for Pixel-Level Modulation.}
We extensively investigated how patch-level semantic context should modulate the pixel-level transformer blocks. Beyond the pixel-wise AdaLN presented in the paper, we experimented with: (i)~\emph{additive conditioning}, where patch context is projected and directly added to the compressed pixel tokens before cross-patch attention; (ii)~\emph{cross-attention conditioning}, where pixel tokens attend to patch context tokens with separate query/key/value projections; (iii)~\emph{self-attention concatenation} (concatenating patch context tokens with pixel tokens along the sequence dimension and processing them jointly in a single self-attention layer); and (iv)~\emph{per-pixel additive conditioning}, where patch context is projected to the full pixel spatial resolution and added before token compaction. Among these alternatives, both cross-attention and self-attention concatenation led to worse generation quality compared to AdaLN. Additive conditioning does not noticeably harm performance. While it is slightly worse, it removes the $P^2$ parameters introduced by pixel-wise AdaLN, making it more scalable to increase the number of PiT blocks.

\paragraph{In-Context Prefix Tokens.}
Inspired by prefix tuning in language models, we experimented with prepending learnable context tokens to the patch sequence at a configurable depth. These tokens, initialized from the class embedding with added learnable positional embeddings, participate in self-attention alongside patch tokens and are stripped before the pixel pathway. This did not lead to measurable improvements in the generation quality on ImageNet class-conditional generation. This mechanism may be more relevant in settings with more complex conditioning (e.g., text-to-image).

\subsection{Training Stability and Optimization}

\paragraph{Representation Alignment Loss is Essential.}
We experimented with removing the feature alignment loss that encourages the patch-level representations to be aligned with the frozen DINOv2 encoder features. Without this auxiliary objective, the training became unstable and eventually diverged.

\paragraph{Prediction Target.}
In addition to velocity prediction, we experimented with $\mathbf{x}$-prediction, where the network directly predicts the clean image $\hat{\mathbf{x}}$ and the loss is computed in velocity space via $\hat{\mathbf{v}} = (\hat{\mathbf{x}} - \mathbf{x}_t) / \max(\sigma(t), \epsilon)$ with $\epsilon=0.05$ for numerical stability. In practice, we found that $\mathbf{x}$-prediction can effectively mitigate loss spikes during training. However, it did not outperform direct velocity prediction in our setting and required additional tuning of both the $\epsilon$ threshold and the logit-normal timestep sampling hyperparameters ($p_{\text{mean}}$, $p_{\text{std}}$), since the optimal timestep distribution differs between velocity and $\mathbf{x}$-prediction parameterizations.

\subsection{Architectural Design}

\paragraph{Encoder--Decoder Bottleneck for Token Compression.}
We explored adding a hierarchical encoder--decoder (ED) bottleneck around the patch-level transformer. The encoder progressively merges tokens via $2\times2$ patch merging stages, processes them with transformer blocks at reduced resolution, and the decoder mirrors this with patch expanding and skip connections. While the ED bottleneck architecture reduced computational cost, the optimal skip connection pattern (we tested many configurations, e.g., varying which encoder stages connect to which decoder stages) was highly sensitive to the overall architecture. The token compaction mechanism in PiT blocks ultimately proved to be a simpler and more robust alternative.

\paragraph{Multi-Scale Pixel Hidden Dimensions.}
In the pixel pathway, we explored varying the per-pixel hidden dimension across a wide range (from 4 to 128). Smaller dimensions were insufficient for good performance, while larger dimensions increased memory consumption in the compaction and expansion projections (which scale as $P^2 \times d_{\text{pixel}} \times d_{\text{attn}}$) without proportional quality gains. We found that a relatively compact (e.g., 16) pixel representation was sufficient when paired with adequate attention dimension in the cross-patch attention.

\paragraph{Shared vs.\ Separate Modulation.}
We tested sharing the AdaLN modulation network across all conditioning blocks, which reduces parameter count. This did not degrade quality substantially for shallower models but became limiting for deeper architectures where different layers may benefit from distinct modulation patterns.

\paragraph{Learnable vs.\ Fixed Positional Embeddings.}
We compared fixed sinusoidal 2D positional embeddings with fully learnable positional embeddings for the pixel-level tokens. Both approaches yielded comparable generation quality. Given the simplicity and resolution-generalizability of sinusoidal embeddings, we opted for fixed embeddings in our final model.

\section{Qualitative Examples}
\begin{figure*}[t]
    \centering
    \begin{subfigure}[b]{0.85\textwidth}
        \centering
        \includegraphics[width=\linewidth]{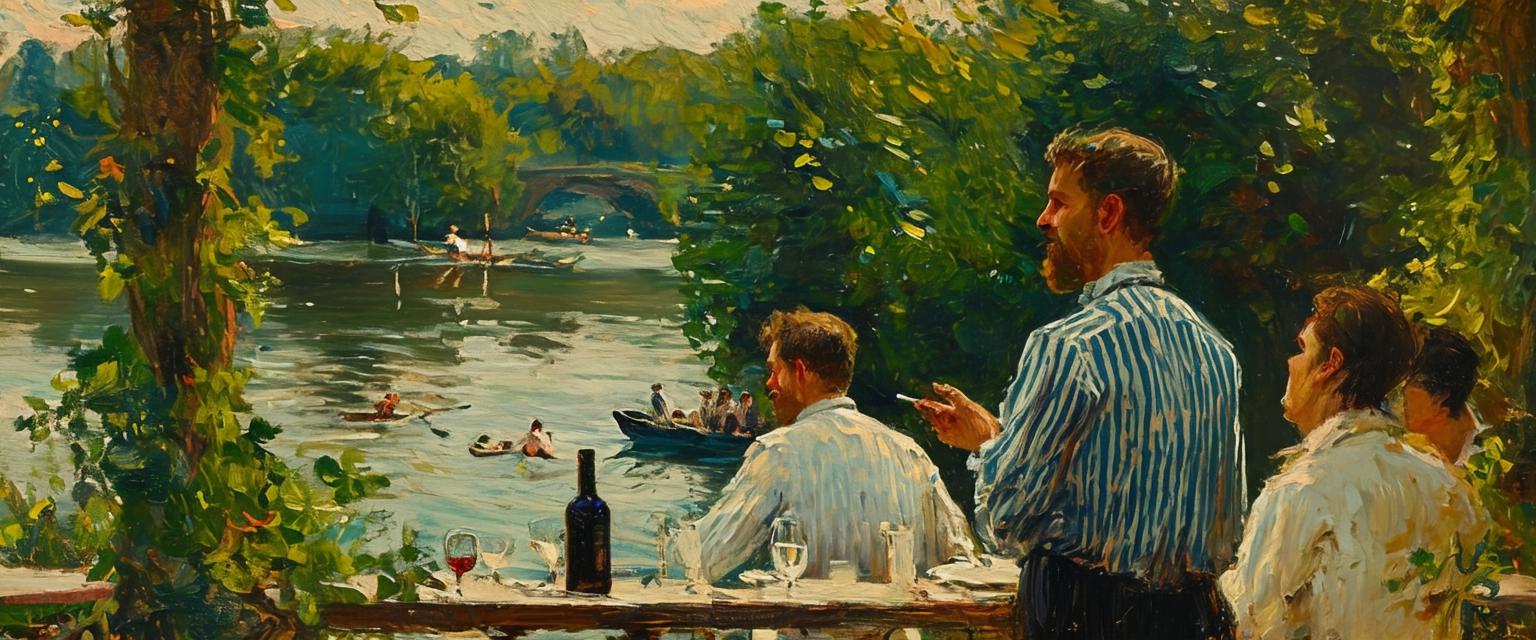}
        \caption*{\tiny A group of figures are gathered at a table near a trellised terrace, overlooking a river with rowers and a boat. The scene, rendered in an Impressionistic style, employs loose brushstrokes and soft, diffused light. The man in the foreground, dressed in a blue and white striped shirt, gestures casually, holding what appears to be a cigarette. The table is set with wine bottles and glasses, indicating a leisurely gathering. Through the trellis, a glimpse of the river reveals rowers in action, with a lone figure in a boat further out. The greenery of the trellis and the river landscape blend, contributing to the painting's overall sense of depth and atmospheric perspective.}
    \end{subfigure}
    \vspace{1em}
    \begin{subfigure}[b]{0.85\textwidth}
        \centering
        \includegraphics[width=\linewidth]{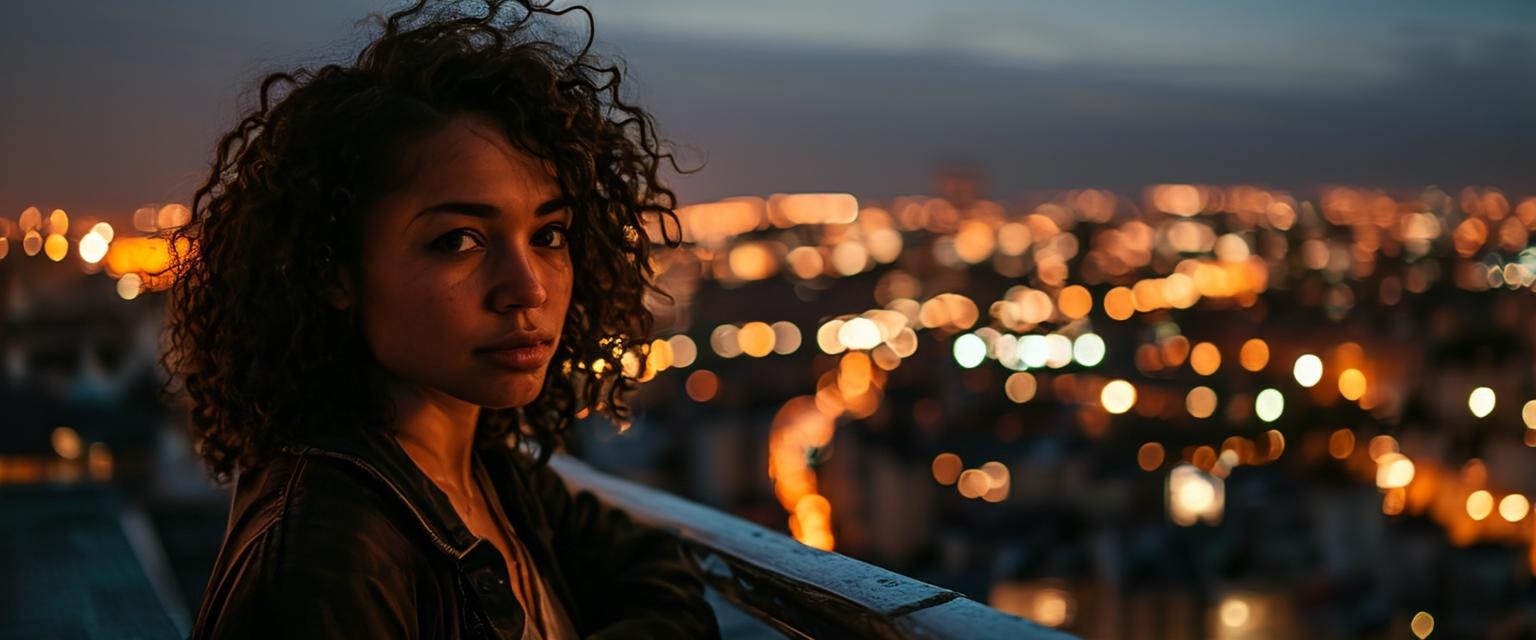}
        \caption*{\tiny A woman stands on a rooftop at dusk, overlooking a cityscape illuminated by twinkling lights. Her curly hair frames a serious expression, and she leans casually against the rooftop railing. The soft lighting of the sunset blends with the artificial glow of the city, creating a warm yet muted atmosphere. The out-of-focus background emphasizes the vastness of the urban landscape, dotted with skyscrapers and distant roads. Her dark jacket adds a layer of contrast, focusing the viewer's attention on her face. The overall style evokes a sense of urban solitude and reflection against a backdrop of a vibrant cityscape.}
    \end{subfigure}
    \vspace{1em}
    \begin{subfigure}[b]{0.85\textwidth}
        \centering
        \includegraphics[width=\linewidth]{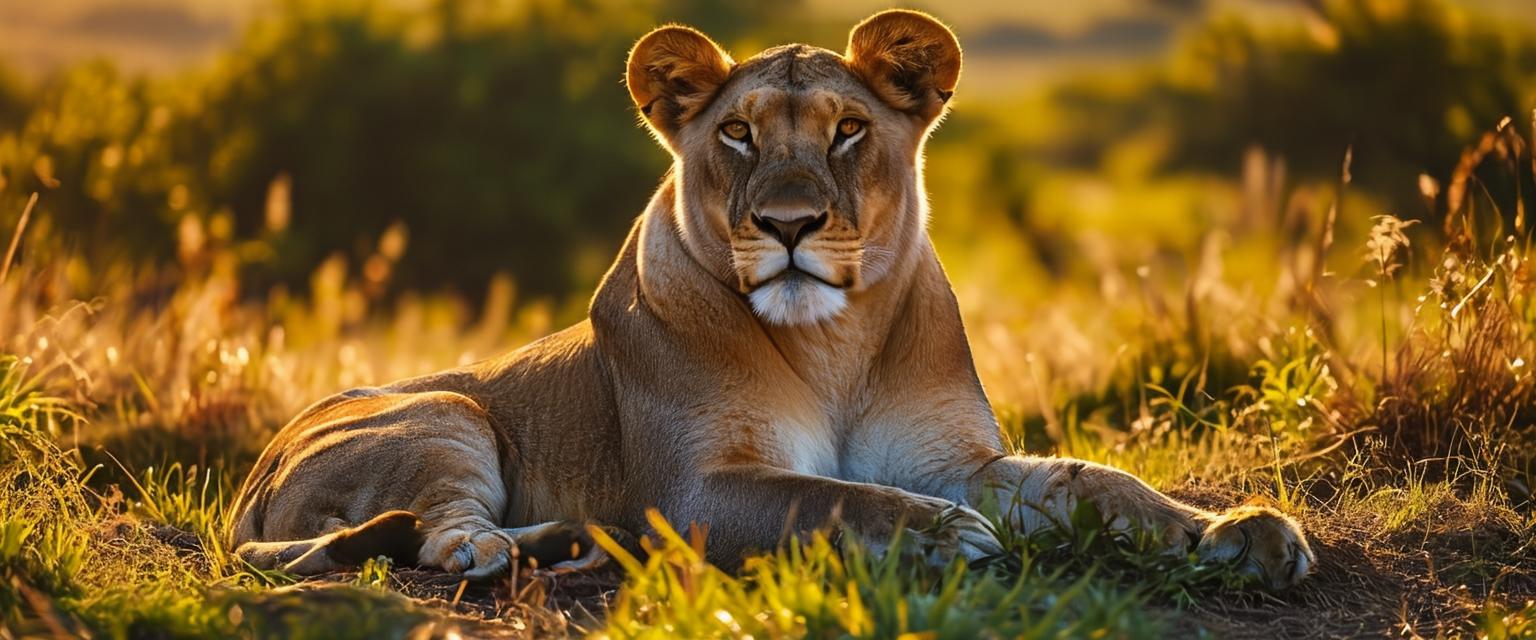}
        \caption*{\tiny A golden-hued lioness rests serenely in a sunlit grassy field. The lioness, bathed in the warm glow of the setting or rising sun, is positioned in the foreground, lying comfortably on a small mound of earth covered with dry grass. Her paws are outstretched, relaxed and slightly crossed. The background is a soft, blurred mix of green and golden foliage, hinting at a savanna-like landscape. The golden light emphasizes the texture of her fur and highlights the contours of her face, adding a sense of calm and natural grandeur to the scene.}
    \end{subfigure}
    \vspace{-3mm}
    \caption{Additional Text-to-Image Generation Results. The caption below each image corresponds to the text prompt used to generate it.}
    \label{fig:app:t2i_qual_0}
\end{figure*}

\begin{figure*}[t]
    \centering
    \begin{subfigure}[b]{0.32\textwidth}
        \centering
        \includegraphics[width=\linewidth]{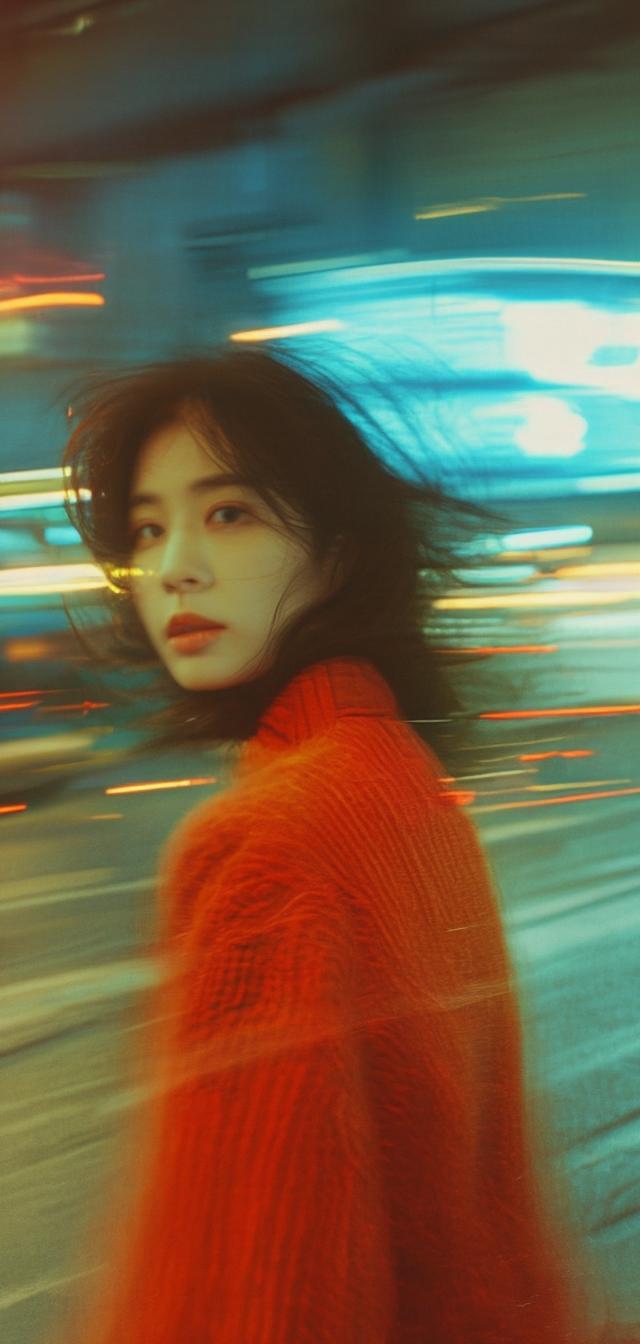}
        \caption*{\scriptsize Photo of a person moving with motion blur, shot with a Leica M6 and VISION3 500T Color Negative Film, reminiscent of a Wong Kar Tai film set.}
        \vspace{1em}
    \end{subfigure}
    \hfill
    \begin{subfigure}[b]{0.32\textwidth}
        \centering
        \includegraphics[width=\linewidth]{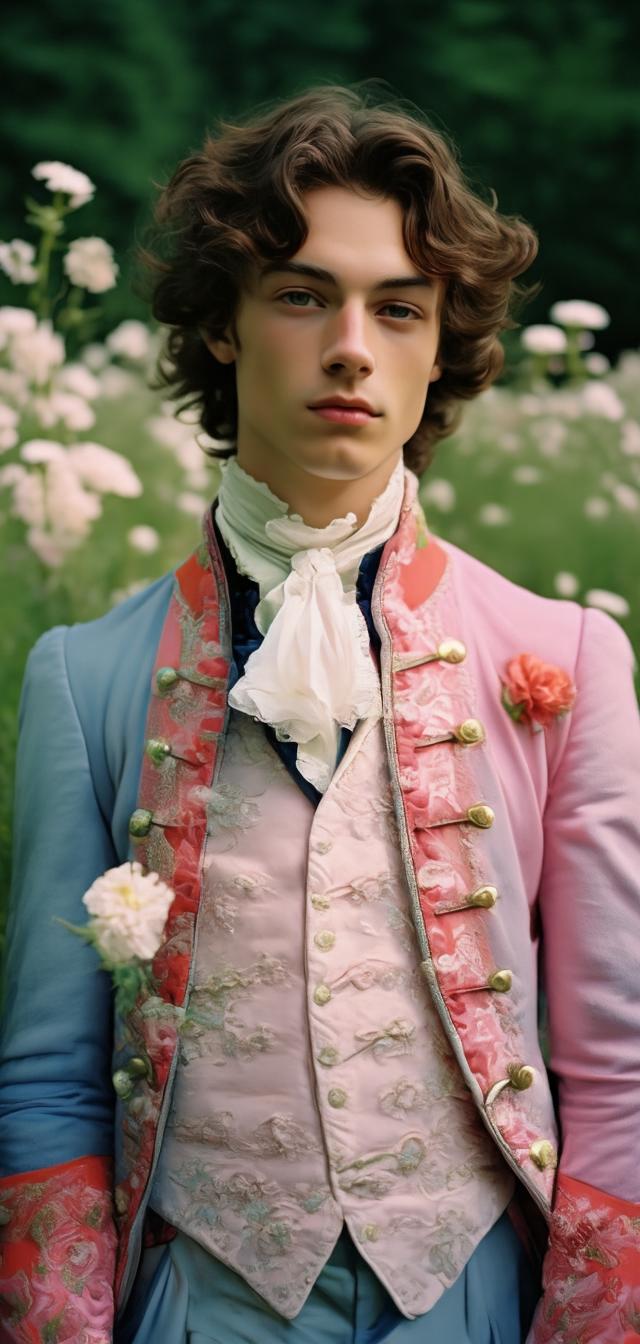}
        \caption*{\scriptsize A young man wearing 18th century noble clothing in blues and pinks and standing in front of the green grass with white flowers.}
        \vspace{1em}
    \end{subfigure}
    \hfill
    \begin{subfigure}[b]{0.32\textwidth}
        \centering
        \includegraphics[width=\linewidth]{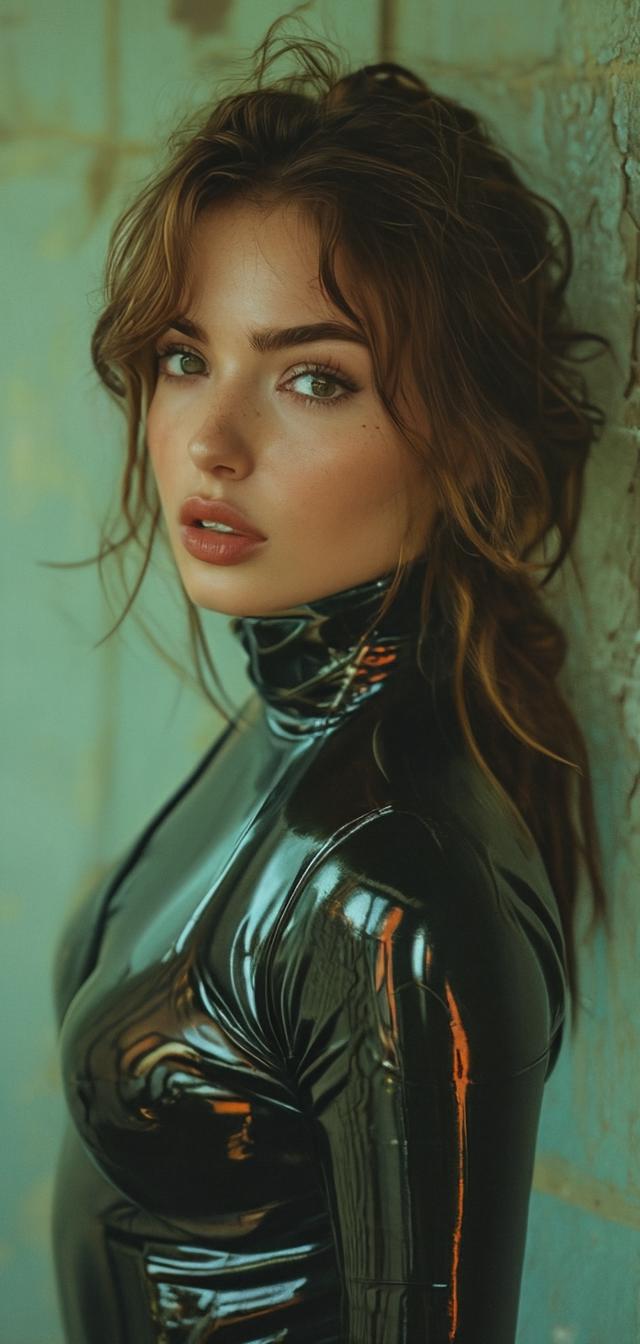}
        \caption*{\scriptsize Portrait shot of a pretty woman, latex suit fashion, contrasting background, fashion magazine cover, 35mm kodachrome.}
        \vspace{1em}
    \end{subfigure}
    \begin{subfigure}[b]{0.32\textwidth}
        \centering
        \includegraphics[width=\linewidth]{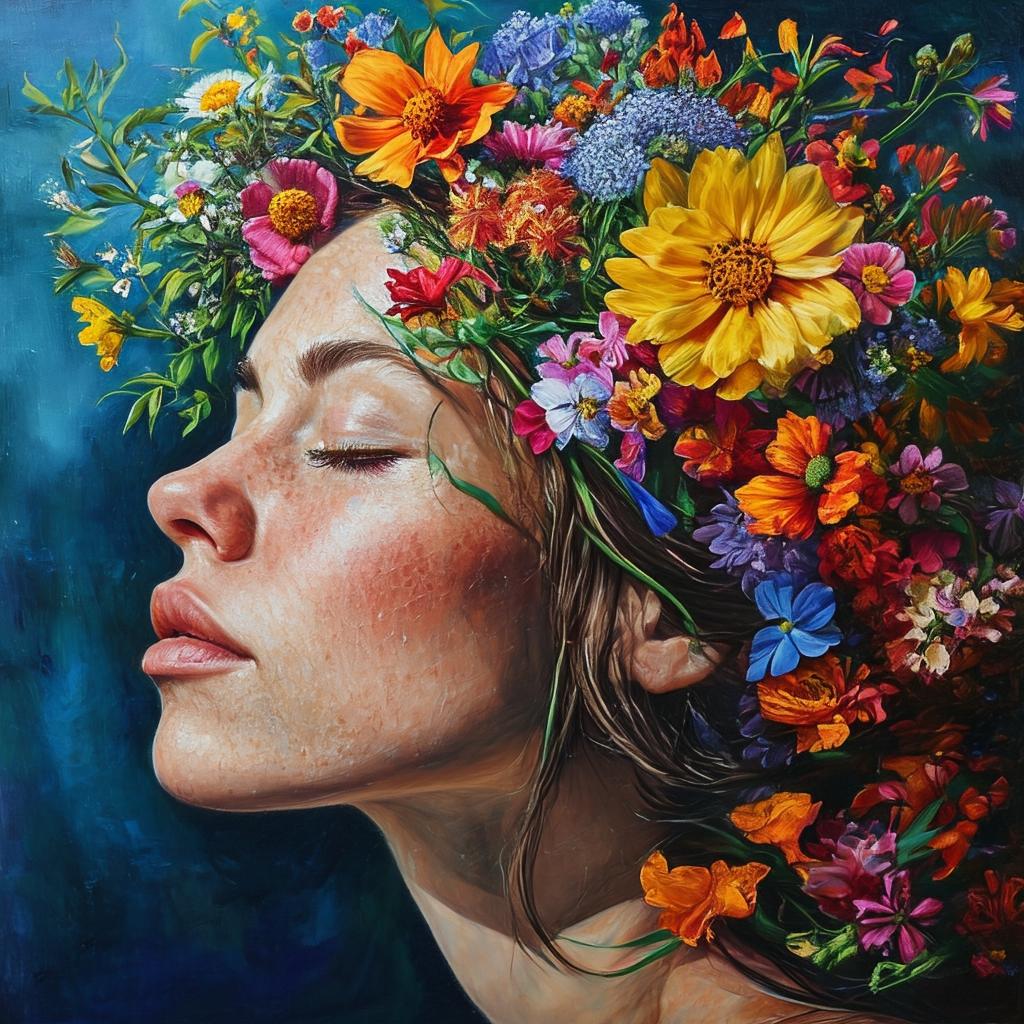}
        \caption*{\scriptsize A portrait of a human growing colorful flowers from her hair. Hyperrealistic oilpainting.}
    \end{subfigure}
    \hfill
    \begin{subfigure}[b]{0.32\textwidth}
        \centering
        \includegraphics[width=\linewidth]{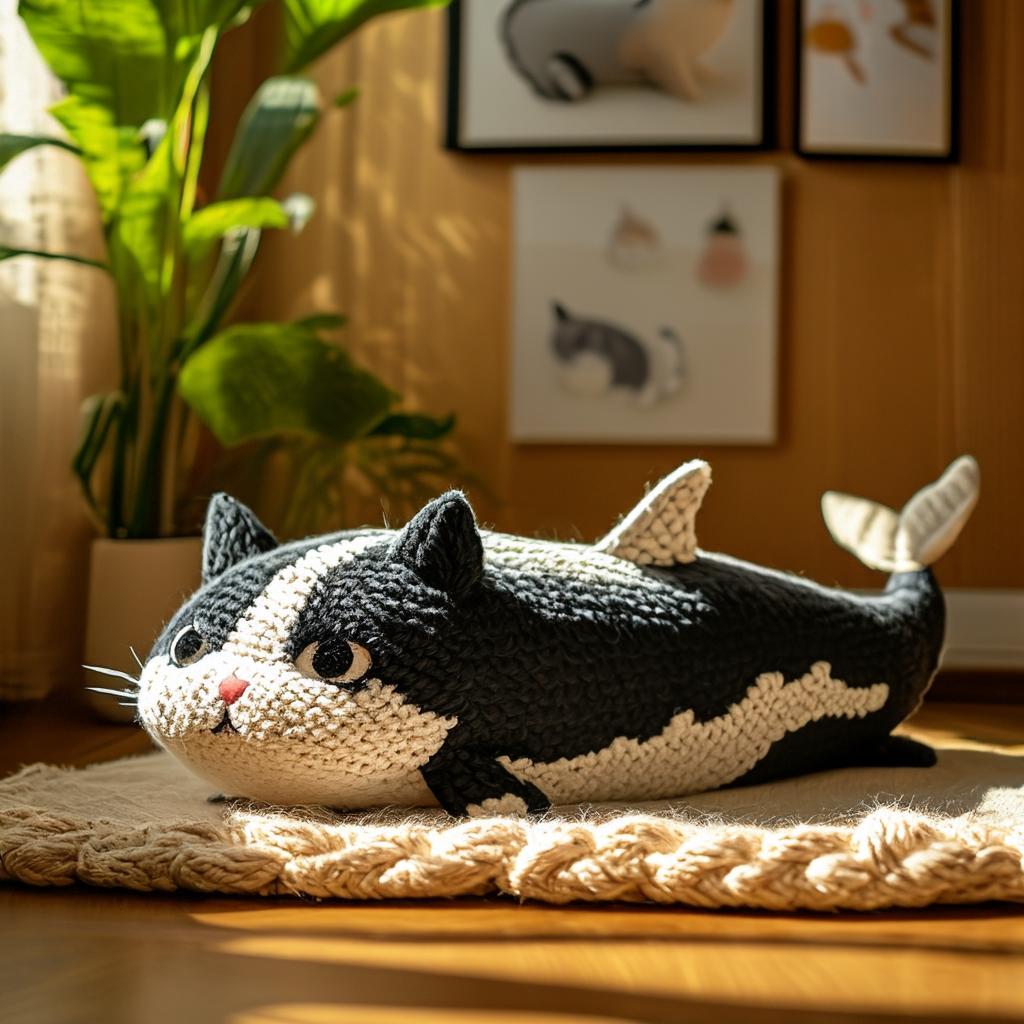}
        \caption*{\scriptsize knitted cat-whale plush toy on rug in warm sunlit living room, cozy decor.}
    \end{subfigure}
    \hfill
    \begin{subfigure}[b]{0.32\textwidth}
        \centering
        \includegraphics[width=\linewidth]{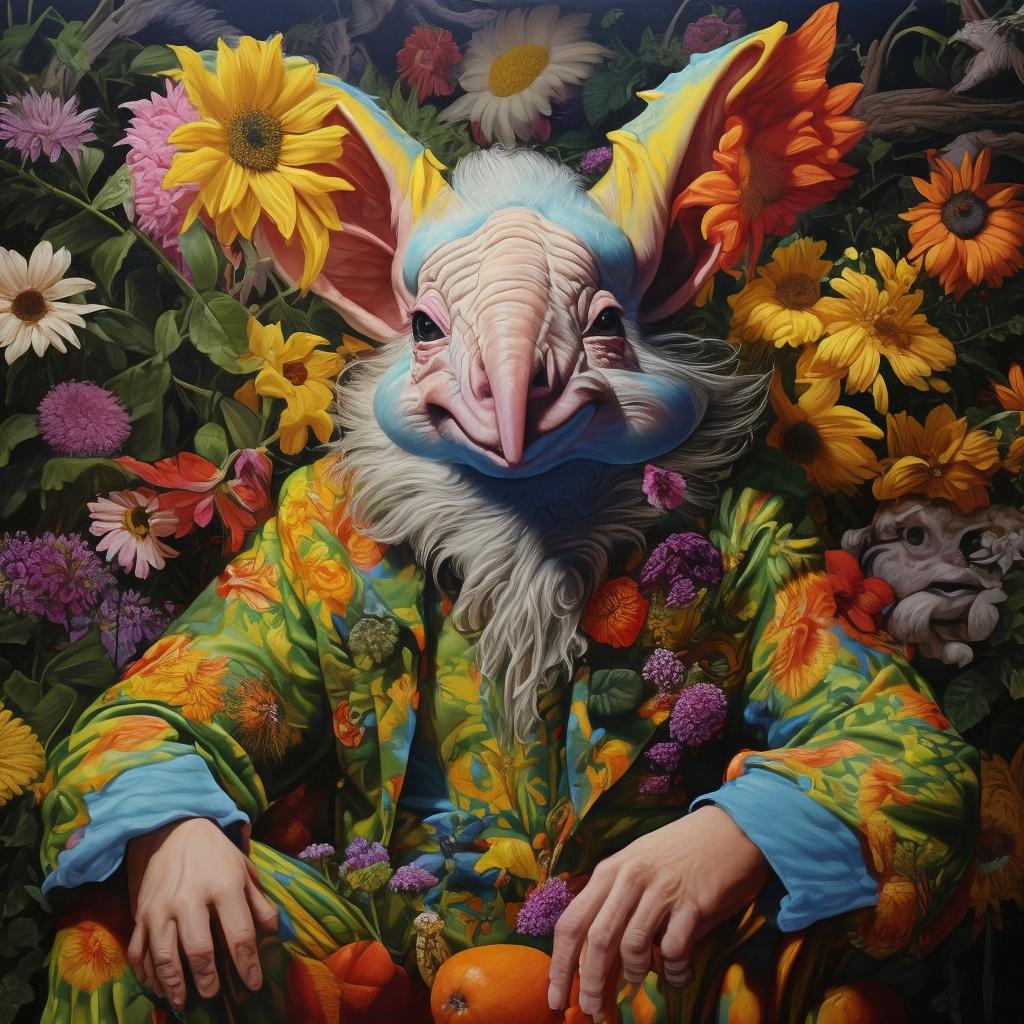}
        \caption*{\scriptsize Behold the Joymonger, photorealistic, 1990s, hyper realism, extremely detailed.}
    \end{subfigure}
    \caption{Additional Text-to-Image Generation Results. The caption below each image corresponds to the text prompt used to generate it.}
    \label{fig:app:t2i_qual_1}
\end{figure*}

\begin{figure*}[t]
    \centering
    \begin{subfigure}[b]{\textwidth}
        \centering
        \includegraphics[width=\linewidth]{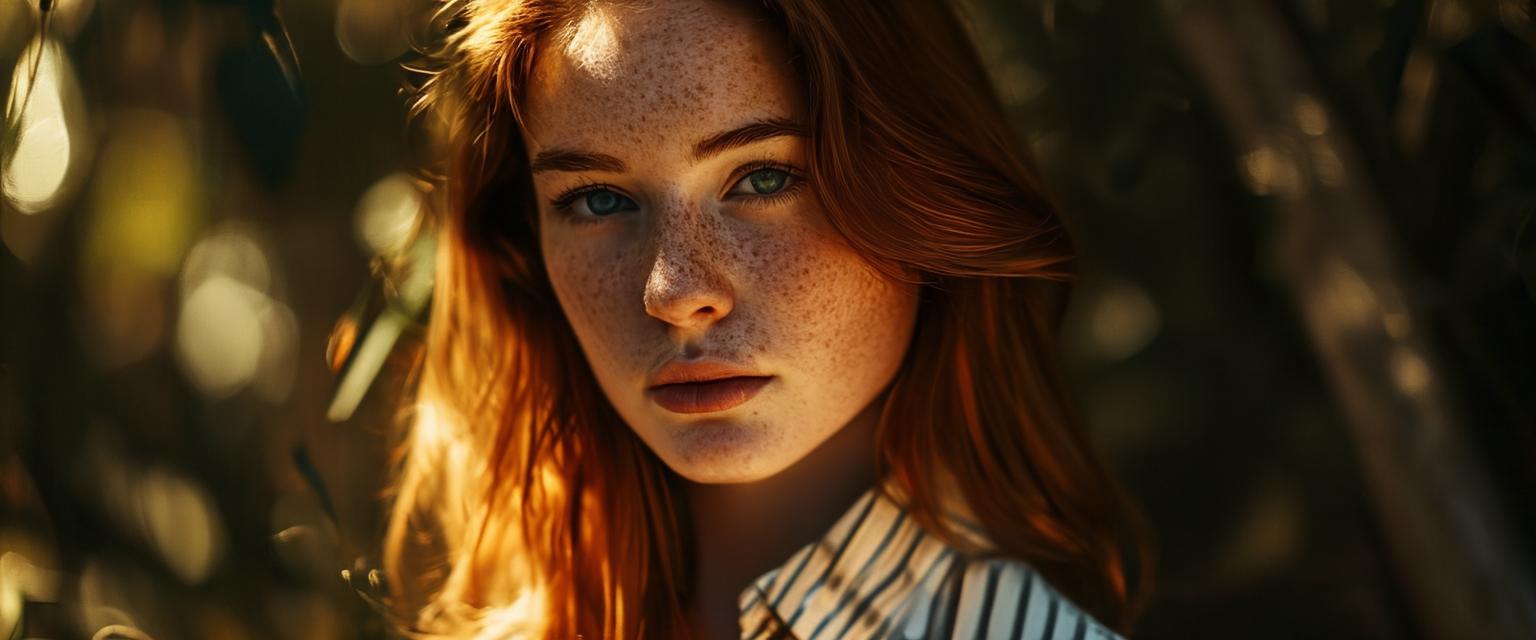}
        \caption*{\scriptsize A young woman with striking eyes gazes directly at the viewer, set against a soft, blurred background. Her long, auburn hair falls loosely around her shoulders, partially obscuring one side of her face, while a vertically striped, collared shirt completes her casual yet elegant look. The lighting is warm and natural, emphasizing the subtle contours of her face and the slight flush in her cheeks. The photograph captures a blend of relaxed confidence and introspective beauty, rendered with a soft focus that lends a dreamlike quality to the image. The surrounding environment is intentionally muted, ensuring that the woman remains the primary focal point. The color palette leans towards earthy tones, enhancing the overall warmth and approachability of the portrait.}
        \vspace{0.5em}
    \end{subfigure}
    \hfill
    \begin{subfigure}[b]{0.32\textwidth}
        \centering
        \includegraphics[width=\linewidth]{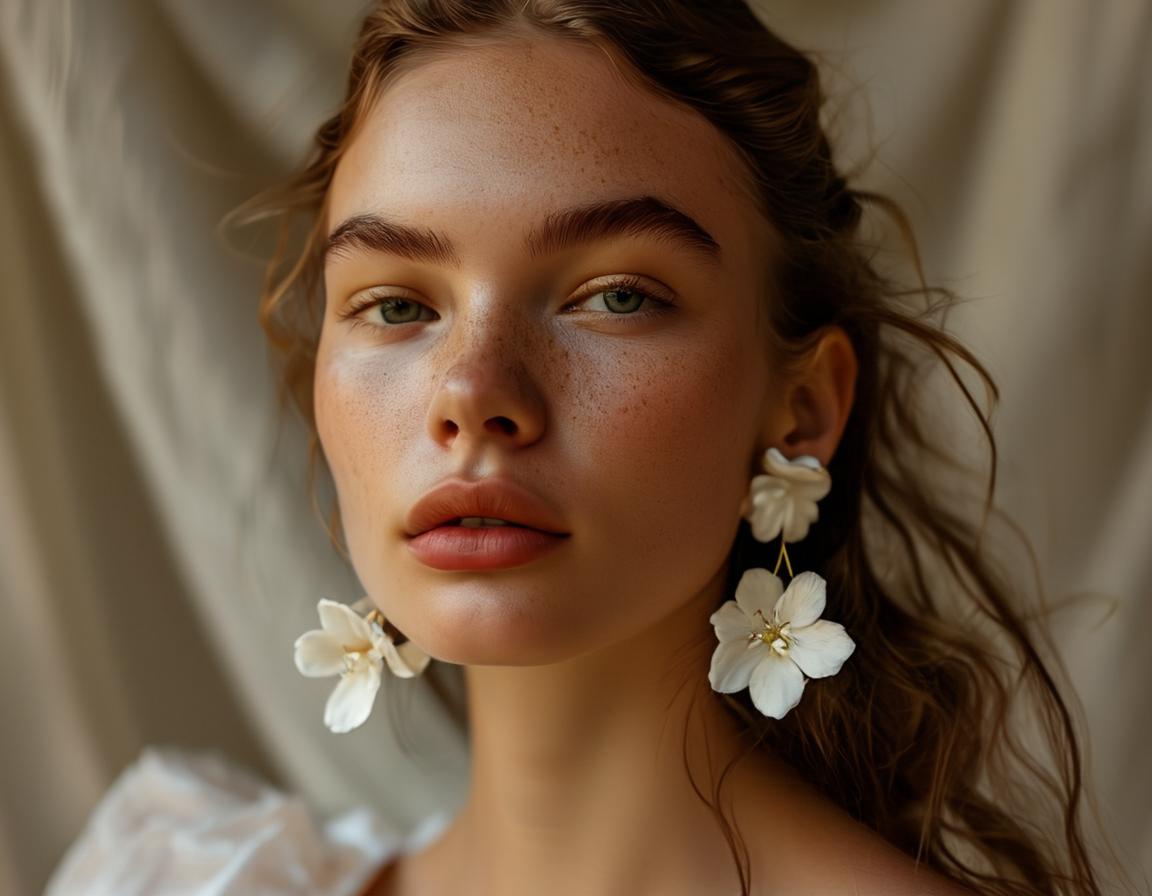}
        \caption*{\scriptsize Close-up portrait of a beautiful Baltic model wearing white flower-shaped earrings, emphasis on the earrings, everything is in full focus, pores and skin imperfections are visible, neutral lighting from a large studio softbox, natural beauty, professional studio shooting.}
    \end{subfigure}
    \hfill
    \begin{subfigure}[b]{0.32\textwidth}
        \centering
        \includegraphics[width=\linewidth]{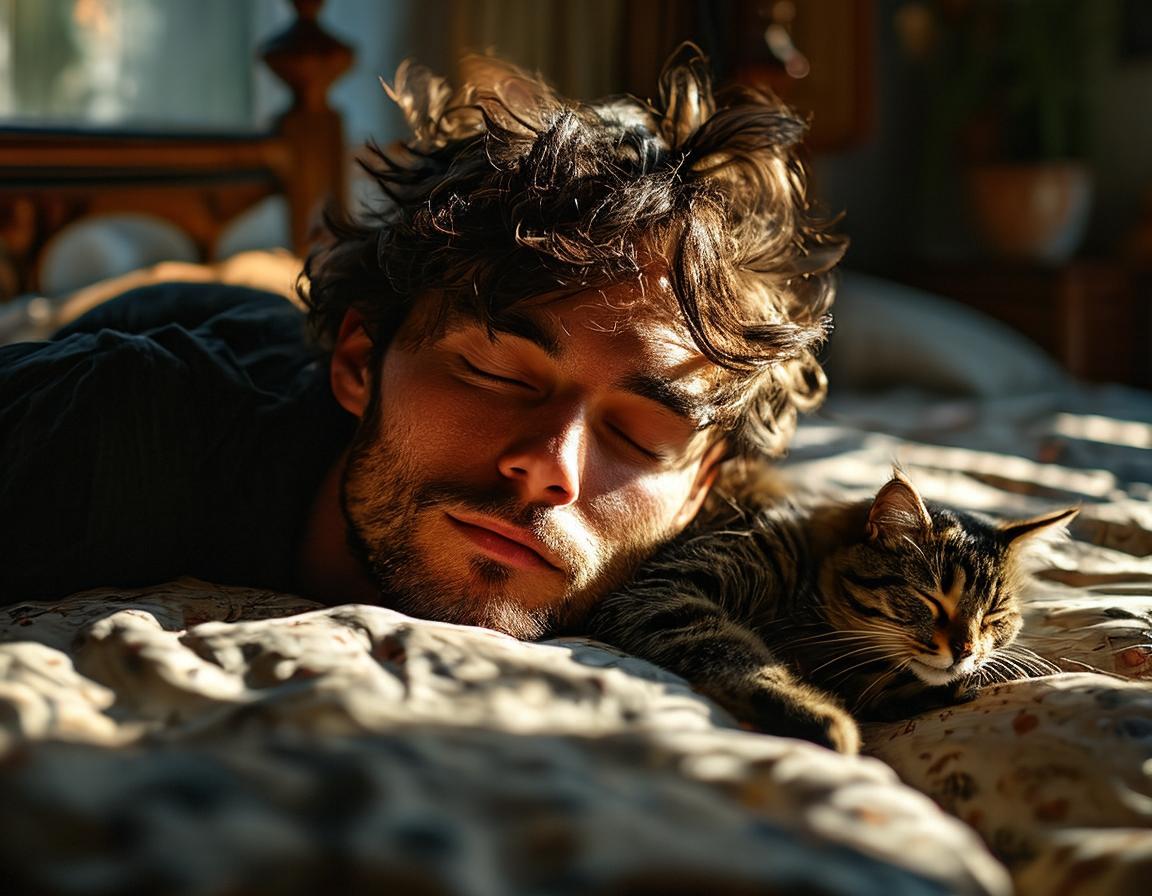}
        \caption*{\scriptsize funny Candid photo, cat sleeping across aman's eyes as he sleeps on his back,blocking his face, man is 25 years old,neck length frizzy black-brown hair,very unruly hair, stubble, wearing blackdress pants, long sleeve white button upshirt, socks, laying on a post bed.}
    \end{subfigure}
    \hfill
    \begin{subfigure}[b]{0.32\textwidth}
        \centering
        \includegraphics[width=\linewidth]{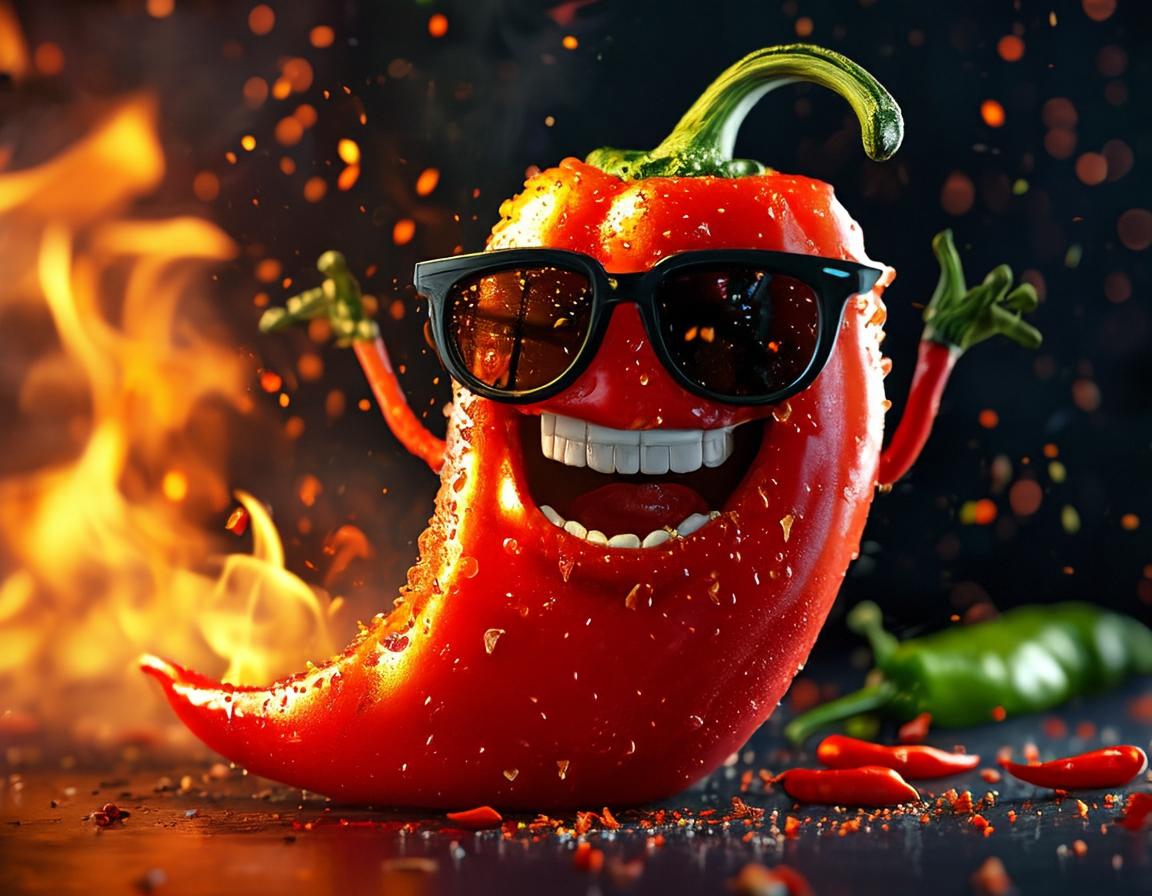}
        \caption*{\scriptsize Ultra-realistic photo of an anthropomorphic chili pepper with a glossy red surface, smiling with human teeth and wearing black sunglasses. Surrounded by realistic flames, the pepper is sharply in focus, while the fiery background is slightly blurred.}
    \end{subfigure}
    \begin{subfigure}[b]{0.32\textwidth}
        \centering
        \includegraphics[width=\linewidth]{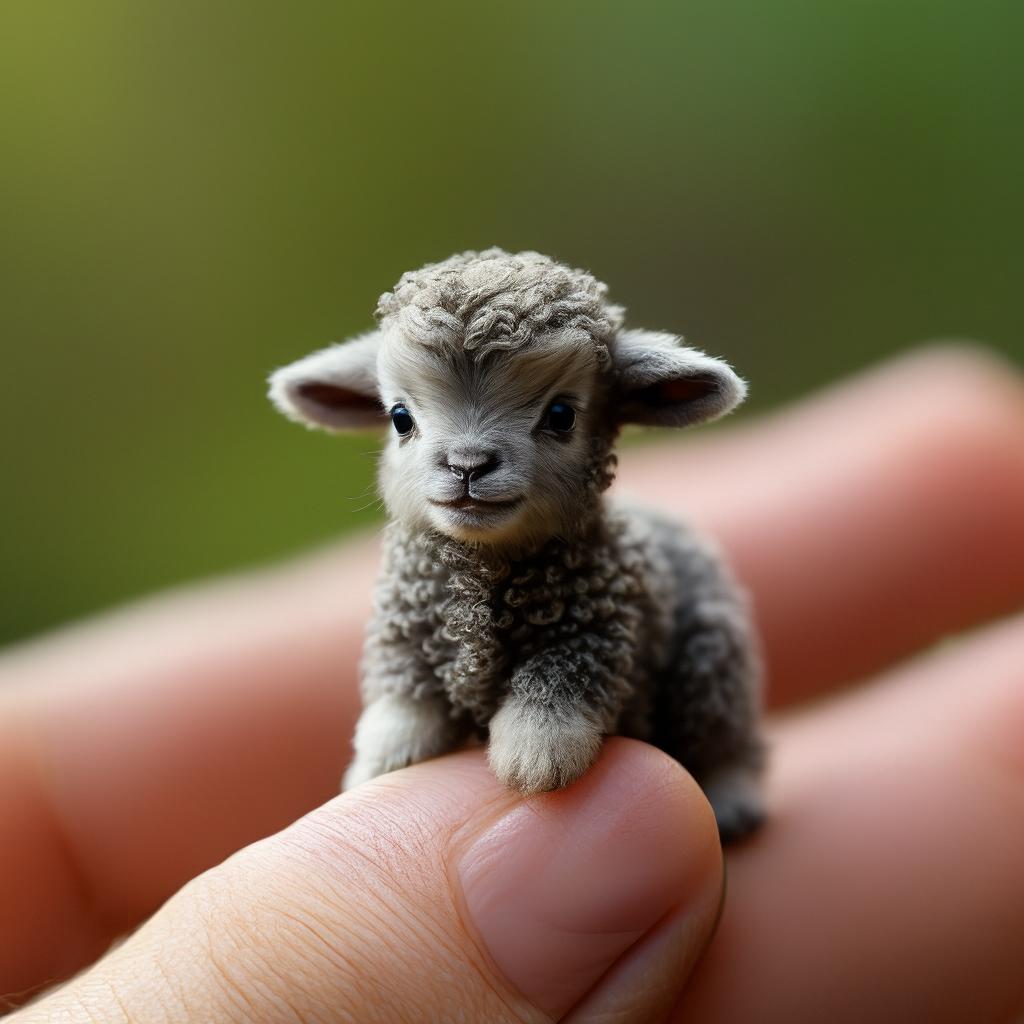}
        \caption*{\scriptsize Tiny fluffy lamb standing on a fingertip, ultra-detailed wool texture, soft natural light.}
    \end{subfigure}
    \hfill
    \begin{subfigure}[b]{0.32\textwidth}
        \centering
        \includegraphics[width=\linewidth]{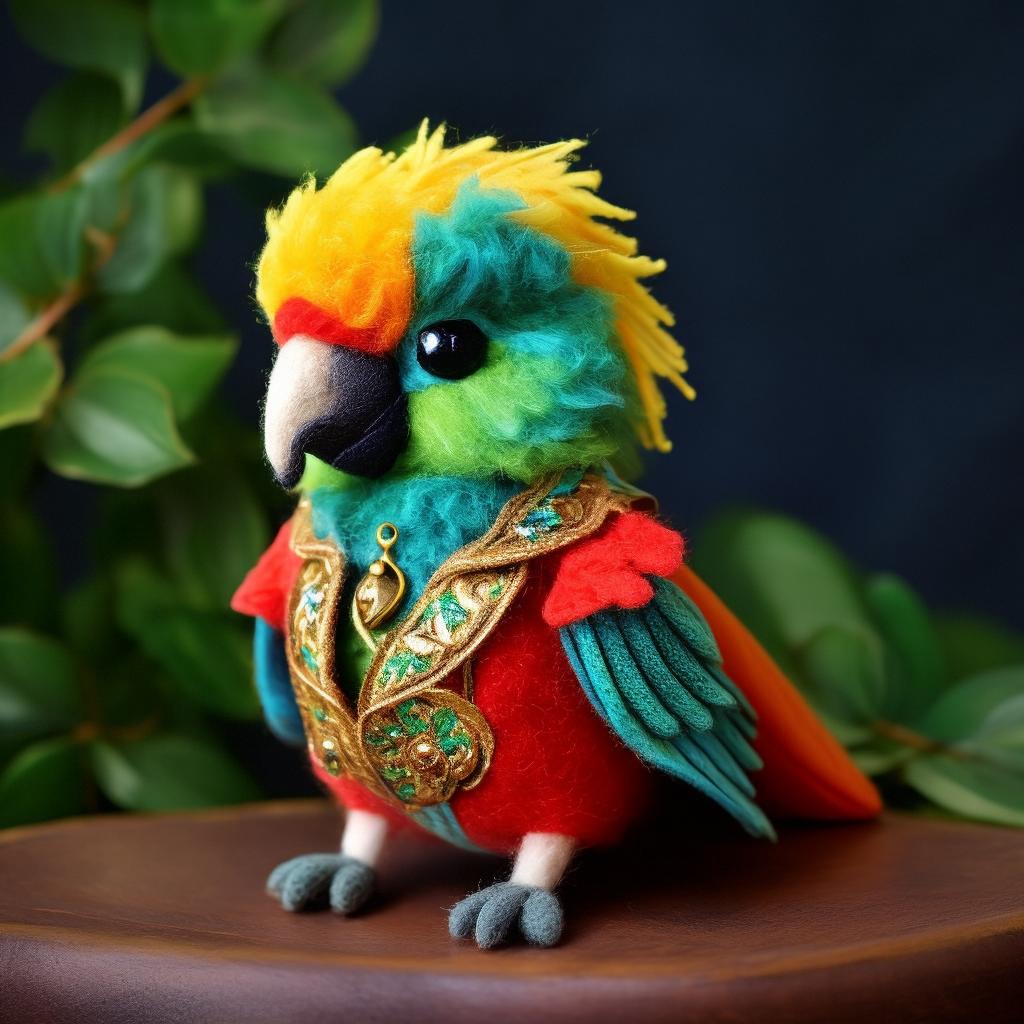}
        \caption*{\scriptsize Colorful woolen parrot plush wearing ornate psychic-style vest, detailed fabric textures.}
    \end{subfigure}
    \hfill
    \begin{subfigure}[b]{0.32\textwidth}
        \centering
        \includegraphics[width=\linewidth]{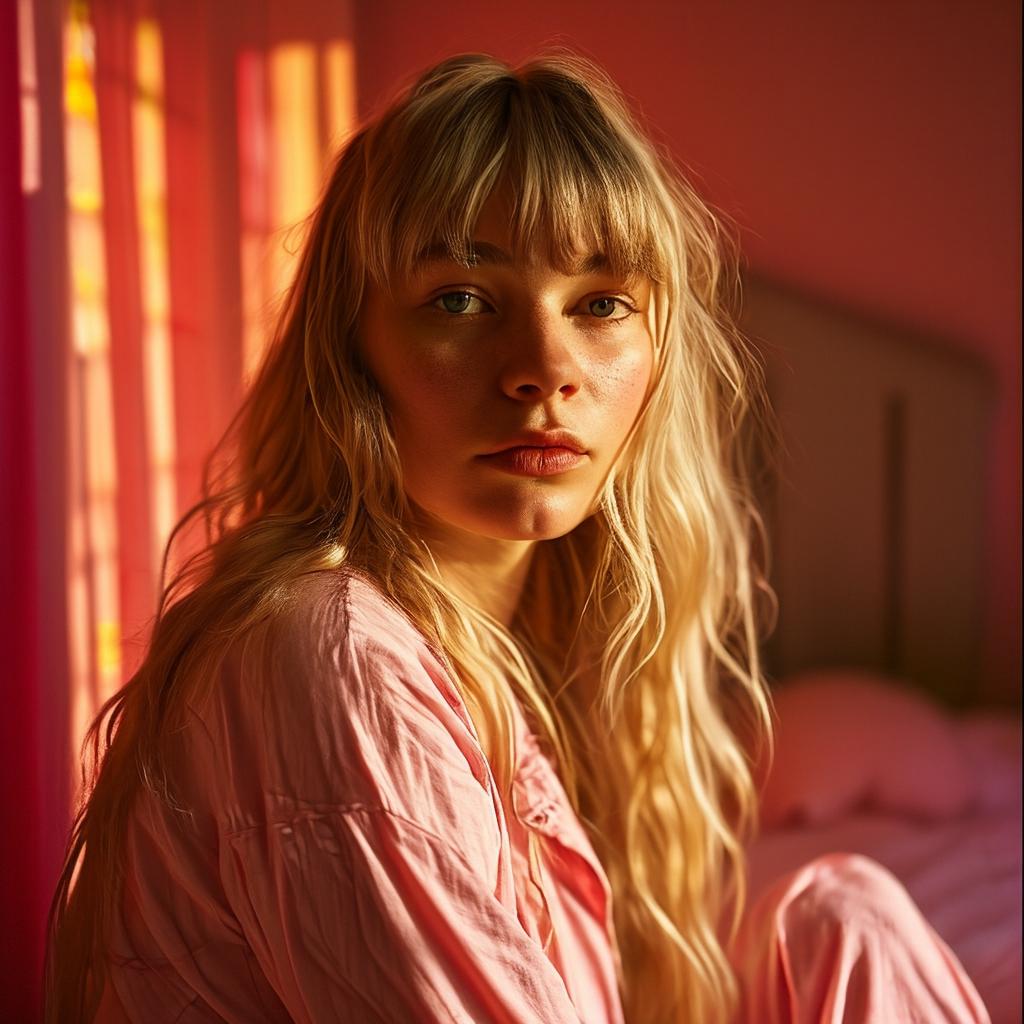}
        \caption*{\scriptsize young woman in warm sunset light, soft shadows, long blonde hair, pink pajamas, calm expression.}
    \end{subfigure}
    \caption{Additional Text-to-Image Generation Results. The caption below each image corresponds to the text prompt used to generate it.}
    \label{fig:app:t2i_qual_2}
\end{figure*}

\begin{figure*}[t]
    \centering
    \begin{subfigure}[b]{0.32\textwidth}
        \centering
        \includegraphics[width=\linewidth]{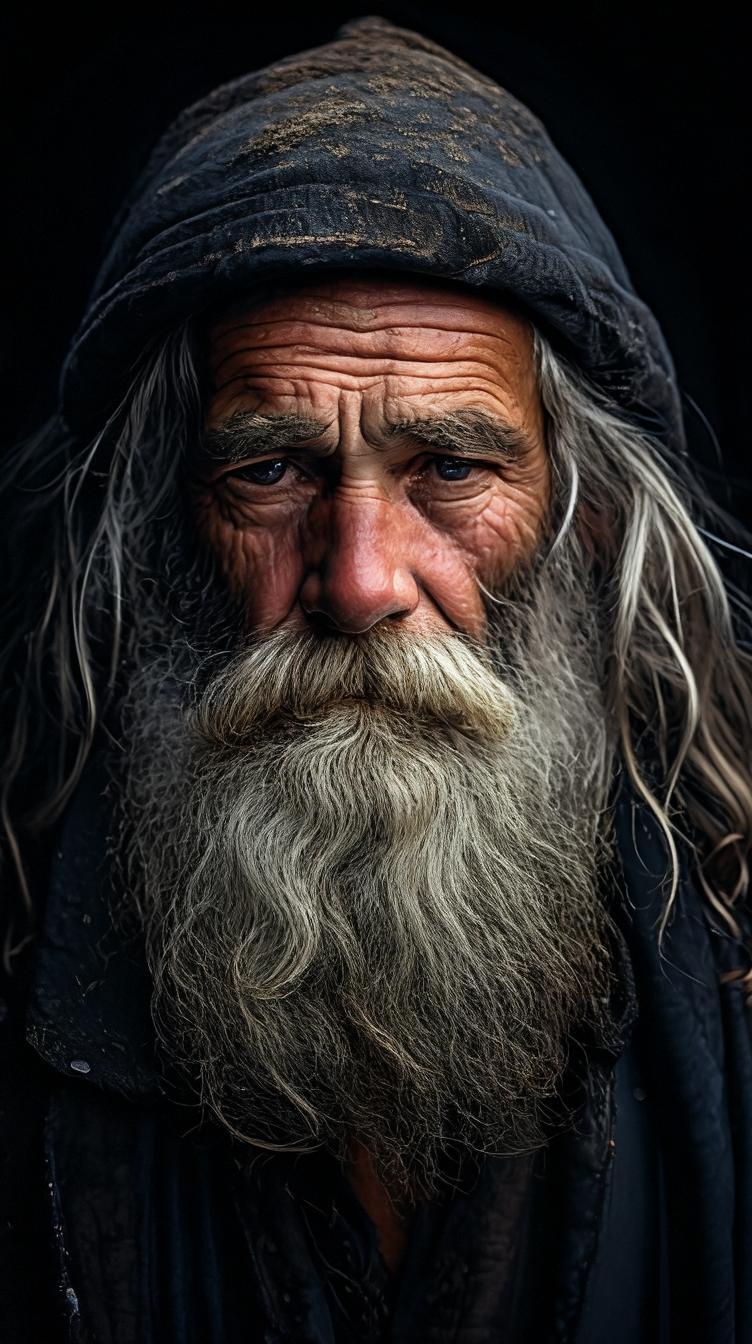}
        \caption*{An image of a very tired old man, fisherman, long beard, black background settings. The facial expression reflects wisdoms and test of time.}
        \vspace{1em}
    \end{subfigure}
    \hfill
    \begin{subfigure}[b]{0.32\textwidth}
        \centering
        \includegraphics[width=\linewidth]{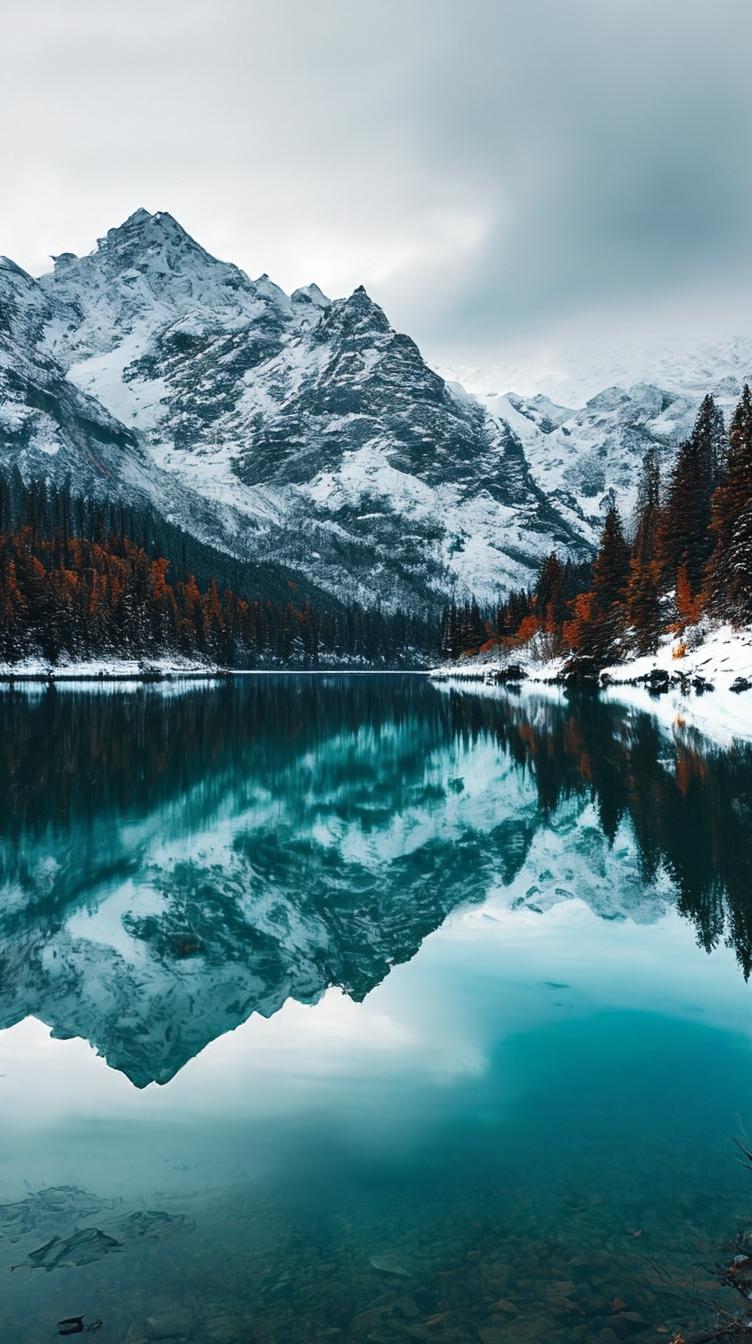}
        \caption*{snow-covered mountains rising above a calm turquoise lake, their peaks perfectly mirrored in the water, framed by dense autumn-tinged pines.}
        \vspace{1em}
    \end{subfigure}
    \hfill
    \begin{subfigure}[b]{0.32\textwidth}
        \centering
        \includegraphics[width=\linewidth]{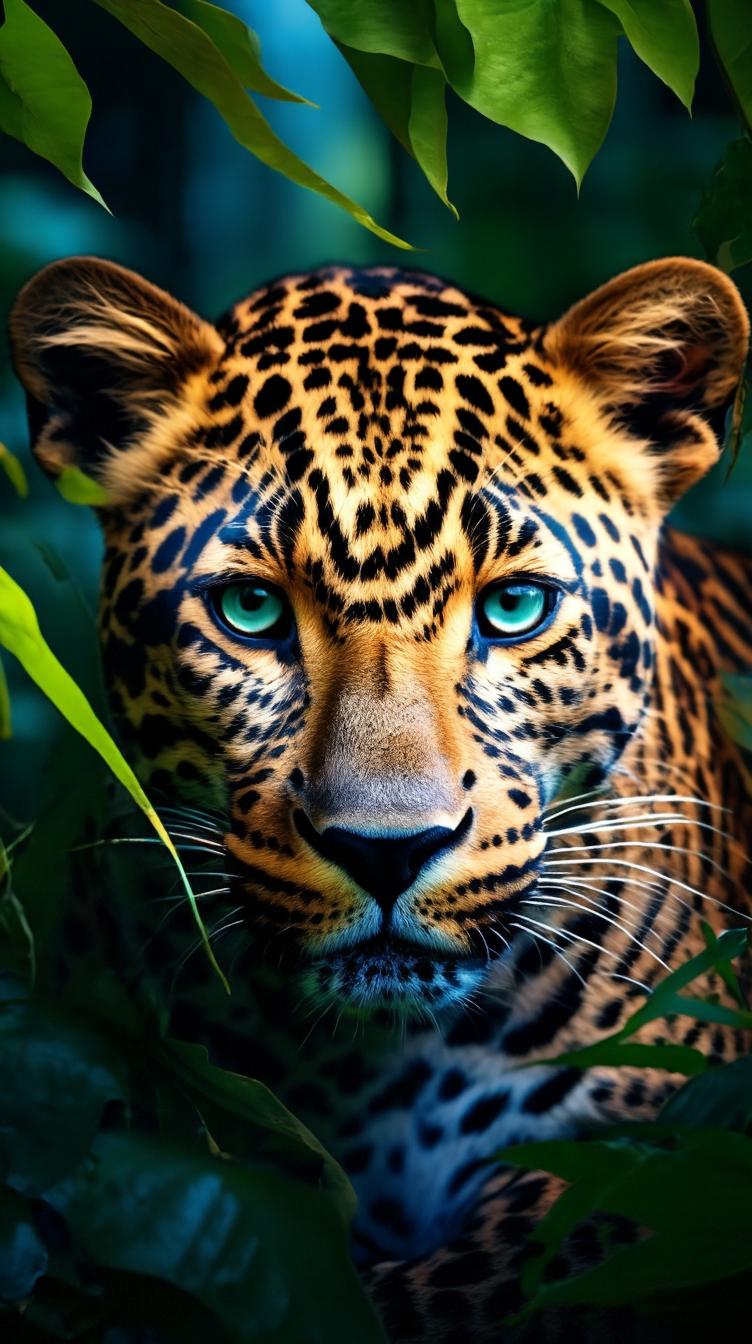}
        \caption*{A leopard hiding in the jungle, a photo-realistic portrait, captured with a Canon EOS R5 camera and a macro lens for detailed wildlife portraits.}
        \vspace{1em}
    \end{subfigure}
    
    \begin{subfigure}[b]{0.32\textwidth}
        \centering
        \includegraphics[width=\linewidth]{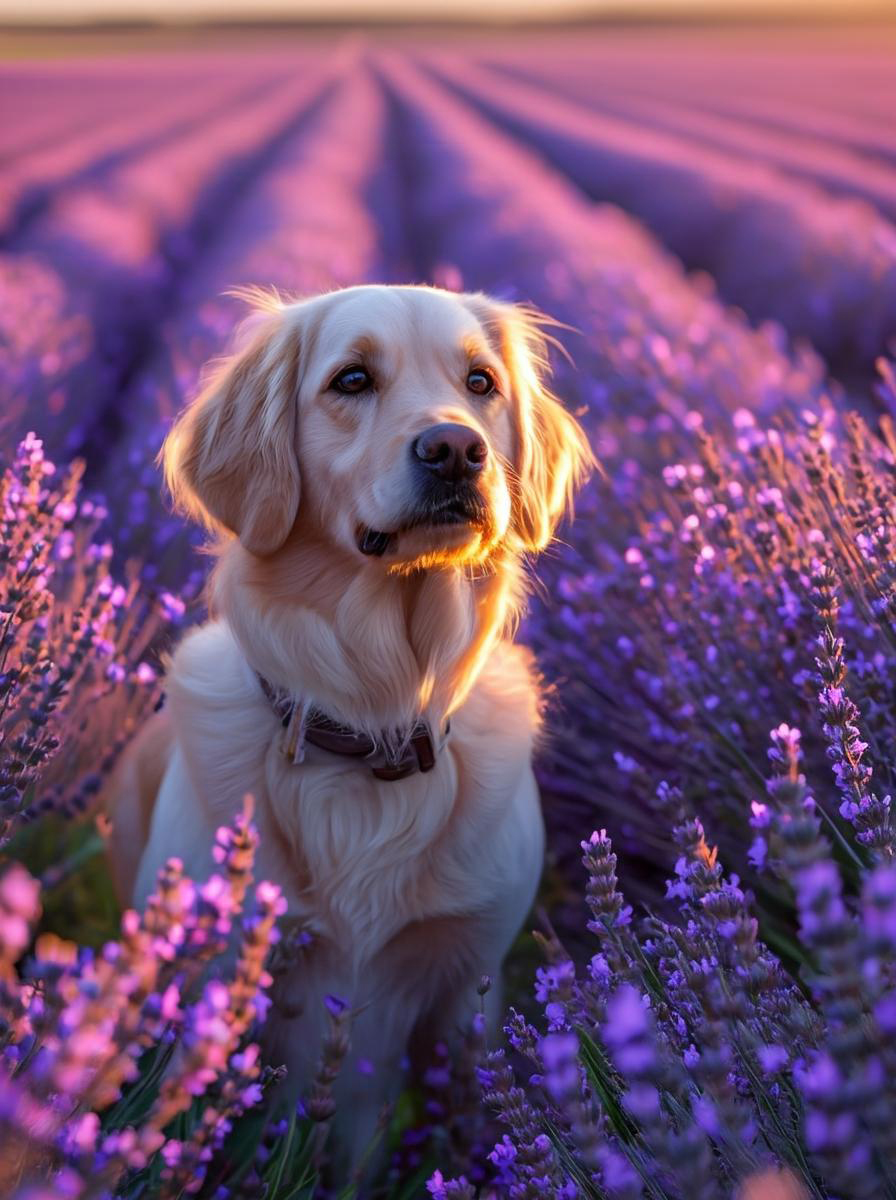}
        \caption*{A serene moment unfolds as a dog leisurely navigates a tranquil lavender field at sunset. The golden light enhances the beauty of the swaying purple blooms.}
    \end{subfigure}
    \hfill
    \begin{subfigure}[b]{0.32\textwidth}
        \centering
        \includegraphics[width=\linewidth]{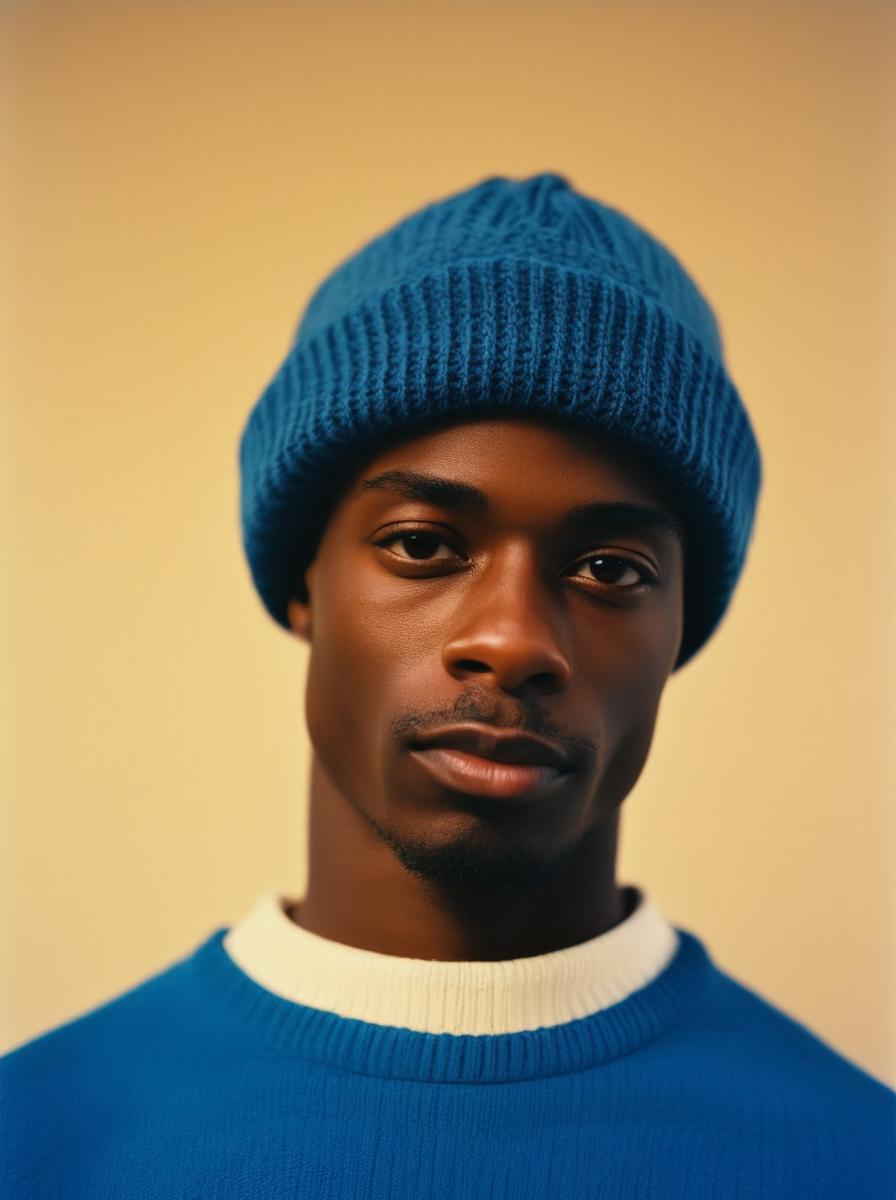}
        \caption*{Close-up shot of an African American man wearing a blue beanie, against a beige background, with a vintage aesthetic, in the style of Kodak film photography.}
    \end{subfigure}
    \hfill
    \begin{subfigure}[b]{0.32\textwidth}
        \centering
        \includegraphics[width=\linewidth]{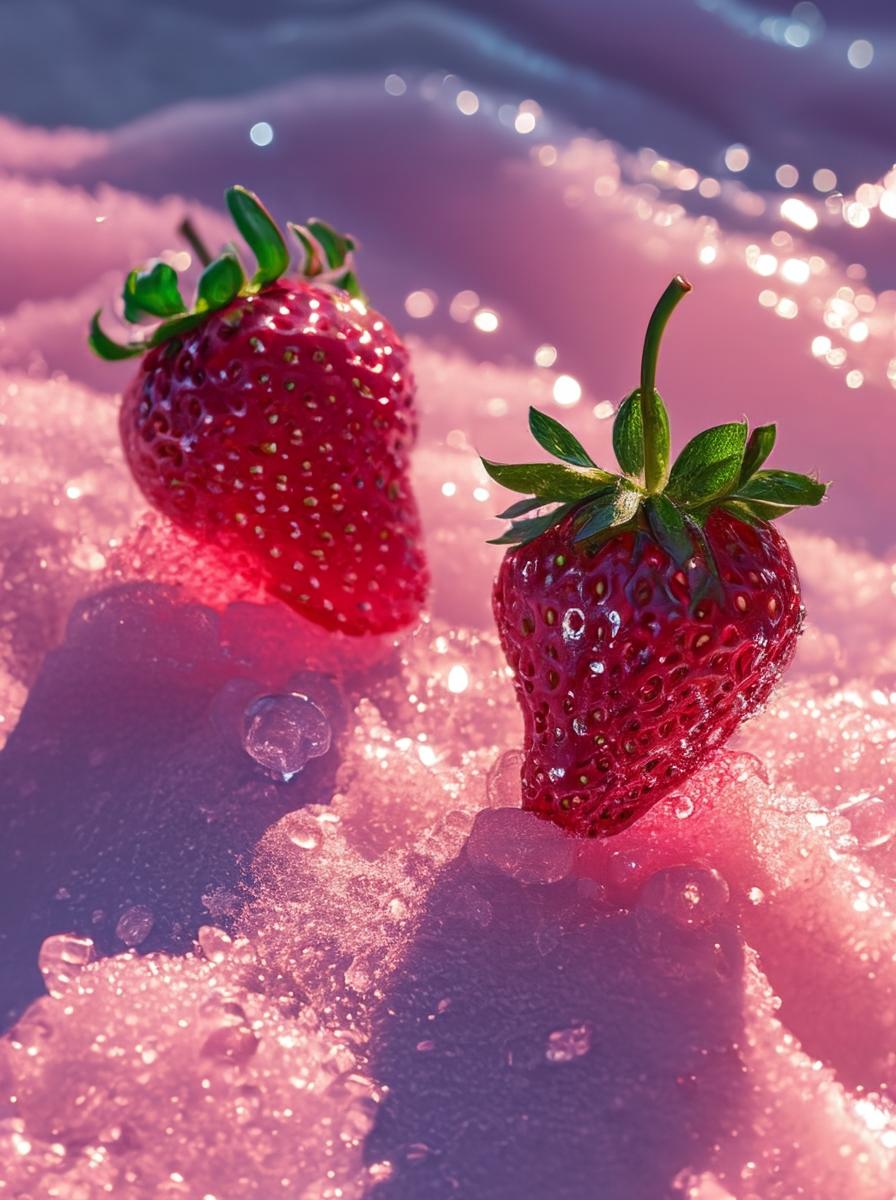}
        \caption*{Two conjoined strawberry. The strawberrys are resting on undulating red-pink holographic icy slush. Vaporwave. Intense color. Ice textures. Solar aesthetic.}
    \end{subfigure}
    \caption{Additional Text-to-Image Generation Results. The caption below each image corresponds to the text prompt used to generate it.}
    \label{fig:app:t2i_qual_3}
\end{figure*}

\begin{figure}[t]
    \centering
    \setlength{\tabcolsep}{0pt}
    \renewcommand{\arraystretch}{0}
    \begin{tabular}{cccc}
        \multicolumn{2}{c}{\includegraphics[width=0.50\columnwidth]{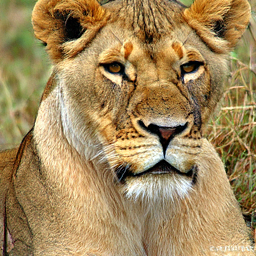}} &
        \multicolumn{2}{c}{\includegraphics[width=0.50\columnwidth]{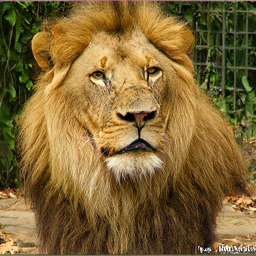}} \\
        \includegraphics[width=0.25\columnwidth]{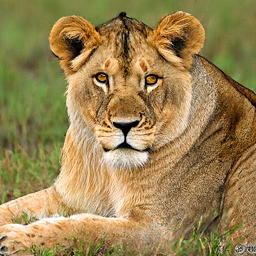} &
        \includegraphics[width=0.25\columnwidth]{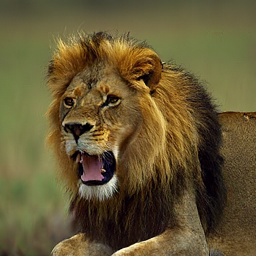} &
        \includegraphics[width=0.25\columnwidth]{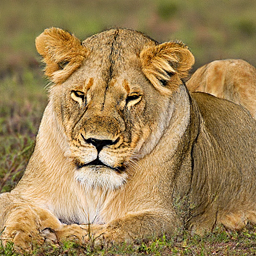} &
        \includegraphics[width=0.25\columnwidth]{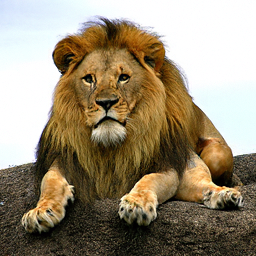} \\
        \includegraphics[width=0.25\columnwidth]{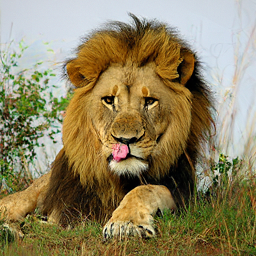} &
        \includegraphics[width=0.25\columnwidth]{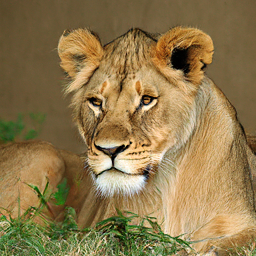} &
        \includegraphics[width=0.25\columnwidth]{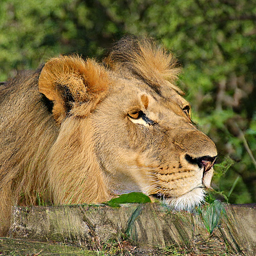} &
        \includegraphics[width=0.25\columnwidth]{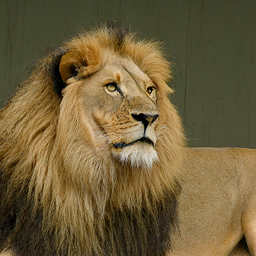}
    \end{tabular}
    \captionof{figure}{\textbf{Uncurated ImageNet 256$\times$256 \model{}-XL samples.} Class 291. CFG scale = 4.0.}
    \label{fig:app:qual_class291}
    \vspace{1em}
    \setlength{\tabcolsep}{0pt}
    \renewcommand{\arraystretch}{0}
    \begin{tabular}{cccc}
        \multicolumn{2}{c}{\includegraphics[width=0.50\columnwidth]{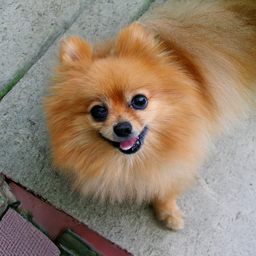}} &
        \multicolumn{2}{c}{\includegraphics[width=0.50\columnwidth]{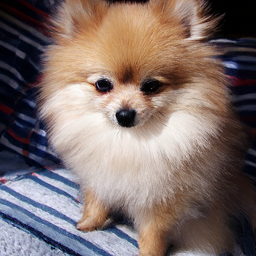}} \\
        \includegraphics[width=0.25\columnwidth]{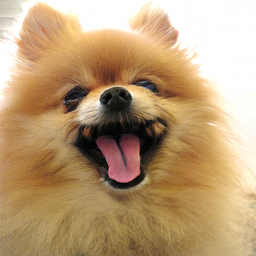} &
        \includegraphics[width=0.25\columnwidth]{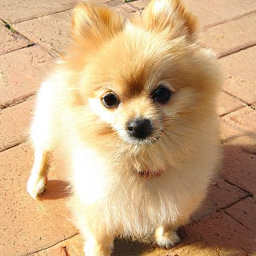} &
        \includegraphics[width=0.25\columnwidth]{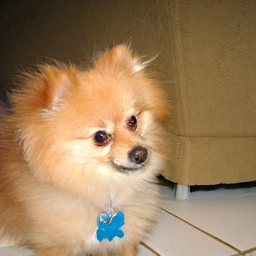} &
        \includegraphics[width=0.25\columnwidth]{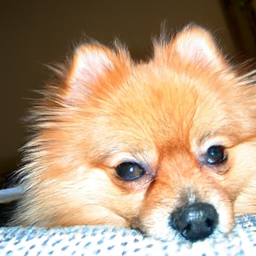} \\
        \includegraphics[width=0.25\columnwidth]{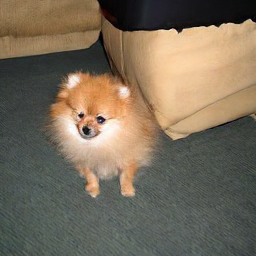} &
        \includegraphics[width=0.25\columnwidth]{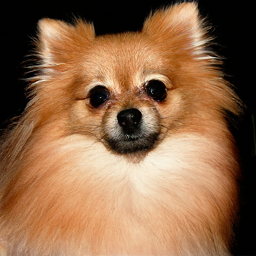} &
        \includegraphics[width=0.25\columnwidth]{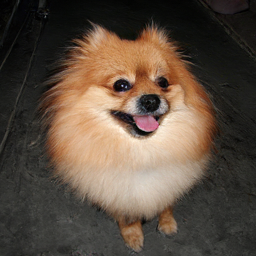} &
        \includegraphics[width=0.25\columnwidth]{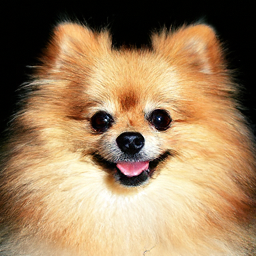}
    \end{tabular}
    \captionof{figure}{\textbf{Uncurated ImageNet 256$\times$256 \model{}-XL samples.} Class 259. CFG scale = 4.0.}
    \label{fig:app:qual_class259}
\end{figure}

\begin{figure}[t]
    \centering
    \setlength{\tabcolsep}{0pt}
    \renewcommand{\arraystretch}{0}
    \begin{tabular}{cccc}
        \multicolumn{2}{c}{\includegraphics[width=0.50\columnwidth]{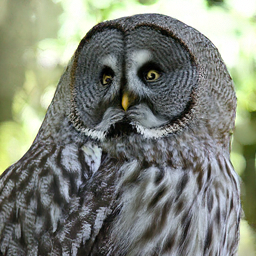}} &
        \multicolumn{2}{c}{\includegraphics[width=0.50\columnwidth]{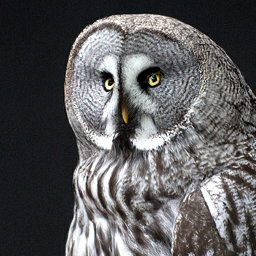}} \\
        \includegraphics[width=0.25\columnwidth]{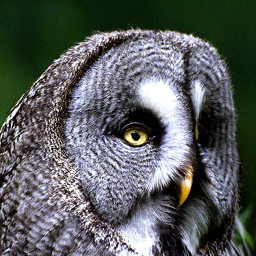} &
        \includegraphics[width=0.25\columnwidth]{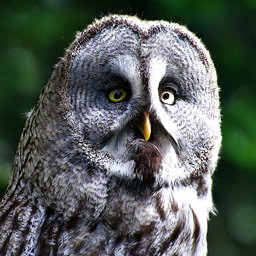} &
        \includegraphics[width=0.25\columnwidth]{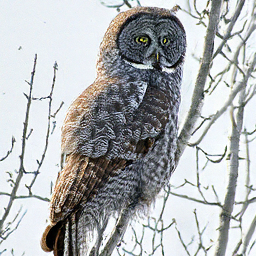} &
        \includegraphics[width=0.25\columnwidth]{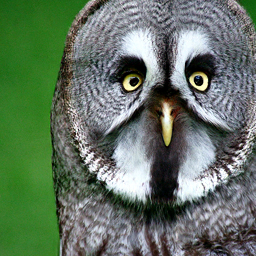} \\
        \includegraphics[width=0.25\columnwidth]{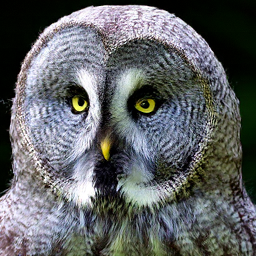} &
        \includegraphics[width=0.25\columnwidth]{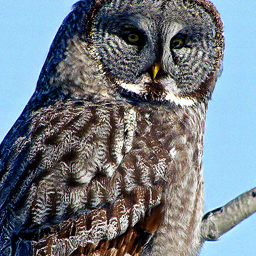} &
        \includegraphics[width=0.25\columnwidth]{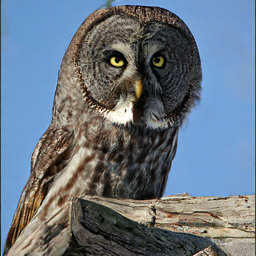} &
        \includegraphics[width=0.25\columnwidth]{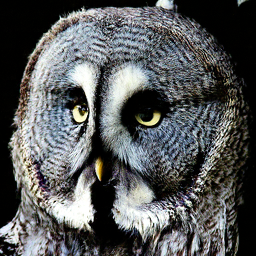}
    \end{tabular}
    \captionof{figure}{\textbf{Uncurated ImageNet 256$\times$256 \model{}-XL samples.} Class 24. CFG scale = 4.0.}
    \label{fig:app:qual_class24}
    \vspace{1em}
    \setlength{\tabcolsep}{0pt}
    \renewcommand{\arraystretch}{0}
    \begin{tabular}{cccc}
        \multicolumn{2}{c}{\includegraphics[width=0.50\columnwidth]{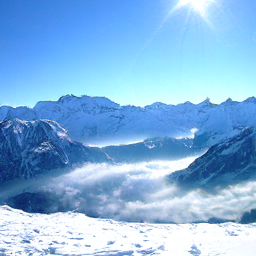}} &
        \multicolumn{2}{c}{\includegraphics[width=0.50\columnwidth]{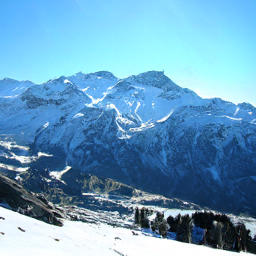}} \\
        \includegraphics[width=0.25\columnwidth]{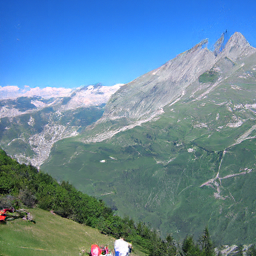} &
        \includegraphics[width=0.25\columnwidth]{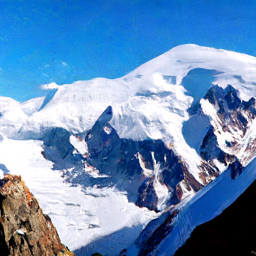} &
        \includegraphics[width=0.25\columnwidth]{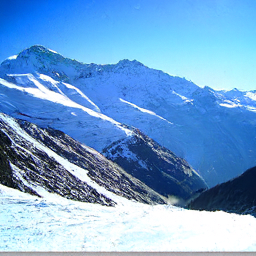} &
        \includegraphics[width=0.25\columnwidth]{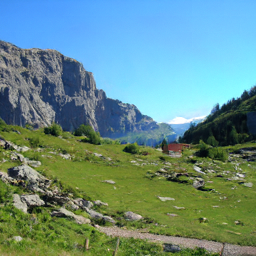} \\
        \includegraphics[width=0.25\columnwidth]{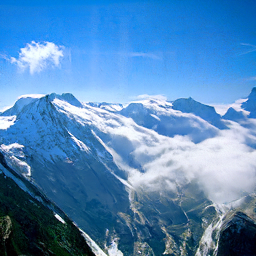} &
        \includegraphics[width=0.25\columnwidth]{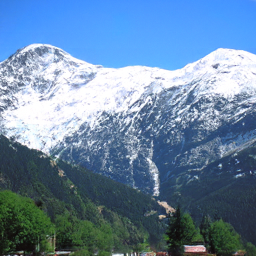} &
        \includegraphics[width=0.25\columnwidth]{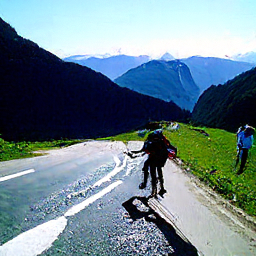} &
        \includegraphics[width=0.25\columnwidth]{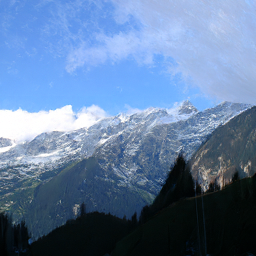}
    \end{tabular}
    \captionof{figure}{\textbf{Uncurated ImageNet 256$\times$256 \model{}-XL samples.} Class 970. CFG scale = 4.0.}
    \label{fig:app:qual_class970}
\end{figure}

\begin{figure}[t]
    \centering
    \setlength{\tabcolsep}{0pt}
    \renewcommand{\arraystretch}{0}
    \begin{tabular}{cccc}
        \multicolumn{2}{c}{\includegraphics[width=0.50\columnwidth]{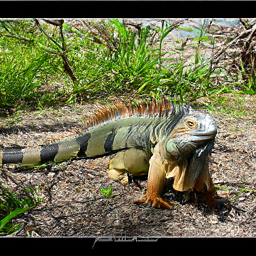}} &
        \multicolumn{2}{c}{\includegraphics[width=0.50\columnwidth]{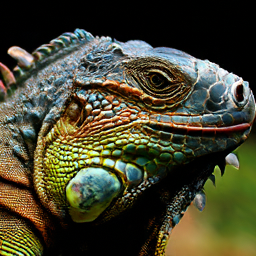}} \\
        \includegraphics[width=0.25\columnwidth]{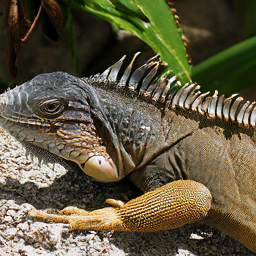} &
        \includegraphics[width=0.25\columnwidth]{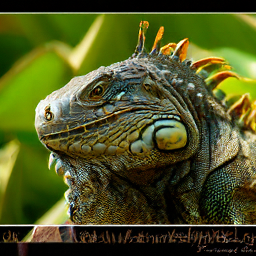} &
        \includegraphics[width=0.25\columnwidth]{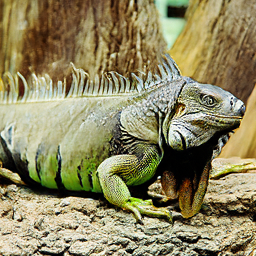} &
        \includegraphics[width=0.25\columnwidth]{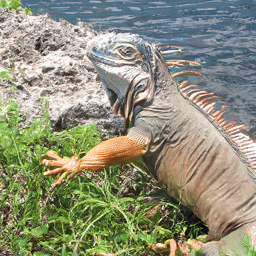} \\
        \includegraphics[width=0.25\columnwidth]{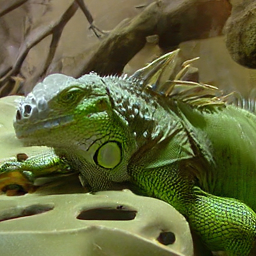} &
        \includegraphics[width=0.25\columnwidth]{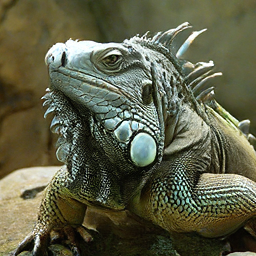} &
        \includegraphics[width=0.25\columnwidth]{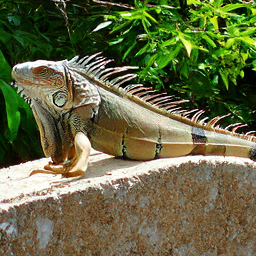} &
        \includegraphics[width=0.25\columnwidth]{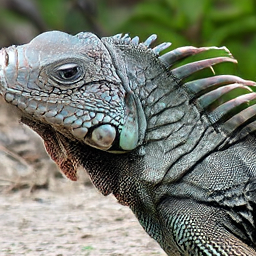}
    \end{tabular}
    \captionof{figure}{\textbf{Uncurated ImageNet 256$\times$256 \model{}-XL samples.} Class 39. CFG scale = 4.0.}
    \label{fig:app:qual_class39}
    \vspace{1em}
    \setlength{\tabcolsep}{0pt}
    \renewcommand{\arraystretch}{0}
    \begin{tabular}{cccc}
        \multicolumn{2}{c}{\includegraphics[width=0.50\columnwidth]{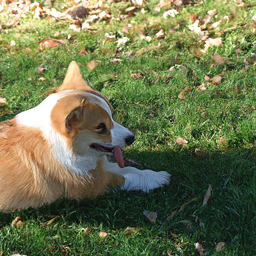}} &
        \multicolumn{2}{c}{\includegraphics[width=0.50\columnwidth]{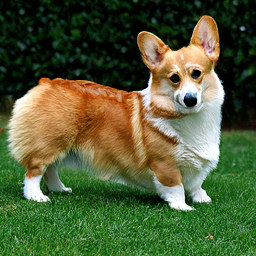}} \\
        \includegraphics[width=0.25\columnwidth]{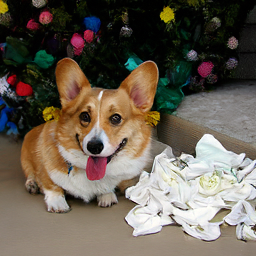} &
        \includegraphics[width=0.25\columnwidth]{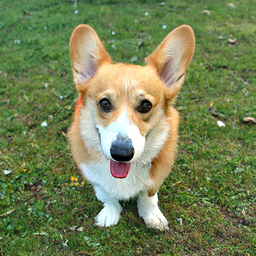} &
        \includegraphics[width=0.25\columnwidth]{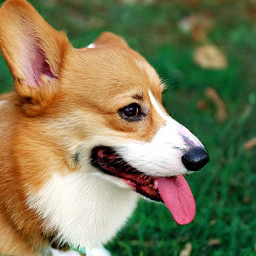} &
        \includegraphics[width=0.25\columnwidth]{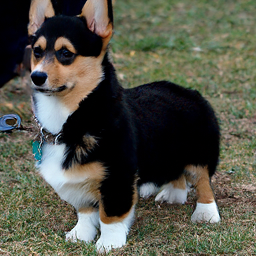} \\
        \includegraphics[width=0.25\columnwidth]{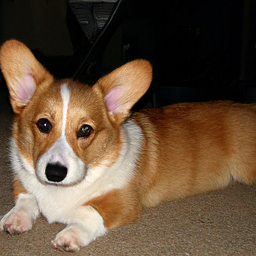} &
        \includegraphics[width=0.25\columnwidth]{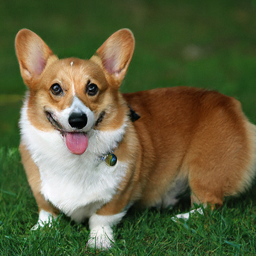} &
        \includegraphics[width=0.25\columnwidth]{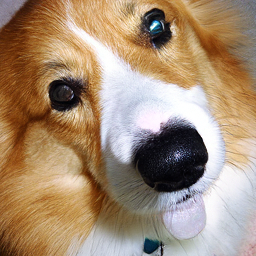} &
        \includegraphics[width=0.25\columnwidth]{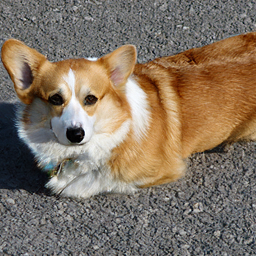}
    \end{tabular}
    \captionof{figure}{\textbf{Uncurated ImageNet 256$\times$256 \model{}-XL samples.} Class 263. CFG scale = 4.0.}
    \label{fig:app:qual_class263}
\end{figure}

\begin{figure}[t]
    \centering
    \setlength{\tabcolsep}{0pt}
    \renewcommand{\arraystretch}{0}
    \begin{tabular}{cccc}
        \multicolumn{2}{c}{\includegraphics[width=0.50\columnwidth]{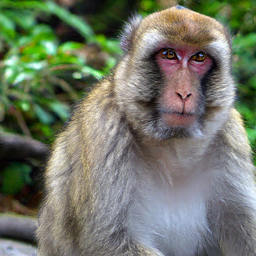}} &
        \multicolumn{2}{c}{\includegraphics[width=0.50\columnwidth]{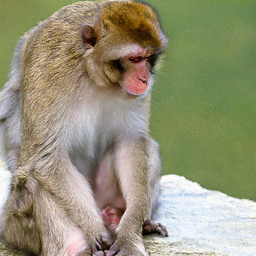}} \\
        \includegraphics[width=0.25\columnwidth]{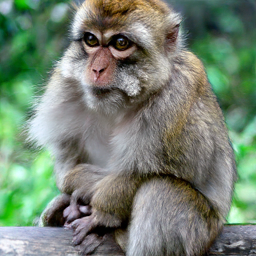} &
        \includegraphics[width=0.25\columnwidth]{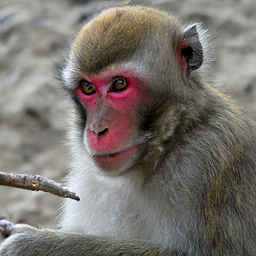} &
        \includegraphics[width=0.25\columnwidth]{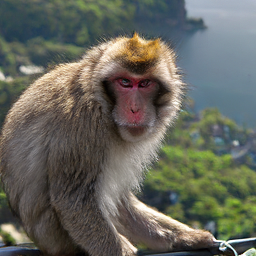} &
        \includegraphics[width=0.25\columnwidth]{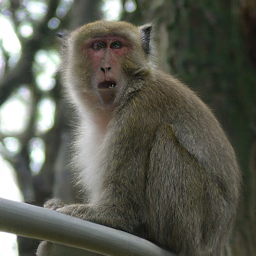} \\
        \includegraphics[width=0.25\columnwidth]{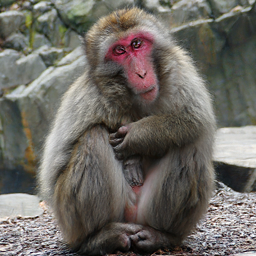} &
        \includegraphics[width=0.25\columnwidth]{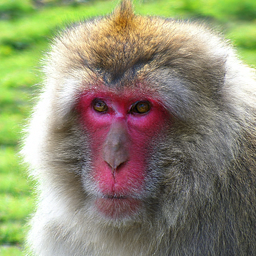} &
        \includegraphics[width=0.25\columnwidth]{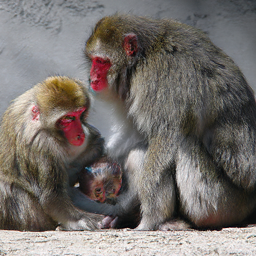} &
        \includegraphics[width=0.25\columnwidth]{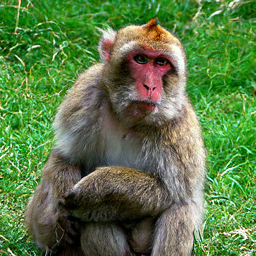} 
    \end{tabular}
    \captionof{figure}{\textbf{Uncurated ImageNet 256$\times$256 \model{}-XL samples.} Class 373. CFG scale = 4.0.}
    \label{fig:app:qual_class373}
    \vspace{1em}
    \setlength{\tabcolsep}{0pt}
    \renewcommand{\arraystretch}{0}
    \begin{tabular}{cccc}
        \multicolumn{2}{c}{\includegraphics[width=0.50\columnwidth]{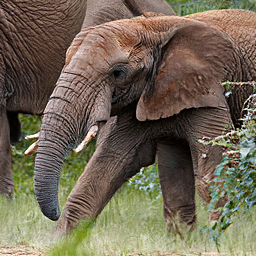}} &
        \multicolumn{2}{c}{\includegraphics[width=0.50\columnwidth]{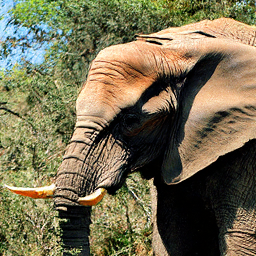}} \\
        \includegraphics[width=0.25\columnwidth]{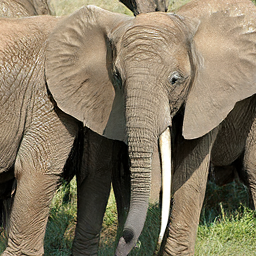} &
        \includegraphics[width=0.25\columnwidth]{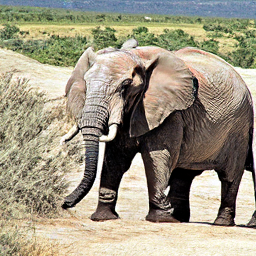} &
        \includegraphics[width=0.25\columnwidth]{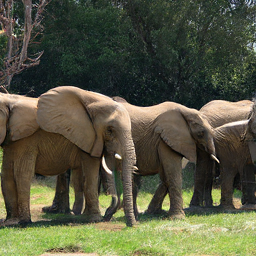} &
        \includegraphics[width=0.25\columnwidth]{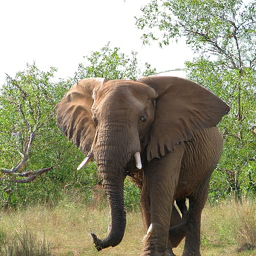} \\
        \includegraphics[width=0.25\columnwidth]{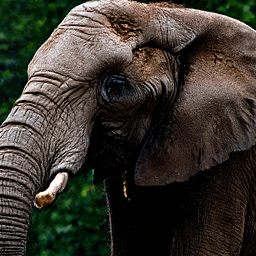} &
        \includegraphics[width=0.25\columnwidth]{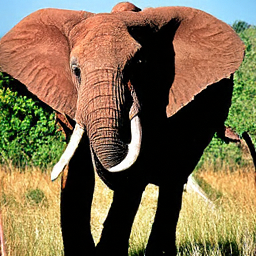} &
        \includegraphics[width=0.25\columnwidth]{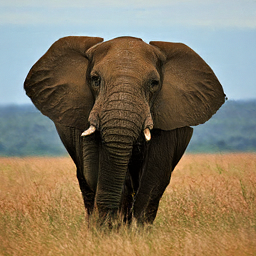} &
        \includegraphics[width=0.25\columnwidth]{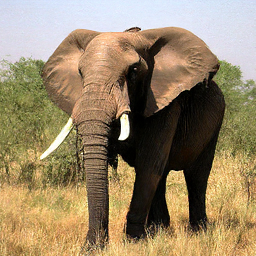} 
    \end{tabular}
    \captionof{figure}{\textbf{Uncurated ImageNet 256$\times$256 \model{}-XL samples.} Class 386. CFG scale = 4.0.}
    \label{fig:app:qual_class386}
\end{figure}

   We include additional qualitative results for ImageNet 256$\times$256 single-class-conditioned image generations, as shown in Figures~\ref{fig:app:qual_class291}--\ref{fig:app:qual_class386}, together with high-resolution (approximately $1024^2$) text-to-image generation results in Figures~\ref{fig:app:t2i_qual_0}--\ref{fig:app:t2i_qual_3}, illustrating the visual quality, diversity, and prompt alignment achieved by \model{}.

\section{Limitations}
Due to the limited model capacity and insufficient high-quality training data, our 1.3B-parameter PixelDiT text-to-image model sometimes struggles to generate objects that are both geometrically and texturally complex, such as human hands and intricate architectural scenes. Additionally, we observe that training pixel-space diffusion models with velocity prediction is prone to loss spikes, particularly for deeper architectures and during long training runs. Although we have identified several stabilization techniques (see Section~\ref{sec:empirical_insights}), fully eliminating loss spikes without sacrificing training efficiency remains an open challenge. In future work, we plan to address these limitations by scaling up the model capacity, curating larger and higher-quality training data, and conducting more foundational research into the training dynamics of pixel-space diffusion to better understand and mitigate loss instabilities.

\clearpage

\clearpage

\clearpage

\clearpage

\clearpage

\end{document}